\documentclass[10pt,twocolumn,letterpaper]{article}

\usepackage[pagenumbers]{iccv} % To force page numbers, e.g. for an arXiv version

\newtheorem{definition}{Definition}
\newtheorem{theorem}{Theorem}

\newtheorem{lemma}[theorem]{Lemma}

\usepackage{adjustbox,booktabs,multirow}
\usepackage{amsmath,bm}

\definecolor{iccvblue}{rgb}{0.21,0.49,0.74}
\usepackage[pagebackref,breaklinks,colorlinks,allcolors=iccvblue]{hyperref}

\def\paperID{*****} % *** Enter the Paper ID here
\def\confName{CVPR}
\def\confYear{2026}

\title{The Mechanics of CNN Filtering with Rectification}

\author{Liam Frija-Altarac\\
\'{E}cole de Technologie Sup\'{e}rieure\\
Montreal, Quebec, Canada\\
{\tt\small liam.frija-altarac.1@ens.etsmtl.ca }
\and
Matthew Toews\\
\'{E}cole de Technologie Sup\'{e}rieure\\
Montreal, Quebec, Canada\\
{\tt\small matt.toews@gmail.com}
}

\begin{document}

\maketitle
\begin{abstract}

This paper proposes elementary information mechanics as a new model for understanding the mechanical properties of convolutional filtering with rectification, inspired by physical theories of special relativity and quantum mechanics. We consider kernels decomposed into orthogonal even and odd components. Even components cause image content to diffuse isotropically while preserving the center of mass, analogously to rest or potential energy with zero net momentum. Odd kernels cause directional displacement of the center of mass, analogously to kinetic energy with non-zero momentum. The speed of information displacement is linearly related to the ratio of odd vs total kernel energy. Even-Odd properties are analyzed in the spectral domain via the discrete cosine transform (DCT), where the structure of small convolutional filters (e.g. $3 \times 3$ pixels) is dominated by low-frequency bases, specifically the DC $\Sigma$ and gradient components $\nabla$, which define the fundamental modes of information propagation. To our knowledge, this is the first work demonstrating the link between information processing in generic CNNs and the energy-momentum relation, a cornerstone of modern relativistic physics.

%This paper proposes elementary information mechanics as a new model for understanding the action of convolutional neural network (CNN) filtering, inspired by physical theories of special relativity and quantum mechanics. Filters are represented as orthogonal odd and even functions, these are shown to be dominated by sum $\Sigma$ and gradient $\nabla$ components respectively, which account for over 90\% of classification accuracy in generic CNN models like VGG16 and Resnet50. Furthermore, under standard convolution + ReLU non-linearity these components leads to three distinct modes of information propagation in activation space, namely 1) a symmetric diffusion mode associated with the sum $\Sigma$ operator and 2) bi-directional vibration or 3) uni-directional translation modes associated with the gradient operator $\nabla$. Information is attenuated by directional vibration and translation modes based on the relative orientations of image and filter gradients, in a manner analogous to Malus' law of polarized optical filtering, and novel visualizations demonstrate that this physical law is learned in CNN training.
\end{abstract}

\section{Introduction}
\label{sec:intro}
%from input-output examples and variants of the back-propagation algorithm~\cite{rumelhart1985learning} 
Deep neural networks are now ubiquitous in virtually all artificial intelligence (AI) and computer vision tasks and were recognized by the 2024 Nobel prizes, in physics for networks based on energy minimization (Boltzmann machines, Hopfield networks~\cite{hinton1984boltzmann,hopfield1982neural}) and in chemistry for predicting protein folding patterns (AlphaFold~\cite{jumper2021highly}). Advantageously, a variety of different network architectures may be trained to solve the same task (e.g. ImageNet classification) without specialized filter design or knowledge of internal signal processing details, however this has also lead to criticism of their black-box nature~\cite{rudin2019stop}. Are there fundamental processes that govern the propagation of information through activation layers of generic neural networks, similarly to the physical processes governing the propagation of energy and mass through real world space?

\begin{figure}
  \centering
\includegraphics[width=\linewidth]{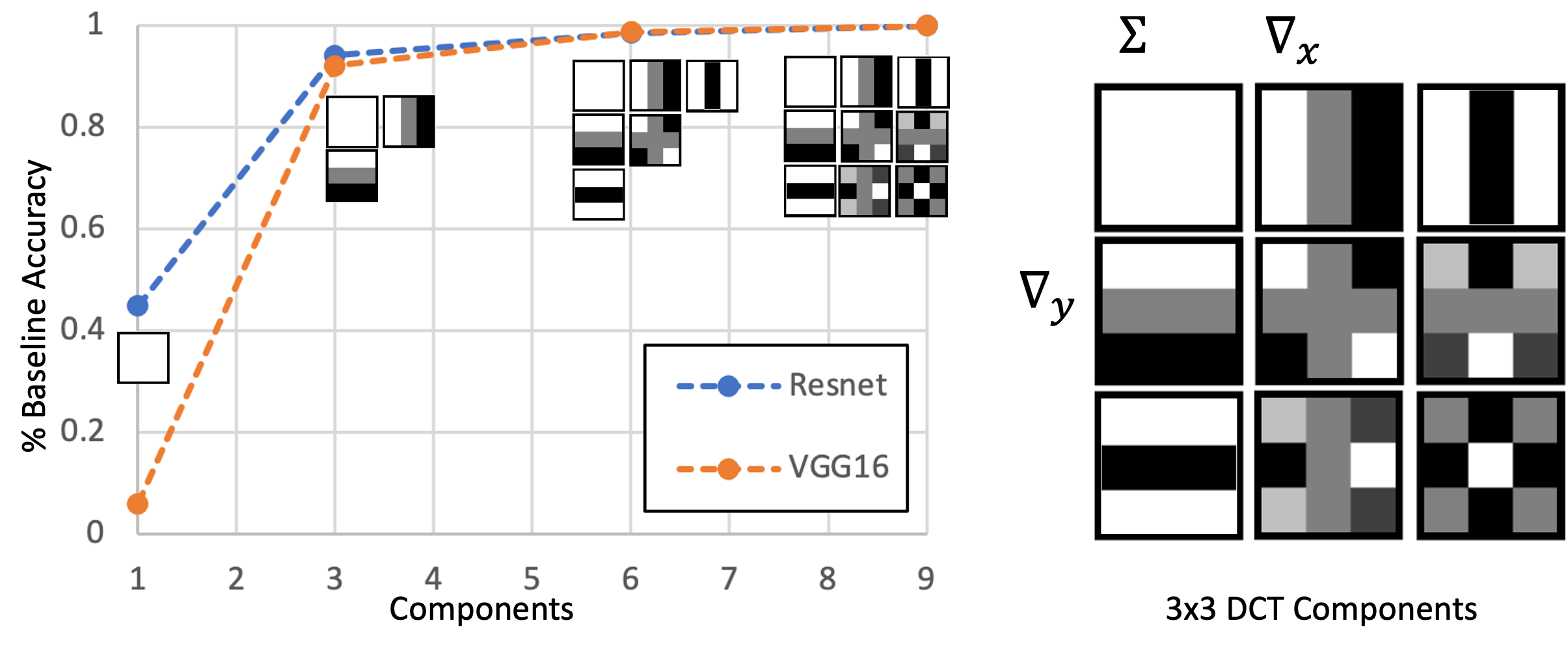} \\
  \caption{Motivating example: training using filters limited to DCT component subsets reveals that the three lowest frequency components (labelled $\Sigma,\nabla_x, \nabla_y$) account the majority of baseline accuracy for Resnet and VGG16 models (94\% and 92\% respectively). Details are provided in \Cref{dct_experiments}.}
  \label{fig:observation}
\end{figure}

We address this question with a new theory called elementary information mechanics, where fundamental frequency components common to all filters account for the majority of network accuracy and exhibit three notable modes of mechanical information propagation: diffusion, vibration and translation. Our theory is motivated by the novel result shown in Figure~\ref{fig:observation}, where $ > 92\%$ of baseline CNN accuracy may be achieved using only the three lowest frequency components of the discrete cosine transform (DCT)~\footnote{The DCT is a well-known basis used for image and video compression, \eg the JPEG algorithm~\cite{wallace1992jpeg}, where low frequency components account for the majority of image information.}. These correspond to the DC or sum $\Sigma$ and gradients $\nabla_x, \nabla_y$, which we show lead to diffusion, vibration and translation of image information under standard convolution + non-linear ReLU operations. They may be interpreted as mass-like and momentum-like terms via analogy to relativistic quantum mechanics, we provide novel demonstrations of various aspects of this analogy.

We investigate elementary information mechanics in the context convolutional neural networks (CNN)~\cite{lecun1989backpropagation} and the ImageNet~\cite{imagenet} classification task, which launched the current GPU-based AI paradigm with AlexNet~\cite{krizhevsky2012imagenet} and led to a variety of unique architectural solutions to the same task \eg the VGG16\cite{simonyan2014very} and Resnet50~\cite{he2016deep} models. While these models are no longer state-of-the-art, their well-known basic structure is ideal for demonstrating the mechanics of information propagation through generic network layers. The remainder of paper describes the theory of elementary information mechanics, including the related work and background concepts, intuitive examples and demonstrations in 1D and 2D, and its application in the well-known VGG16 and Resnet50 networks.

\section{Related Works}

Modern deep learning began in 2012, when CNNs were trained using highly parallelized processing on a graphics processing unit (GPU)~\citep{krizhevsky2009learning}, achieving a significant increase in performance on the benchmark ImageNet dataset. This allowed the development of various CNN architectures such as VGG~\citep{simonyan2014very} and Resnet~\citep{he2016deep}. The success of CNN models included the use of ReLU non-linearity~\citep{nair2010rectified, maas2013rectifier}, correct weight initialization \citep{glorot10a,he15} and the use of small filter sizes, \ie $3\times3$ pixel sizes. ReLU is a major component of modern neural architectures, having replaced the sigmoid activation function that was widely used in earlier networks. 
Its main advantage lies in reducing the vanishing gradient problem that often arises with saturating nonlinearities such as the sigmoid~\citep{glorot2011deep}.

A primary novelty of this work is in analyzing filtering with rectification, a non-linear operation. Linear filtering, including wavelets~\citep{haar1909theorie,gabor1946theory} and binomial filtering~\citep{aubury1996binomial} in approximating Gaussian filtering~\cite{wells1986efficient}, has been studied for over a century, whereas rectified filtering has only recently become ubiquitous in Deep Neural Networks. We find distinct properties when analyzed in terms of even and odd filter components.  

Our work investigates specifically the mechanical action of filters, particularly even radially symmetric and odd components of small filters. An interesting trend in modern CNNs has been the use of low-resolution filters of odd square dimensions, most notably  $3\times3$. Intuitively, low-resolution filters allow a larger number of channels and improved classification,  $3\times3$  being the most popular choice for 2D CNNs~\cite{simonyan2014very,he2016deep}. %Even dimensions such as $2\times2$ are avoided, as the result of convolution cannot be stored symmetrically with respect to the original data, leading to an undesirable shift~\citep{wu2019convolutionevensizedkernelssymmetric}. 
%Nevertheless, \citet{wu2019convolutionevensizedkernelssymmetric} show that this issue can be mitigated through appropriate padding strategies, achieving comparable accuracies to $3 \times 3$ models while using fewer parameters.  
%Given the use of small filters, one might reasonably assume that useful aspects of 
Such small filters are dominated by primary low-frequency components, \ie DC and gradient components, which have distinct effects upon activation information under rectified convolution. Our work presents a novel result that indeed, CNN accuracy is dominated by the average and gradient spectral components, which may be generalized to arbitrary filter sizes through even and odd functions, \ie radial symmetry and antisymmetry.

A closely related work is that of~\citet{luo2016understanding} who examined how the effective receptive field (ERF) of filters evolves during training. They empirically and analytically observed that the ERF of a randomly initialized CNN grows linearly with depth $n$ at a rate of $\mathcal{O}(\sqrt{n})$, though, relative to the theoretical receptive field,  it shrinks at a rate of $\mathcal{O}(1/\sqrt{n})$. After training, however, \citet{luo2016understanding} observed that the ERF typically grows to match the theoretical receptive field, though the underlying cause remains unclear.
Our work suggests that the increase in size of the ERF is determined by the odd component, which causes directional displacement of information. 

Other works have also looked at the symmetric properties of filters~\cite{cohen2016group}. The symmetry of average filters was shown to rise toward the network output\cite{alsallakh2025symmetries}, and filters have been shown to cluster into small sets of  gradient-like kernels\cite{babaiee2024unveiling}. However, these works have not studied the mechanical effects under rectified convolution.

In signal processing, the frequency spectrum of natural images is known to follow an inverse power law, where the majority of spectral energy is concentrated into low-frequency components. Various spectral bases may be used. The discrete Fourier transform (DFT) uses the complex-valued exponential $e^{-i2\pi fx} = \cos(2\pi fx)+i\sin(2\pi fx)$ and integer frequency multiples $f = \{1,2,3,4,\dots\}$. The discrete cosine transform (DCT) is a real-valued basis widely used in image compression~\cite{wallace1992jpeg}, which when compared to the FFT contains twice the number cosine components $\cos(2\pi fx)$ sampled at half-integer frequency multiples $f = \{\frac{1}{2},1,\frac{3}{2},2,\dots\}$. The primary difference is in the boundary conditions imposed, DFT components represent strictly periodic boundaries while the DCT allows both periodic and antiperiodic boundaries, and thus is able to represent discontinuities in natural images with a smaller number of low frequency components~\cite{wallace1992jpeg}.

\begin{figure}
  \centering
\includegraphics[width=\linewidth]{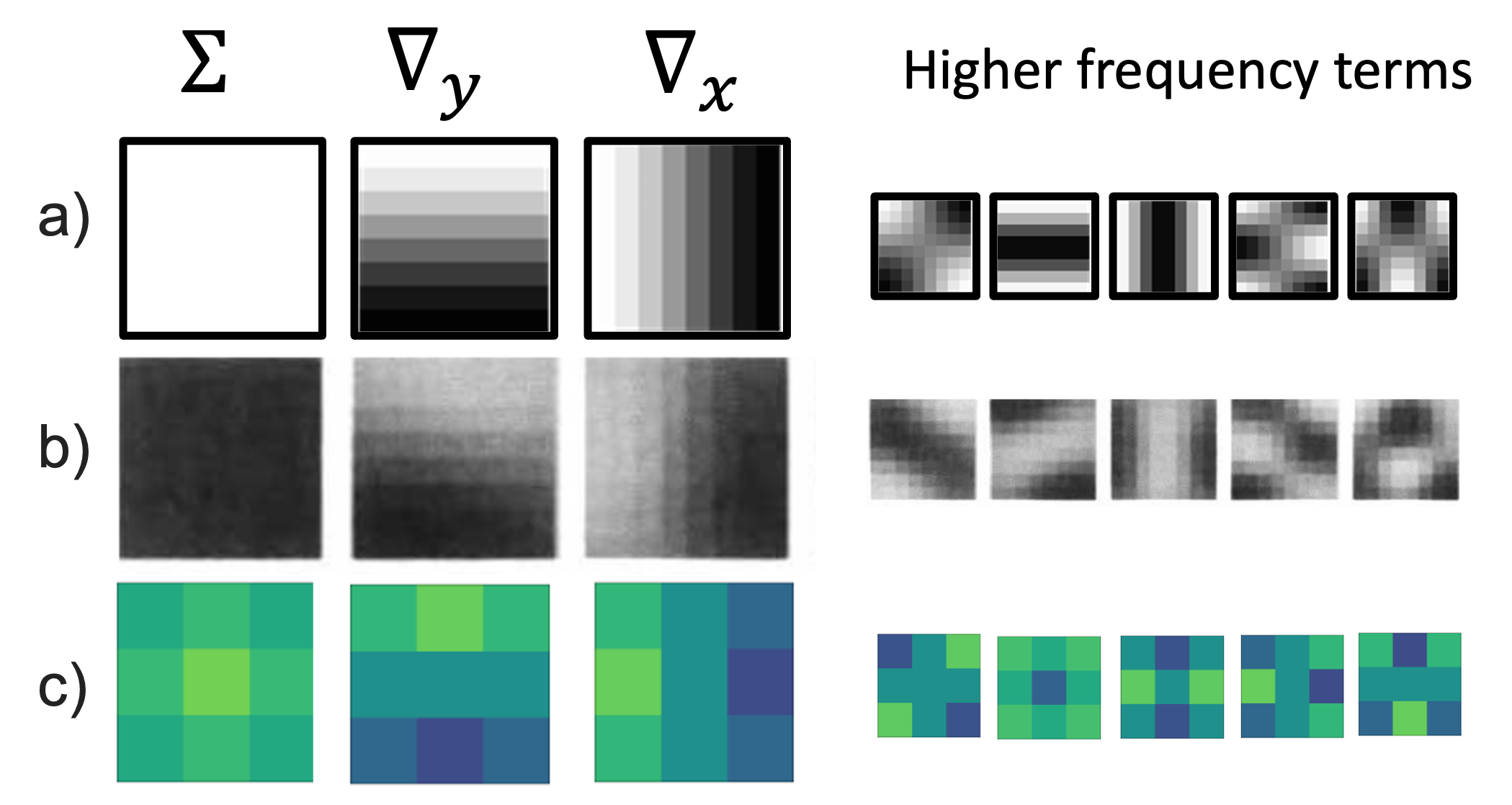}
  \caption{The three most significant components of real linear decompositions resemble sum $\Sigma$ and gradient $\nabla_x, \nabla_y$ filters, including a) JPEG DCT coefficients~\cite{wallace1992jpeg} b) Natural image PCs~\cite{olshausen1996emergence} c) CNN PCs (VGG16)~\cite{fukuzaki2022principal}.}
  \label{fig:Frequencies}
\end{figure}

Principal component analysis (PCA) provides a real-valued data-driven frequency decomposition, and as shown in Figure~\ref{fig:Frequencies}, the PCs of both generic CNN filters~\cite{fukuzaki2022principal} and natural images~\cite{olshausen1996emergence} reveal the discrete cosine transform (DCT) structure including sum $\Sigma$ and gradient $\nabla = \{\nabla_x,\nabla_y \}$ components. 
The DCT has recently been used to improve generative diffusion~\cite{ning2024dctdiff} and transformer~\cite{pan2024dct} models, however to our knowledge, the mechanical properties of DCT coefficients have not yet been studied, neither with or without rectification.

\section{Background}

This paper proposes elementary information mechanics, based on an analogy between relativistic quantum mechanics (QM) in the physical world and information processing in generic CNN activation image space, which bear notable superficial similarities. In both, the state of energy or information is organized within an image $I(x,y,z,t)$ of up to three spatial dimensions $(x,y,z)$ and one time or layer dimension $t$. In both systems, energy or information transform over time as a result of linear operators. 

In physics, special relativity (SR)~\cite{einstein1905electrodynamics} assumes that the laws of physics to applied identically at all inertial reference frames (\ie invariance), that the maximum speed of light through physical space is a constant $c = \frac{dx}{dt}=3e8$meters/second. Furthermore, the rest energy of a stationary mass $m$ is equal to $E=mc^2$~\cite{einstein1905does}. Quantum mechanics~\cite{griffiths2020introduction} (QM) relates the energy $E$, rest mass $m$ and momentum $p$ of elementary particles within the framework of SR, where two equations are pertinent in our analogy. The first is the quadratic energy-momentum relation~\eqref{eq:energy-momentum} equating the squared energy $E^2$ of a particle to the sum of its squared rest energy $(mc^2)^2$ due to mass $m$ and squared energy $(pc)^2$ due to momentum $p$:
\begin{align}
 E^2 = (mc^2)^2 + (pc)^2.
 \label{eq:energy-momentum}
\end{align}
The second is the linear Dirac equation~\cite{dirac1928quantum} equating mass with appropriately defined energy and momentum operators:
% \begin{align}
% %% mc^2 = i\hslash c\sum_{\mu = 0}^3\gamma^{\mu}\partial/\partial_{\mu},
% %i\hslash \partial/\partial_{t}
% \gamma^0 E = mc^2 + c\sum_{\mu=1}^3\gamma^{\mu}p_\mu,
%  \label{eq:dirac}
% \end{align}
% where in Equation~\eqref{eq:dirac}, the right hand side is the energy operator $E = i\hslash\partial/\partial_t$ defined by the partial derivative in time $t$ and Planck's reduced constant $\hslash$, and on the left hand side $p_\mu = -i\hslash\partial/\partial_\mu = -i\hslash\nabla_\mu$ is the momentum operator defined by the partial derivative in either of the three spatial directions, and $\alpha_\mu$ are complex-valued 4x4 matrices that ensure anticommutation and consistency between Equations~\eqref{eq:energy-momentum} and ~\eqref{eq:dirac}, implementing the Clifford algebra $\mathbb{C}(1,3)$ of 4D time-space~\cite{griffiths2020introduction}.
\begin{align}
E = mc^2 + pc,
 \label{eq:dirac}
\end{align}
for example in Equation~\eqref{eq:dirac}, the right hand side is the energy operator $E = i\hslash\partial/\partial_t$ defined by the partial derivative in time $t$ and Planck's reduced constant $\hslash$, and on the left hand side $p = -i\hslash\partial/\partial_x = -i\hslash\nabla_x$ is the momentum operator defined by the gradient in spatial direction $x$. By analogy, our model adopts orthogonal linear operators including an odd gradient $\nabla$ and a even sum $\Sigma$, which function analogously to momentum $p$ and rest energy mass $m$.

%where mass and momentum are expressed as a linear basis of gamma matrices (similar to Pauli matrices or quanternions).
%\begin{align}
% E = mc^2 + ipc
%\end{align}
%The energy $E$ momentum $p$ are expressed as operators $E = \partial/\partial_t$ and $\hat{p} = -ih \partial/\partial_x$ while a fixed rest mass is ascribed for standard particles such as the electron or $m=0$ for the massless photon.

In deep networks, the CNN~\cite{fukushima1983neocognitron,lecun1989backpropagation} assumes that linear filter operators are applied identically at all image locations (\ie invariance), and that the maximum speed of information through activation space is a constant limited to half the filter width $\frac{dx}{dt} = \frac{width-1}{2}$ pixels/layer. The CNN differs from relativistic QM in that filter operators and image activations are all real-valued, and in the non-linear ReLU operation which discards all negative-valued activations following linear convolution. Our theory is thus inspired by relativistic QM in adopting a mass-energy-momentum framework as in Equations~\eqref{eq:energy-momentum} and~\eqref{eq:dirac}, however using on real-valued filter operators found in CNNs.

A variety of works investigate physics-informed neural network models~\citep{karniadakis2021physics,raissi2017physics}, including conservation laws and dynamical systems~\citep{muller2023exact,grzeszczuk1998neuroanimator}, e.g. turbulence~\citep{van2024energy}, partial differential equations (PDEs)~\citep{raissi2018deep} and PDE-based architectures\citep{smets2023pde}, and physical models for quantum computing~\cite{bondesan2021hintons}. Our framework makes use of even and odd functions, which are closely linked to theories of mathematical symmetry, group and gauge equivariance~\citep{cohen2016group,cohen2019gauge}. To our knowledge, we are the first to propose and study mechanical modes of information propagation via analogy to relativistic QM in the activation space of generic CNNs.

%In the study of generic CNN filter weights, tools such as Activation Maximization~\citep{simonyan2013deep,Erhan2009Visualizing}, Deconvolution networks~\citep{zeiler2014visualizing}, GradCam~\citep{selvaraju2020grad} allow visualizing inputs maximizing filters or salient regions. The structure of trained filter weights has been studied using non-negative matrix factorization~\cite{petrov2021weight}, PCA~\cite{fukuzaki2022principal}, showing regularity of structure. The symmetry of average filters was shown to rise toward the network output\cite{alsallakh2025symmetries}, and filters have been shown to cluster into small sets of similar gradient-like kernels\cite{babaiee2024unveiling}.
%Binomial filtering has been studied in the context of signal processing~\cite{aubury1996binomial}, e.g. in approximating Gaussian filtering~\cite{wells1986efficient}, however not in the context of non-linearity.

To our knowledge, our work is the first to propose a mechanical description of how filter components lead to the propagation of salient information across 2D image space from one layer to the next.

%There is much interest in modelling phenomena in the physical 3D world space, with quantum mechanics accounting for mass, charge, spin, 

%Interpretability review

%\begin{figure}
%  \centering
%\includegraphics[width=\linewidth]{figs/Previous.png} \\
%\includegraphics[width=\linewidth]{figs/Motivation.png} \\
%  \caption{The three most significant components of linear decompositions resemble DC $\Sigma$ and gradient $\nabla_x, \nabla_y$ filters, including a) JPEG DCT coefficients~\cite{wallace1992jpeg} b) Natural image PCs~\cite{olshausen1996emergence} c) CNN PCs (VGG16)~\cite{fukuzaki2022principal}. The lower graphs show baseline accuracy and reconstruction error vs. number of DCT components, where the circles show that only three $\Sigma,\nabla_x, \nabla_y$ components lead to $> 90\%$ of baseline accuracy and $< 10\%$ reconstruction error for Resnet and VGG16 models.}
%  \label{fig:Frequencies}
%\end{figure}

\section{Elementary Information Mechanics}

We propose to analyze CNN filters in terms of frequency components, primarily low frequency sum $\Sigma$ and gradient $\nabla$ components, which exhibit distinct models of information propagation: diffusion, vibration and translation. We refer to this as {\em elementary information mechanics}, which is particularly relevant for small discrete filter sizes, \eg $3 \times 3$ pixels, where the spectrum is composed of a small number of components. Unlike quantum mechanics describing physical phenomena via complex linear operators, information mechanics here are based on real linear and non-linear operators, \ie convolution and rectification (ReLU).

\subsection{1D Binomial Theory With ReLU}

The primary modes of information propagation under convolution with non-linear ReLU may be characterized in the minimal 1D context with pixel filters of size $2\times 1$, i.e. sum $\Sigma = [1,1]$ and gradient $\nabla = [-1,1]$ operators, and then generalized to higher dimensions via radial symmetry (and antisymmetry). Here, we demonstrate how convolution and ReLU of the sum component results in diffusion and a binomial pyramid, while with the gradient component results in vibration or translation. 

Let $I^{(0)} = \begin{bmatrix} 0 & 1 & 0 \end{bmatrix}$ an impulse in a 1D activation image with a single non-zero pixel $I^{(0)}[0]=1$ at centre position $x=0$ at time or layer $t=0$. CNNs typically apply convolution followed by rectification using the so-called rectification logic unit (ReLU) function defined as:
\begin{align}
  ReLU(I) = \max(0,I) = \frac{I + |I|}{2}.
\end{align}

The activation image $I^{(t)}$ a time $t$ is the result of convolution $*$ with a operator or filter $F$ followed by ReLU:
\begin{align}
  I^{(t)} = ReLU(I^{(t-1)}*F),
\end{align}

Filter $F$ may take on one of three values $F \in \{\Sigma, \nabla_-, \nabla_+\}$, where $\Sigma = [1, 1]$ is the sum operator, and $\nabla_- = [1, -1]$ and $\nabla_+ = [-1,1]$ represent left-handed and right-handed gradient operators (note that $\nabla_-=-\nabla_+$). Inspection reveals that convolving image $I^{(0)} = \begin{bmatrix} 0 & 1 & 0 \end{bmatrix}$ with these operators followed by ReLU leads to three distinct transformations of the original image information, i.e. symmetric diffusion and directional shifting to the left or right:
\begin{align}
 ReLU([0,1,0]*\Sigma~~~) &= [0,1,1,0], &\text{Diffusion} \notag \\
 ReLU([0,1,0]*\nabla_-) &= [0,1,0,0], &\text{Shift Left} \notag \\
 ReLU([0,1,0]*\nabla_+) &= [0,0,1,0], &\text{Shift Right} 
\end{align}

Repeated convolution with the even sum operator $\Sigma = [1,1]$ leads to the well-known binomial pyramid, which may be viewed as random walk where the value of $I^{(t)}(x)$ represents the number of possible paths leading to position $x$. If activations are normalized to sum to unit length $|I^{(t)}|=1$, then the image approaches a Normal distribution with standard deviation $\sigma = \sqrt{t}$, i.e. as $t\rightarrow\infty$, $I^{(t)}\rightarrow Normal(\sqrt{t})$. 
\begin{gather*}
I^{(0)} = \begin{bmatrix} 0 & \textcolor{black}{\text{1}} & 0 \end{bmatrix},\\
I^{(1)} = \begin{bmatrix} 0 & \textcolor{black}{\text{1}} & \textcolor{black}{\text{1}} & 0 \end{bmatrix},\\
I^{(2)} = \begin{bmatrix} 0 & \textcolor{black}{\text{1}} & 2 &\textcolor{black}{\text{1}} & 0 \end{bmatrix},\\
I^{(3)} = \begin{bmatrix} 0 & \textcolor{black}{\text{1}} & 3 & 3 &\textcolor{black}{\text{1}} & 0\end{bmatrix},\\
I^{(4)} = \begin{bmatrix} 0 & \textcolor{black}{\text{1}} & 4 & 6 & 4 &\textcolor{black}{\text{1}} & 0 \end{bmatrix},\\
I^{(5)} = \begin{bmatrix} 0 & \textcolor{black}{\text{1}} & 5 & 10 & 10 & 5 &\textcolor{black}{\text{1}} & 0 \end{bmatrix},\\
I^{(6)} = \begin{bmatrix} 0 & \textcolor{black}{\text{1}} & 6 & 15 & 20 & 15 & 6 &\textcolor{black}{\text{1}} & 0 \end{bmatrix},
\end{gather*}
The diffusion mode results from even convolution, where information propagates identically in all directions. While the leading edge of diffusion propagates at maximum velocity, binomial theorem states that the effective receptive field after $t$ layers approximates a Gaussian distribution with standard deviation $\sigma = \sqrt{t}$, as has been experimentally verified\cite{luo2016understanding}. Diffusion has been studied through the Gaussian scale-space~\cite{lindeberg1994scale}, and in currently popular diffusion generative models~\cite{sohl2015deep}.

Repeated convolution with the odd gradient operators $F \in \{\nabla_-, \nabla_+\}$ causes information to follow unique paths in a random walk, as shown in the following illustration. Red and blue paths result from sequential application of exclusively $\nabla_-$ or $\nabla_+$ operators, where information translates at maximum constant speed in either the left or right directions, respectively. The green path results from alternating between $\nabla_-$ and $\nabla_+$ and vibration mode. Assuming that left and right operators to be equiprobable, the translation paths (red, green) lead to the least probable outcome, and the green path represents the most probable outcome. 

\begin{gather*}
I^{(0)} = \begin{bmatrix} 0 & \textcolor{black}{\text{1}} & 0 \end{bmatrix},\\
I^{(1)} = \begin{bmatrix} 0 & \textcolor{black}{\text{1}} & \textcolor{black}{\text{1}} & 0 \end{bmatrix},\\
I^{(2)} = \begin{bmatrix} 0 & \textcolor{red}{\text{1}} & \textcolor{green}{\text{1}} &\textcolor{blue}{\text{1}} & 0 \end{bmatrix},\\
I^{(3)} = \begin{bmatrix} 0 & \textcolor{red}{\text{1}} & \textcolor{green}{\text{1}} & 0 &\textcolor{blue}{\text{1}} & 0\end{bmatrix},\\
I^{(4)} = \begin{bmatrix} 0 & \textcolor{red}{\text{1}} & 0 & \textcolor{green}{\text{1}} & 0 &\textcolor{blue}{\text{1}} & 0 \end{bmatrix},\\
I^{(5)} = \begin{bmatrix} 0 & \textcolor{red}{\text{1}} & 0 & 0 & \textcolor{green}{\text{1}} & 0 &\textcolor{blue}{\text{1}} & 0 \end{bmatrix},\\
I^{(6)} = \begin{bmatrix} 0 & \textcolor{red}{\text{1}} & 0 & 0 & \textcolor{green}{\text{1}} & 0 & 0 &\textcolor{blue}{\text{1}} & 0 \end{bmatrix},
\end{gather*}

The vibration and translation modes represent a departure from standard linear signal processing, and result from the convolution of an input with odd filters followed by ReLu non-linearity. Vibration follows from convolutions alternating between left and right gradient filters, and produces no net motion of information within the image. Translation follows from convolutions with constant left or right gradient filters, and results in information propagating at the maximum speed, here c=0.5 pixels per layer. Such vibration and translation modes can be expected in wavelet scattering models~\cite{bruna2013invariant}.

%In the following sections, we generalize sum and gradient filtering to 2D image space and larger filter sizes via symmetry and even and odd functions.

\subsection{Symmetry}

Symmetry may be used to generalize the primary diffusion, vibration and translation to 2D and larger filter sizes via symmetry. Specifically, any image or filter $F$ may be decomposed as a sum $F = f_e + f_o$ of orthogonal even (i.e. symmetric) $f_e$ and odd (i.e. antisymmetric) $f_o$ functions with the properties $f_e(-x)=f_e(x)$ and $f_o(-x)=-f_o(x)$. Furthermore, as even and odd components are orthogonal $f_e \perp f_o$, their Euclidean magnitudes form a Pythagorean relationship $\|F\|^2 = \|f_e\|^2 + \|f_o\|^2$. These properties are described by the following quadratic and linear equations:
\begin{align}
 \|F\|^2 &= \|f_e\|^2 + \|f_o\|^2, \label{eq:even_odd_quad}\\
 F &= f_e + f_o.
 \label{eq:even_odd_linear}
\end{align}

\begin{figure}
  \centering
  \begin{tabular}{cc}
     
\includegraphics[width=.45\linewidth]{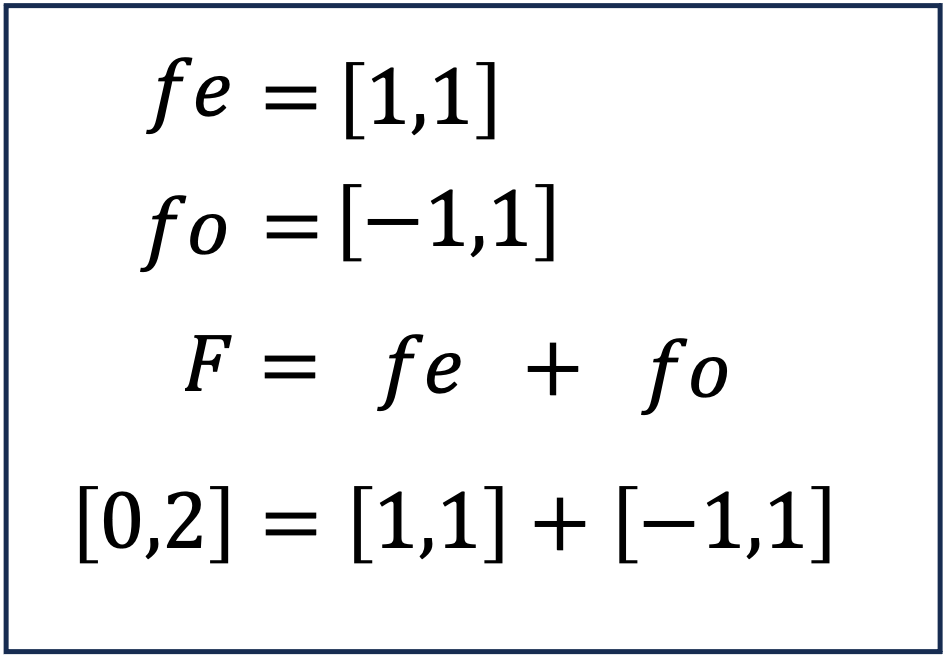}  &  
\includegraphics[width=.45\linewidth]{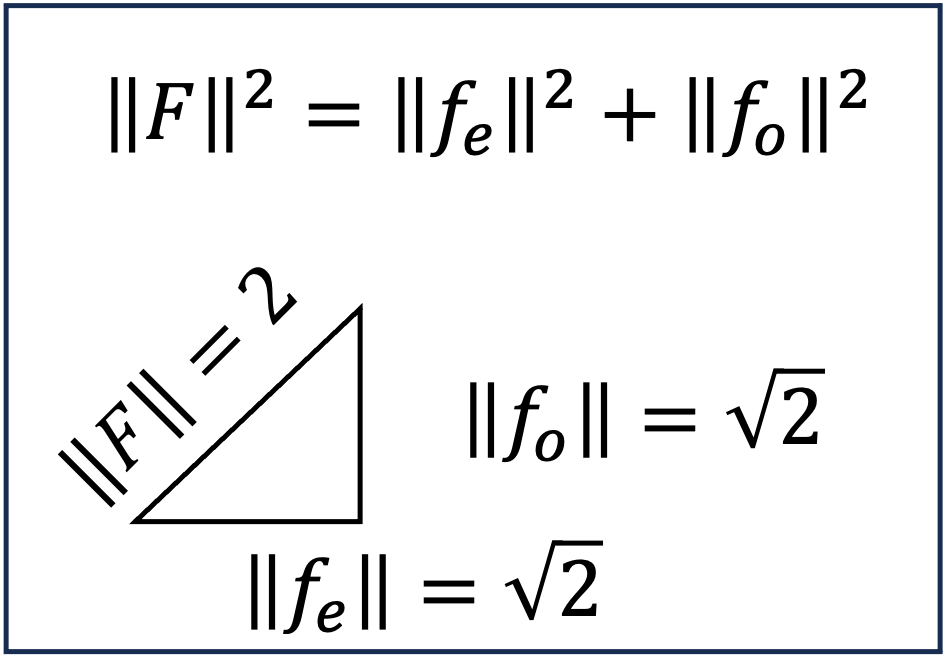} \\
   a)    &  b)
  \end{tabular}
  \caption{a) The decomposition of an example 1D filter $F=[0,2]$ into even $f_e=[1,1]$ and odd $f_o=[-1,1]$ components based on left-right symmetry. b) The Pythagorean relationship between orthogonal even and odd components.}
  \label{fig:geometry}
\end{figure}

\begin{figure}
  \centering
  \includegraphics[width=\linewidth]{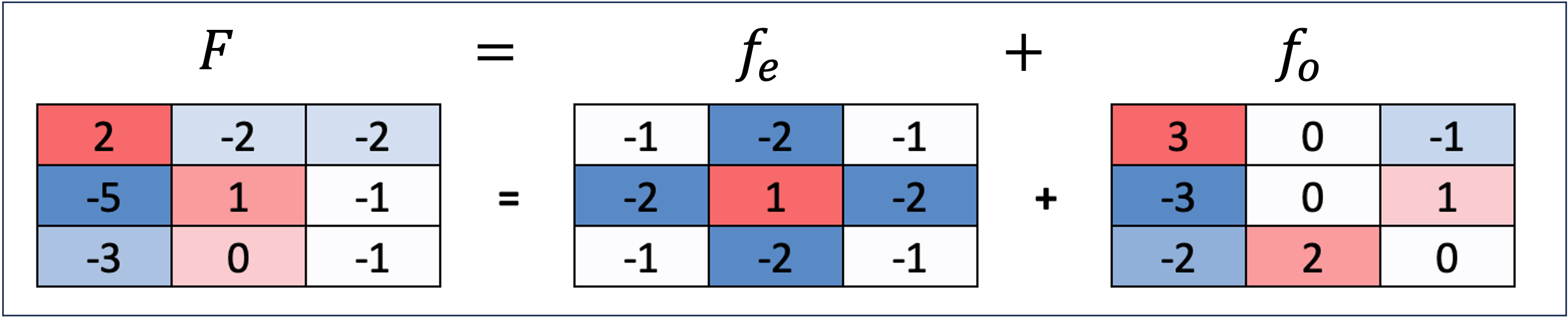} \\
  a) Example 2D filter \\
  \begin{tabular}{cc}
   \includegraphics[width=0.40\linewidth]{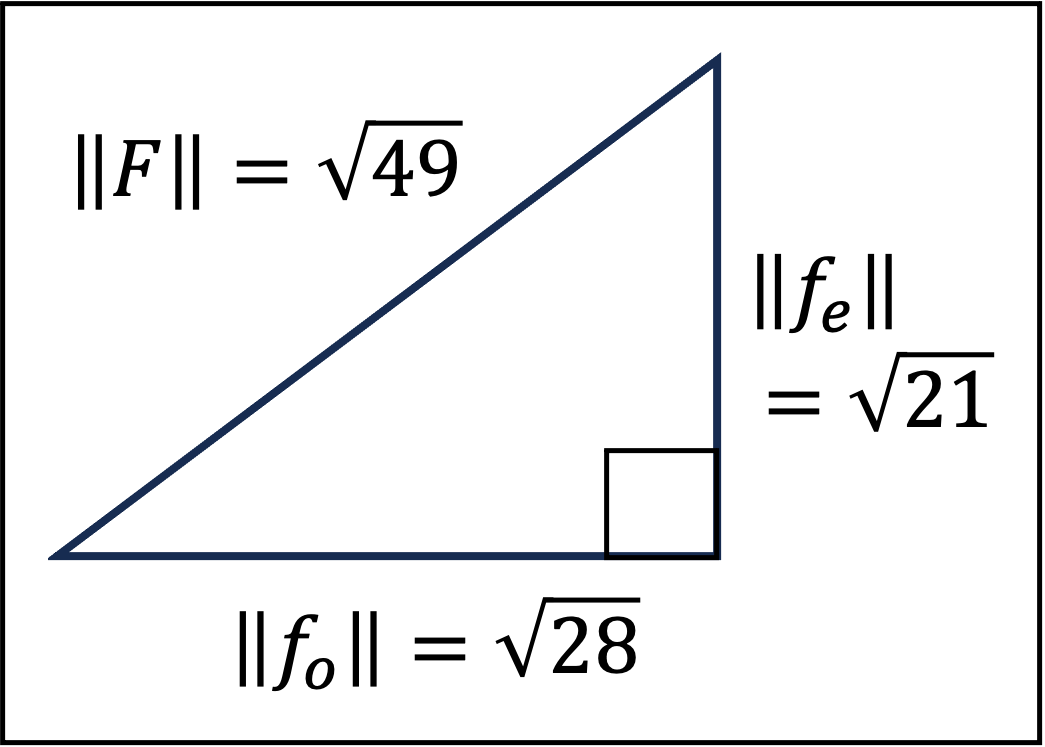}    &  \includegraphics[width=0.52\linewidth]{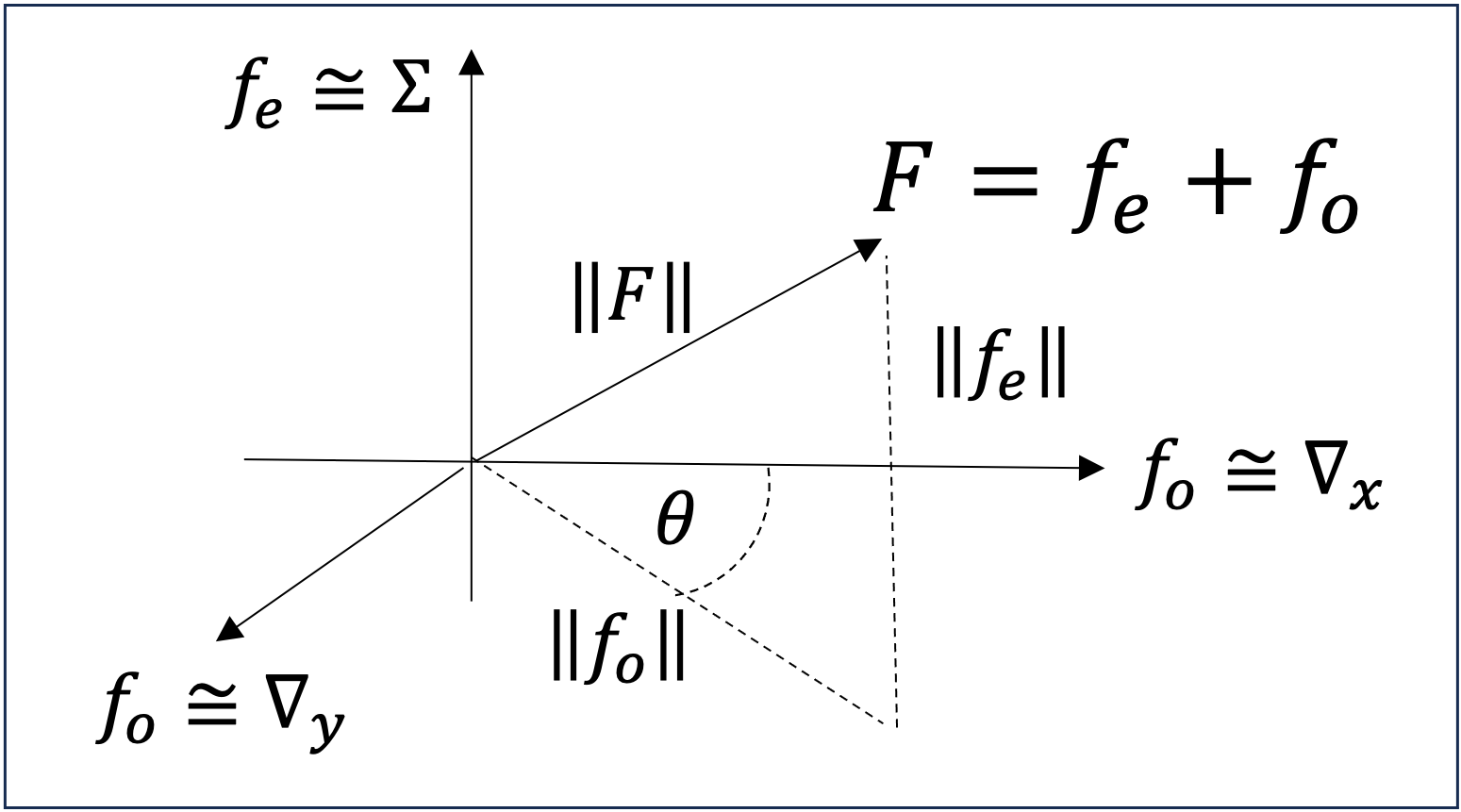} \\
   b) Magnitude Triangle & c) 3D Visualization
  \end{tabular}
  
  \caption{Illustrating the geometry of a 2D filter. a) shows an example decomposition of a 2D filter $F$ into even $f_e$ and odd $f_o$ components. b) shows the Pythagorean relationship between the Euclidean magnitudes of orthogonal $f_e$ and $f_o$ components. c) illustrates 3D filter component space with a vertical even axis $f_e \approx \Sigma$ and a horizontal gradient plane $f_o \approx \{\nabla_x, \nabla_y\}$.}
  \label{fig:geom_2D}
\end{figure}

Figures~\ref{fig:geometry} a) and b) illustrate an example of decomposition into primary even and odd components, which for a minimal 1x2 filter are sum $\Sigma = [1,1]$ and gradient $\nabla = [-1,1]$ filters. In generalizing symmetry from 1D to 2D, filter orientation is no longer a binary left-right variable but a continuous angle $\theta$ in the 2D image plane. Figure~\ref{fig:geom_2D} a) provides an example of a 2D filter decomposed according to even $f_e$ and odd $f_o$ components. Figure~\ref{fig:geom_2D} b) illustrates the magnitude triangle formed by orthogonal even $f_e$ and odd $f_o$ components. Figure~\ref{fig:geom_2D} c) shows how a single filter channel may be visualized in a 3D space according to dominant low frequency components, with an even vertical axis defined by the sum $f_e \approx \Sigma$ component, and an odd horizontal plane defined by gradient $f_o \approx \{\nabla_x,\nabla_y\}$.

%Figure~\ref{fig:geom_2D} d) shows examples of multichannel filters (from VGG16) dominated by either 1) even or 2) odd components. Note how filters are generally mean zero, indicating conservation of 1) sign in the case of even sum $\Sigma$ filters and of 2) momentum in the case of odd gradient $\nabla$ orientation with noticeable bidirectionality in the horizontal image plane, indicating dominant vibration (as opposed to translation) mode of propagation.

With regard to frequency components, the sum component is an even or symmetric function $\Sigma \in f_e$ and the gradient components are odd or antisymmetric functions $\nabla_x, \nabla_y \in f_o$. Higher order DCT components may generally be either even, odd or mixed, but since they contribute only weakly to classification, we leave their analysis to future work and assume $f_e \propto \Sigma$ and $f_o \propto \cos{\theta}\nabla_x + \sin{\theta}\nabla_y$ where $\theta$ is the gradient orientation. Finally for a given filter \( F \), we define \( \beta^2 \) as a mixing ratio between even and odd components, defined by the ratio of odd component contribution \( \| f_o \|^2 \) to the total energy \( \| F \|^2 \) as follows:
\begin{align}
    \beta^2 = \frac{\| f_o \|^2}{\| f_o \|^2 + \| f_e \|^2}.\label{eq:beta_def}
\end{align}
So for example, a filter $F$ may be defined by mixing arbitrary odd and even components according to ratio $\beta$ as follows:
\begin{align}
    F = \|F\|\beta \hat{f_o} + \|F\|\sqrt{1-\beta^2}\hat{f_e},
    \label{eq:beta_mixing}
\end{align}
where in Equation~\eqref{eq:beta_mixing}, $\hat{f_e}=\frac{f_e}{\|f_e\|}$ and $\hat{f_o}=\frac{f_o}{\|f_o\|}$ represent even and odd components normalized to unit length. 

\subsection{The Mechanics of Information Propagation} \label{ssec:mec_prop}

Rectified convolution has distinctly different mechanical effects upon activation information for and odd filter components, which are not present in convolution without rectification. Even filters will cause information to propagate isotropically and do not alter the activation map centre of mass, whereas odd filters will generally cause information to propagate directionally according to angle $\theta$ and thus shift the centre of mass.

The shift in centre of mass may be expressed in terms of the expected value of the filter impulse response. For an even filter,
\begin{align}
E_{f_e}[x,y] = \frac{\sum_{x,y}[x,y]f_e(x,y)}{\sum_{x,y}f_e(x,y)} = [0,0],
\end{align}
and thus by linear superposition, filtering with an even component $f_e$ results in no net displacement of the centre of mass, with or without rectification.

For an odd filter $f_o$, the expected value of the impulse response following rectification equals the centre of mass of the positively signed weights $f_o^+ = \max\{f_o,0\}$
\begin{align}
    E_{f_o}[x,y] = \frac{\sum_{x,y}[x,y]ReLU[f_o(x,y)]}{\sum_{x,y}ReLU[f_o(x,y)]}, \notag \\
    = \frac{\sum_{x,y}[x,y]f_o^+(x,y)}{\sum_{x,y}f_o^+(x,y)} \ne [0,0],
    \label{eq:expected_displacement_odd}
\end{align} 
and thus filtering with an odd component $f_o$ generally leads to a net displacement of the centre of mass.

\begin{figure}[htbp]
  \centering
  \begin{subfigure}{\columnwidth}
    \includegraphics[width=\linewidth]{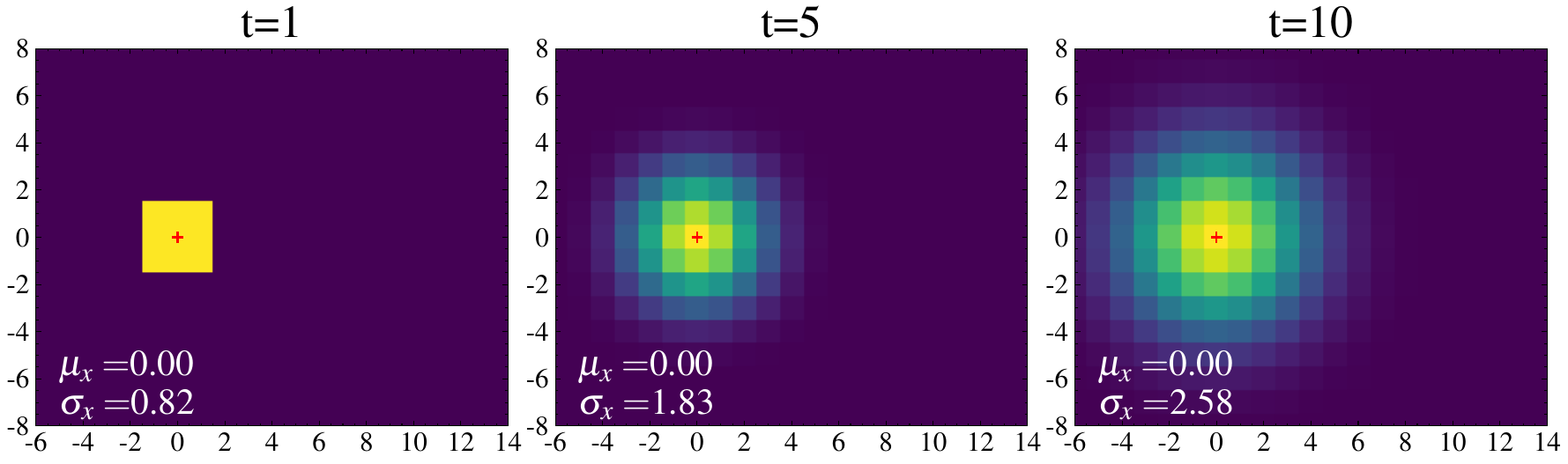}
    \caption{$\Sigma$ $(\beta^2=0)$}
    \label{fig:prop_sym}
  \end{subfigure}\\
     \begin{subfigure}{\columnwidth}
    \includegraphics[width=\linewidth]{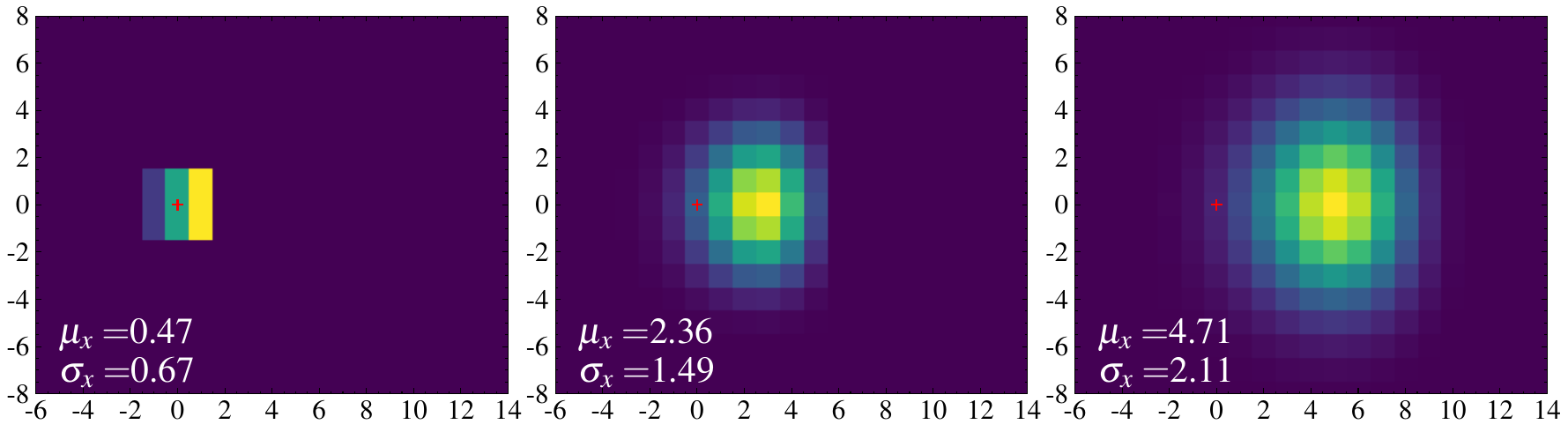}
    \caption{$\Sigma + \nabla_x$  $(\beta^2=0.25)$}
    \label{fig:prop_mix}
  \end{subfigure} \\
  \begin{subfigure}{\columnwidth}
    \includegraphics[width=\linewidth]{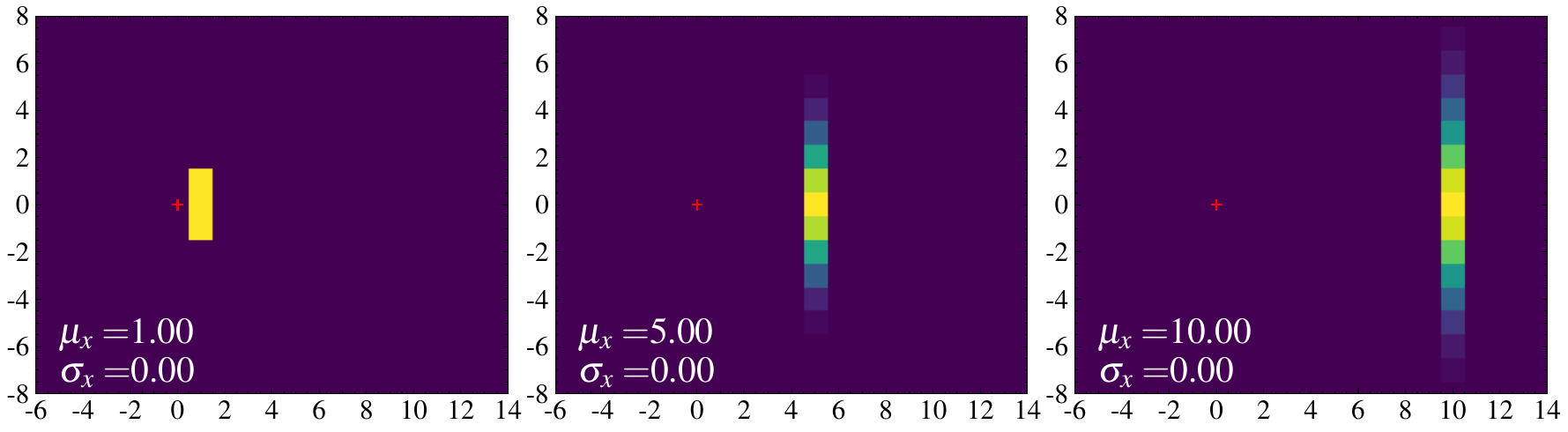}
    \caption{$\nabla_x$ $(\beta^2=1)$}
    \label{fig:prop_anti}
  \end{subfigure}
  
  \caption{The effect of repeated convolution+ReLU of an impulse test pattern with a $3\times3$ kernel mixing DC $\Sigma$ and gradient $\nabla_x$ components at three different ratios $\beta^2=\{0,.25,1\}$.}
  \label{fig:prop_diag_3x3_relu}
\end{figure}

\begin{figure}[htbp]
  \centering
  \begin{subfigure}{\columnwidth}
    \includegraphics[width=\linewidth]{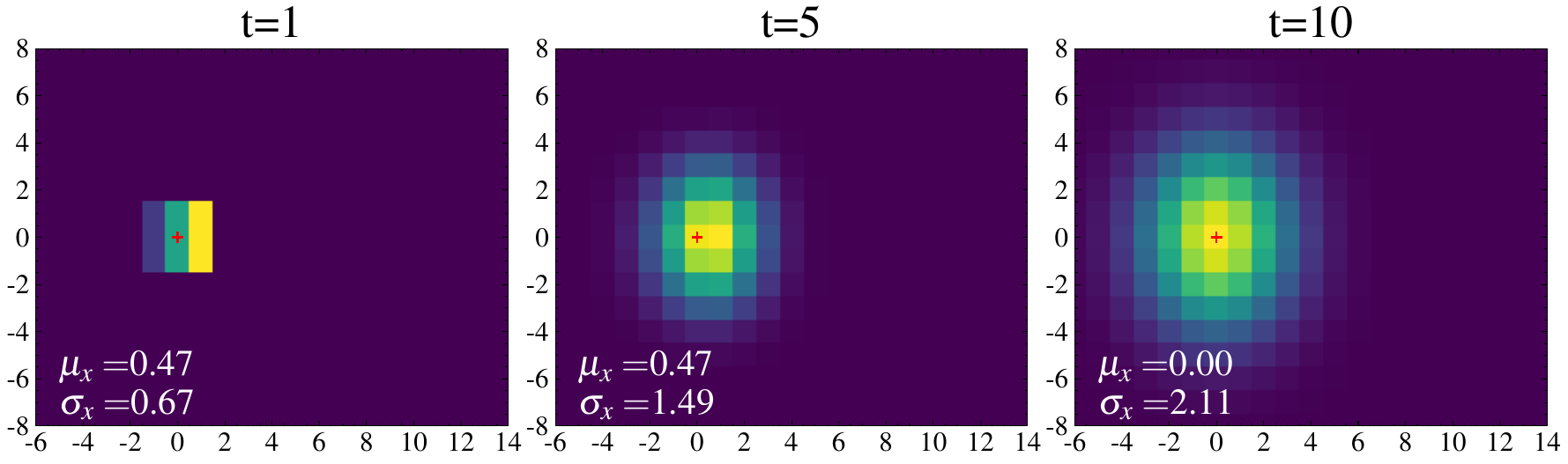}
        \caption{$\Sigma \pm \nabla_x$ $(\beta^2=0.25)$}
    \label{fig:prop_mix_bipol}
  \end{subfigure}\\
     \begin{subfigure}{\columnwidth}
    \includegraphics[width=\linewidth]{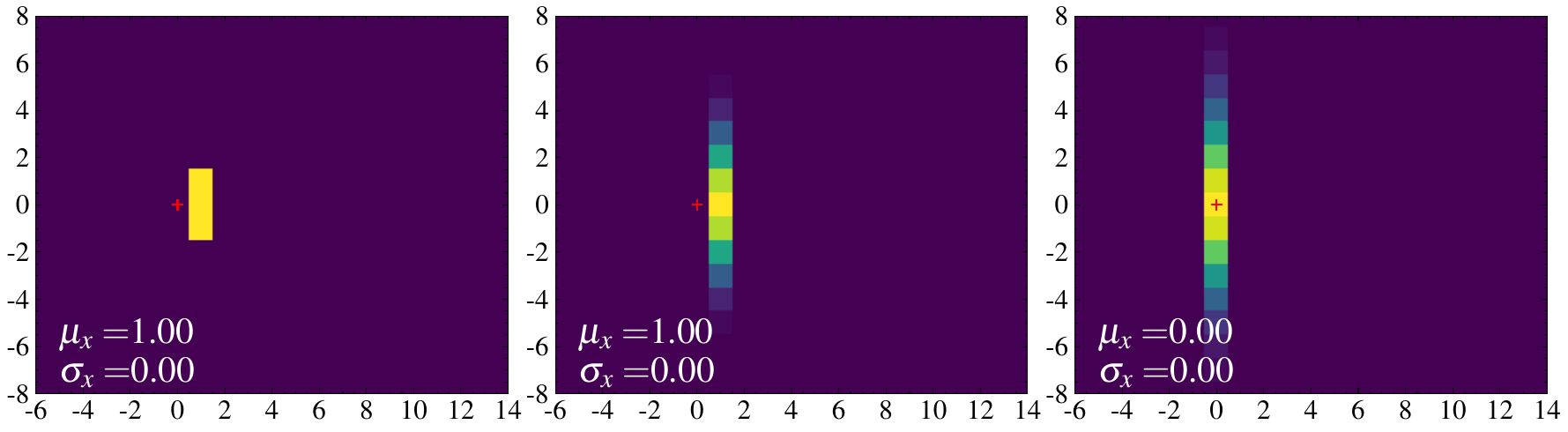}
    \caption{$\pm \nabla_x$ $(\beta^2=1)$}
  \label{fig:prop_anti_bipol}
  \end{subfigure} \\
  \caption{The effect of repeated convolution+ReLU of an impulse test pattern with a $3\times3$ kernel mixing DC $\Sigma$ and alternating gradient $\pm\nabla_x$ components at mixing ratios $\beta^2=\{.25,1\}$.}
  \label{fig:prop_bipol}
\end{figure}

\Cref{fig:prop_diag_3x3_relu} demonstrates these modes of propagation in the 2D image plane, following sequential convolution of a test pattern (single pixel, impulse), where sum $\Sigma$ and horizontal gradient $\nabla_x$ components are combined according to various mixing ratios $\beta^2$ (\Cref{eq:beta_mixing}). Pure diffusion (\cref{fig:prop_sym}) results from $\beta^2=0$. 
Vibration (\Cref{fig:prop_bipol}) is caused by sequential convolution at alternating orientations $\pm\nabla$ . 

There is no net motion for either diffusion  (\cref{fig:prop_sym}) or vibration  (\cref{fig:prop_mix_bipol,fig:prop_anti_bipol}). The test pattern horizontal size (standard deviation $\sigma$) increases with diffusion and decreases with increased $\beta^2$, and the bulk of the test pattern disappears for $\beta^2 > 0.5$. Pure vibration (\cref{fig:prop_anti_bipol}) with $\beta^2=1$ captures a single test pattern edge. Pure gradient (\cref{fig:prop_anti}) results in a test pattern edge propagating rightwards at maximum velocity in a manner reminiscent of a propagating electromagnetic wavefront.

% \begin{figure}
%   \centering
%   \begin{tabular}{cc}
%     \includegraphics[width=.33\linewidth]{figs/2d_prop_test_pattern.png}   &  \includegraphics[width=.33\linewidth]{figs/2d_prop_beta_ratio.png}\\
%      a)Test Pattern  & b) Beta Ratio\\
%   \end{tabular}
% \includegraphics[width=\linewidth]{figs/2d_prop_three_modes.png} \\
% c) Propagation (80 layers)
%   \caption{Demonstrating the three modes of propagation by sequentially filtering a circular test pattern (a) with filter mixing sum $\Sigma$ and horizontal gradient $\nabla_x$ components according to a range of $\beta^2 \in [0,1]$ (b). $\beta^2 = 0$ corresponds to pure diffusion (1), $\beta^2 = 1$ to pure vibration (2) or translation (3). Vibration results from alternating right-left gradients $\nabla_x$, $-\nabla_x$, and translation from unidirectional right $\nabla_x$. $\sigma$ represents the image pattern variance in the horizontal direction.}
%   \label{fig:prop2D}
% \end{figure}

\section{Relativistic Energy-Momentum Relation}

The mechanical effects of even and odd filters under rectified convolution may be thought of as {\it rest energy} or {\it momentum energy} in the sense that the former leaves the centre unchanged while the latter causes a shift. This is not the case for standard convolution, and it is the prime novelty of this work. We are thus able to draw a novel analogy between filtering components and physical operators in a relativistic energy-momentum formulation as follows.

Even-odd Equations~\eqref{eq:even_odd_quad} and~\eqref{eq:even_odd_linear} are analogous to the energy-momentum Equation~\eqref{eq:energy-momentum} and the Dirac Equation~\eqref{eq:dirac}, respectively. Specifically, assuming natural units for constants $\hslash = c = 1$, similarities are the squared filter magnitude and energy $\|F\|^2 \sim E^2$, the gradient components and the momentum $(\nabla_x, \nabla_y) \sim (p_x, p_y)$, and the sum component and mass $\Sigma \sim m$. Note that quantum particle mass is typically treated as a constant (\eg the electron mass = 9.1e-31 kg), however in our model, the sum component $\Sigma$ may be viewed as a rest mass-like operator, integrating activations within a neighbourhood while leaving the centre of mass unchanged.

Most importantly, we are able to establish the equivalence of our even-odd decomposition and the Lorentz transform of special relativity, as follows. The rest energy of an unmoving massive particle is $E=mc^2$ where $m$ is the rest mass and $c=\frac{dx_{max}}{dt}$ is the speed of light (i.e. the maximum speed possible). The total energy of a particle moving at velocity $v = \frac{dx}{dt}$ is quantified using the velocity ratio $\beta = \frac{v}{c} \in [0,1]$ as follows. First, define the so-called Lorentz gamma factor $\gamma$, which may be expanded as a Taylor series:
\begin{align}
\gamma &= \frac{1}{\sqrt{1-\beta^2}} = 1 + \frac{1}{2}\beta^2 + \frac{3}{8}\beta^4 + \frac{5}{16}\beta^6 + \dots  
\label{eq:lorentz_gamma}
\end{align}
Then the particle energy may be defined as
\begin{align}
E = \gamma mc^2 = mc^2 + (\gamma -1)mc^2 = mc^2 + pc,
\label{eq:relativistic_energy}
\end{align}
where $mc^2$ is the rest energy and $pc$ is the energy due to momentum $p$, respectively. For example, the total energy of a slowly moving mass (i.e. $\beta << 1$) leads to the following expression:
\begin{align}
E = \gamma mc^2 \approx \left(1 + \frac{1}{2}\beta^2\right)mc^2 =  mc^2 + \frac{1}{2}mv^2,
\end{align}
where the energy due to momentum $pc = \frac{1}{2}mv^2$ may be approximated by the familiar Newtonian expression of a mass $m$ moving at velocity $v=\frac{dx}{dt}$.

We note that substituting our expression for $\beta^2$ from Equation~\eqref{eq:beta_def} into the definition of Lorentz gamma factor $\gamma$ from Equation~\eqref{eq:lorentz_gamma} leads to the following expression:
\begin{align}
\gamma^2 &= \frac{1}{1 - \beta^2} = \frac{1}{1 - \frac{\|f_o\|^2}{\|f_e\|^2+\|f_o\|^2}} = 1 + \frac{\|f_o\|^2}{\|f_e\|^2},
  \label{eq:gamma_beta}
\end{align}
while using the definition of relativistic energy in Equation~\eqref{eq:relativistic_energy}
\begin{align}
\gamma^2 &= 1 + \frac{p^2}{(mc)^2}.
\label{eq:gamma_energy}
\end{align}
From Equations~\eqref{eq:gamma_beta} and~\eqref{eq:gamma_energy}, we thus establish equivalence of the following ratios. First, the speed ratio $\beta$
\begin{align}
\frac{v}{c} = \frac{\|f_o\|}{\|f\|} = \beta,
\label{eq:ratio_speed}
\end{align}
where in Equation~\eqref{eq:ratio_speed} states that the speed of displacement $v$ relative to the maximum speed $c$ is equivalent to the ratio of odd filter energy $\|f_o\|$ to total filter energy $\|f\| = \sqrt{\|f_o\|^2+\|f_e\|^2}$.

Second, the energy ratio 
\begin{align}
\frac{pc}{mc^2} = \frac{\|f_o\|}{\|f_e\|},
\label{eq:ratio_energy}
\end{align}
where Equation~\eqref{eq:ratio_energy} states that the ratio of energy due to momentum $pc$ vs rest energy $mc^2$ is equivalent to the ratio of odd filter energy $\|f_o\|$ to even filter energy $\|f_e\|$.

\begin{figure}
  \centering
 \begin{subfigure}{0.49\linewidth}
 \includegraphics[width=\linewidth]{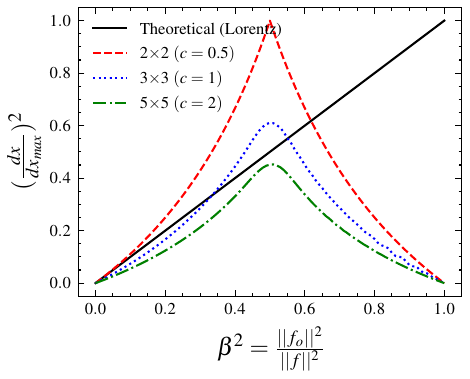} 
\caption{No ReLU}
\end{subfigure} 
\begin{subfigure}{0.49\linewidth}
    \includegraphics[width=\linewidth]{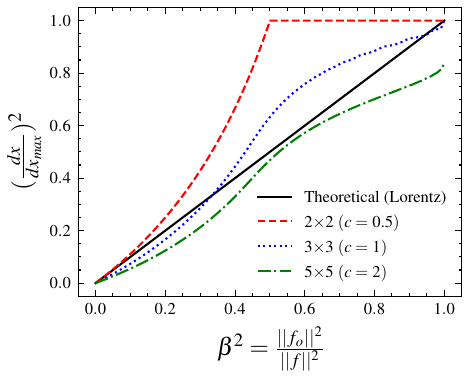} 
    \caption{ReLU}
\end{subfigure}
  \caption{Comparing the squared distance travelled $(dx/dx_{max})^2$ by the information centre of mass vs the squared mixing ratio $\beta^2$, the distance is normalized according to the maximum distance $dx_{max}$ determined by the filter size.}
  \label{fig:propagation_graphs}
\end{figure}

These two ratios allow us to quantitatively determine, from odd and even filter components, the speed at which information propagates directionally vs diffusively under rectified convolution, \eg in CNNs, via a direct analogy to relativistic physics, which, to our knowledge, is novel. Figure~\ref{fig:propagation_graphs} (b) shows that over various filter sizes and $\beta$ mixing ratios, $3\times3$ kernels closely approximate the theoretical Lorentz transform in terms of displacement, rising steadily and monotonically from $\beta=0$ to $\beta=1$. Note $2\times2$ kernels result in unusual behaviour, and $5 \times 5$ kernels do not reach maximum speed as the expected displacement of a $5\times5$ gradient kernel according to Equation~\eqref{eq:expected_displacement_odd} is less than the maximum. In contrast, Figure~\ref{fig:propagation_graphs} (a) shows that for convolution without rectification, displacement reaches a peak at $\beta^2=0.5$ then falls to zero for $\beta=1$, and does not follow a Lorentzian relationship.

%Purely even filtering with $\beta = 0$ the leads to zero net displacement $dx = 0$, purely odd filtering with $\beta = 1$ to a maximum displacement $dx = dx_{max} = \frac{Width-1}{2}$ as determined by the filter  $Width$, and mixed filtering $0 < \beta < 1$ to intermediate displacements $0 < dx < dx_{max}$. 

%For this reason, the odd gradient component $f_o = \nabla = [-1, 1]$ may be considered to represent kinetic energy, and indeed in quantum mechanics it defines the momentum operator $\hat{p} = -i\hslash\nabla$. Similarly, the even sum component $f_e = \Sigma = [1,1]$ may be viewed as representing potential or rest energy by integrating activations to a mass-like value with zero net displacement.

%\begin{align}
%\beta^2 &= \frac{\|f_a\|^2}{\|f_s\|^2+\|f_a\|^2}, \\
%1 - \beta^2 &= 1 - \frac{\|f_a\|^2}{\|f_s\|^2+\|f_a\|^2} = \frac{\|f_s\|^2}{\|f_s\|^2+\|f_a\|^2} 
%\end{align}

%Our work defines $\beta$ as the ratio of antisymmetric energy $\|f_a\|$ to the total energy $f = \sqrt{\|f_a\|^2 + \|f_s\|^2}$. Then $\beta^2$ may be expressed as

%To validate 
%We study the propagation of information we compare the impact of $\beta$ upon the convolution of an impulse response, where $\beta$ mixes between a sum $(\beta)\Sigma$ and $(1-\beta)\nabla$ both with and without ReLU. We measure the mean

%Without rectification a), the distance travelled $dx^2$ attains a maximum at $\beta^2=0.5$ then drops to 0. With ReLU b) $dx$ rises monotonically and attains a higher value.

 \section{DCT Training Experiment}\label{dct_experiments}

Convolutional kernels may be expressed in terms of the DCT frequency decomposition, and approximated by small sets of the lowest low-frequency components, which are dominant in natural images \citep{ruderman1993statistics}. Here, we evaluate CNN training with various subsets of DCT components to assess their impact on classification accuracy. We hypothesize that the majority of accuracy may be achieved by small subsets of low-frequency components. We find that a 3-component representation including the even DC component $\Sigma$ and the odd first-order oriented gradients ($\nabla_x$, $\nabla_y$) leads to an effective approximation capturing $> 92\%$ of classification accuracy.

We evaluate how individual frequency components contribute to network behaviour.  In a standard convolutional layer, each input channel $C_i$ is convolved with a distinct $k \times k$ kernel $f_i$, and the resulting feature maps are summed to produce the filter output $Y$ (see \Cref{eq:basic_conv_layer}). Because any kernel or image can be represented as a linear combination of DCT bases, we introduce a modified convolutional layer that incorporates this decomposition. During forward propagation, each kernel is generated as a weighted sum of $N$ DCT bases, where the weights (DCT coefficients)  $\{ \omega_i \}$ are learned parameters (see \Cref{eq:dct_conv_layer}).
%Our proposed experimental setup is illustrated in \Cref{fig:dct_experiment_setup}. 
\begin{align}
    Y &= \sum_{j=1}^{C} f_j \ * \ C_j \label{eq:basic_conv_layer} \\
    &= \sum_{j=1}^{C} \left ( \sum_{i=1}^{N} \omega_i \  DCT_i \right) \ * \ C_j \label{eq:dct_conv_layer}
\end{align}

When using the full spectrum ($N = k^2$), the layer behaves identically to a standard convolution. Since kernel generation (\Cref{eq:dct_conv_layer}) is a linear operation, backpropagation remains unchanged, meaning any convolutional layer in a trained CNN can be replaced by its DCT-based counterpart without altering the models behaviour. By varying $N$, we analyze how each frequency band contributes to network performance.

%\begin{figure*}[htb]
%\centering
%
%		\includegraphics[width=0.7\linewidth]{figs/dct_experiment.pdf} 
%\caption{Illustrating a DCT convolution layer, where each kernel is a linear sum of $N$ DCT basis components scaled by a learnable weighting parameter $\omega$. With the exception of the kernel generation step, the overall mechanism is identical to a standard convolution layer}
%\label{fig:dct_experiment_setup}
%\end{figure*}

%We consistently find that most of the representation in trained CNNs is encoded by the lowest three frequency components, corresponding to the DC ($\Sigma$) and first-order gradient bases ($\nabla_x, \nabla_y$). 

%Our DCT experiments follow a similar method to~\cite{chkeinski2020dct}. 

%\subsection{Spectra of ImageNet-trained Kernels}

This experiment seeks to evaluate the contribution of DCT frequency component subsets toward the task of classification on the Imagenet dataset (ILSVRC2012)~\cite{imagenet}. We evaluate VGG16~\citep{simonyan2014very} and ResNet50~\cite{he2016deep} models by progressively reducing the number of preserved DCT coefficients $\omega$ representing each $3\times3$ convolutional kernel, down to a single DC component.

We initialize the DCT weights $\omega$ by projecting pretrained $3\times3$ kernels from VGG16~\citep{simonyan2014very} and ResNet50~\citep{he2016deep}. Figure~\ref{fig:dct-decomp-full} shows the average spectral energy across all layers. In both models, the majority of energy lies in the three lowest frequency components ($\Sigma$, $\nabla_x$, $\nabla_y$), confirming that learned filters rely on DC and gradient structure.

\begin{figure}
  \centering

\begin{subfigure}{0.49\linewidth}
    \includegraphics[width=\linewidth]{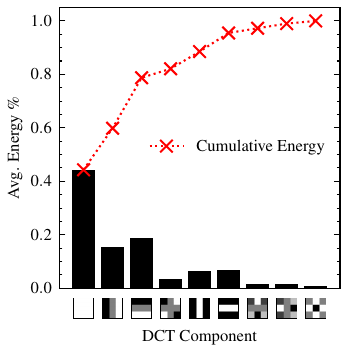}
    \caption{VGG16}
    \label{fig:dct-full-vgg}
  \end{subfigure} 
  %\hspace*{0.0000\linewidth}
  \begin{subfigure}{0.49\linewidth}
    \includegraphics[width=\linewidth]{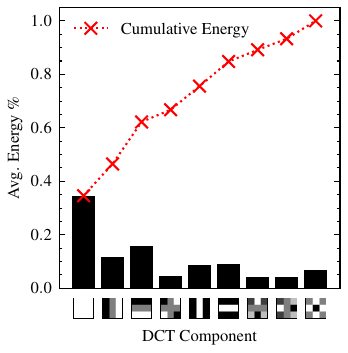}
    \caption{Resnet50}
    \label{fig:dct-full-resnet}
  \end{subfigure}
  
\caption{Spectral DCT decomposition $\omega_i$ of all $3\times3$ convolutional filters in all layers of \subref{fig:dct-full-resnet}~Resnet50 and \subref{fig:dct-full-vgg}~VGG16. We find that in both models the majority of the weights are comprised of low order DC and Gradients ($\Sigma+\nabla$)  } 
  \label{fig:dct-decomp-full}
\end{figure} 

\begin{figure}
  \centering
    \begin{subfigure}{0.49\linewidth}
    \includegraphics[width=\linewidth]{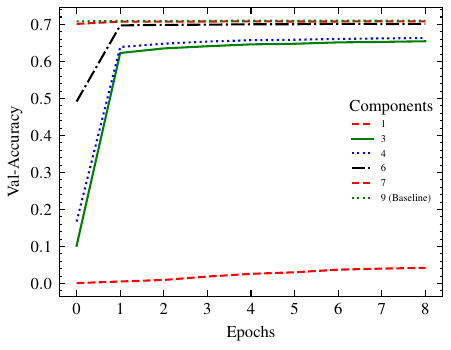}
    \caption{VGG16}
    \label{fig:dct-train-vgg}
  \end{subfigure} 
%  \hspace*{0.0000\linewidth}
    \begin{subfigure}{0.49\linewidth}
    \includegraphics[width=\linewidth]{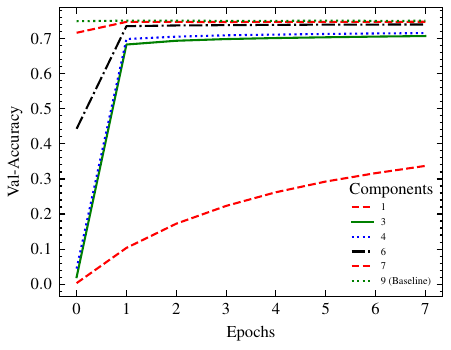}
    \caption{Resnet50}
    \label{fig:dct-train-resnet}
  \end{subfigure}
  \caption{Preserving $N$ DCT components of learned $3\times3$ weights (Trained on ImageNet) and retraining only those components on ImageNet. Note that 3 components ($\Sigma+\nabla$) account for approximately 93\% of the baseline representation }
  \label{fig:dct-train}
\end{figure}

We then evaluate accuracy when only the lowest $N$ DCT components are preserved. At initialization (Epoch 0), truncation significantly reduces accuracy. However, minimal fine-tuning on the preserved coefficients is sufficient to recover performance close to the original models. As shown in Figure~\ref{fig:dct-train}, both VGG16 and ResNet50 achieve over 92\% of baseline accuracy using only the three lowest components, despite these accounting for only 79\% and 62\% of each models spectral energy, respectively. \\

Both models were fine-tuned for 8 epochs with a learning rate $\eta=10^{-5}$ and batch size 256. Following the protocol in \cite{he2016deep}, during training, we randomly resize the shorter side of each image between $[256, 480]$, then randomly crop a $224\times224$ patch and apply random horizontal flipping. For validation, we resize the shorter side to 256 and centre crop a $224\times224$ patch. Initial weights were taken from the pretrained models available in Keras~\cite{chollet2015keras}.
%Code will be provided upon acceptance.

\section{Discussion} 

In this work, we propose a description of information propagation to describe the action of filtering operators on information in CNN activation maps. The theory considers an even-odd filter decomposition, which for small $3\times3$ filters is dominated by three fundamental components, the sum $\Sigma$ and gradients $(\nabla_x, \nabla_y)$. These are shown to result in three distinct modes of information propagation: 1) symmetric diffusion associated with the sum $\Sigma$ and 2) oriented vibration and/or translation associated with gradient operators $(\nabla_x, \nabla_y)$ applied bi-directionally or uni-directionally.

Odd and even kernels can be combined using a mixing ratio $\beta$, which is linearly related to the resulting propagation speed when convolutions are applied with ReLU. Notably, we find that ReLU leads to behaviour closely following Lorentz transform of relativistic physics and enables maximal displacement of information.

%It is worth noting an in-plane angle $\theta = \arctan(\nabla_y / \nabla_x)$ that is defined by odd quadrature pair components. Sequential convolutions applied at differing orientations $\theta$ introduce attenuation in the resulting filter responses, analogous to Malus’ law in polarized light filtering. An analysis of this effect lies beyond the scope of this paper and is left for future work.

%The functioning of elementary information mechanics is presented through intuitive demonstrations 1D and 2D space and even-odd function theory. Filter kernels may be visualized in a 3D space defined by a vertical sum $\Sigma$ axis and a horizontal odd gradient plane. Results reveal a number of novel and surprising insights. Elementary sum and gradient components alone account for over 90\% of accuracy in well-known context VGG16 and Resnet50 models and ImageNet classification task.  The weights of trained network filters account explicitly for angular attenuation reminiscent of Malus' law. Filters in early layers tend to be dominated by bi-polar gradient filters and thus the vibration mode, filters in deep layers by sum filters and thus the diffusion mode.

We name this model elementary information mechanics as it considers the fundamental sum and gradient operators, which are present in filters of all sizes, in various frequency decompositions including the DCT, PCA and Haar wavelets, in popular hand-crafted filtering processes including Gaussian diffusion scale-space~\cite{lindeberg1994scale} and the scattering transform~\cite{bruna2013invariant}, and here demonstrated investigated for the first time trained filters of generic CNNs.

Future work will investigate the use of elementary information mechanics in various aspects of deep neural networks \eg optimization, initialization, domain adaptation. Higher order frequency components may be used to represent other mechanical aspects such as angular momentum, these become more important in larger filters. The theory could be applied to understand the mechanics and optimize models other than CNNs, such as the transformer~\cite{pan2024dct} or generative diffusion \cite{ning2024dctdiff} where the use of DCT components has shown promise.

Finally, our theory of elementary information mechanics is inspired by an analogy to relativistic quantum mechanics, specifically the quadratic energy-momentum equation and the linear Dirac equation accounting for the mass, energy and momentum of a relativistic particle. Specifically, even sum $\Sigma$ and odd gradient $\nabla$ operators are analogous to mass $m$ and momentum $p$ components in the energy-momentum relation. As elementary information mechanics is defined in terms of standard real-valued filters operating on positive-valued CNN activations, it may prove useful understanding the link between the propagation of energy and mass in real physical world space and the propagation of information within activation image space of deep networks.  To our knowledge, this is the first work demonstrating the link between information processing in generic CNNs and the energy-momentum relation, a cornerstone of modern relativistic physics.

Code for generating the results in this paper may be found at~\footnote{\url{https://github.com/liamaltarac/Information-Mechanics}}.

{
    \small
    \bibliographystyle{ieeenat_fullname}
    \bibliography{main}

@String(ICCV= {Int. Conf. Comput. Vis.})

@String(ICCV  = {ICCV})

@article{einstein1905electrodynamics,
  title={On the electrodynamics of moving bodies},
  author={Einstein, Albert and others},
  journal={Annalen der physik},
  volume={17},
  number={10},
  pages={891--921},
  year={1905}
}

@article{einstein1905does,
  title={Does the inertia of a body depend upon its energy-content},
  author={Einstein, Albert},
  journal={Annalen der physik},
  volume={18},
  number={13},
  pages={639--641},
  year={1905}
}

@article{dirac1928quantum,
  title={The quantum theory of the electron},
  author={Dirac, Paul Adrien Maurice},
  journal={Proceedings of the Royal Society of London. Series A, Containing Papers of a Mathematical and Physical Character},
  volume={117},
  number={778},
  pages={610--624},
  year={1928},
  publisher={The Royal Society London}
}

@article{fukushima1983neocognitron,
  author={Fukushima, Kunihiko and Miyake, Sei and Ito, Takayuki},

  journal={IEEE Transactions on Systems, Man, and Cybernetics}, 

  title={Neocognitron: A neural network model for a mechanism of visual pattern recognition}, 

  year={1983},

  volume={SMC-13},

  number={5},

  pages={826-834},

  doi={10.1109/TSMC.1983.6313076}}

@article{lecun1989backpropagation,
  title={Backpropagation applied to handwritten zip code recognition},
  author={LeCun, Yann and Boser, Bernhard and Denker, John S and Henderson, Donnie and Howard, Richard E and Hubbard, Wayne and Jackel, Lawrence D},
  journal={Neural computation},
  volume={1},
  number={4},
  pages={541--551},
  year={1989},
  publisher={MIT Press}
}

@article{wallace1992jpeg,
  title={The JPEG still picture compression standard},
  author={Wallace, Gregory K},
  journal={IEEE transactions on consumer electronics},
  volume={38},
  number={1},
  pages={xviii--xxxiv},
  year={1992},
  publisher={IEEE}
}

@article{olshausen1996emergence,
  title={Emergence of simple-cell receptive field properties by learning a sparse code for natural images},
  author={Olshausen, Bruno A and Field, David J},
  journal={Nature},
  volume={381},
  number={6583},
  pages={607--609},
  year={1996},
  publisher={Nature Publishing Group UK London}
}

@article{fukuzaki2022principal,
  title={Principal Components of Neural Convolution Filters},
  author={Fukuzaki, Shota and Ikehara, Masaaki},
  journal={IEEE Access},
  volume={10},
  pages={104328--104336},
  year={2022},
  publisher={IEEE}
}

@article{pan2024dct,
  title={Dct-based decorrelated attention for vision transformers},
  author={Pan, Hongyi and Hamdan, Emadeldeen and Zhu, Xin and Biswas, Koushik and Cetin, Ahmet Enis and Bagci, Ulas},
  journal={arXiv preprint arXiv:2405.13901},
  year={2024}
}

@article{ning2024dctdiff,
  title={DCTdiff: Intriguing Properties of Image Generative Modeling in the DCT Space},
  author={Ning, Mang and Li, Mingxiao and Su, Jianlin and Jia, Haozhe and Liu, Lanmiao and Bene{\v{s}}, Martin and Salah, Albert Ali and Ertugrul, Itir Onal},
  journal={arXiv preprint arXiv:2412.15032},
  year={2024}
}

@inproceedings{he2016deep,
  title={Deep residual learning for image recognition},
  author={He, Kaiming and Zhang, Xiangyu and Ren, Shaoqing and Sun, Jian},
  booktitle={Proceedings of the IEEE conference on computer vision and pattern recognition},
  pages={770--778},
  year={2016}
}

@article{simonyan2014very,
  title={Very deep convolutional networks for large-scale image recognition},
  author={Simonyan, Karen and Zisserman, Andrew},
  journal={arXiv preprint arXiv:1409.1556},
  year={2014}
}

@inproceedings{bondesan2021hintons,
  title={The Hintons in your neural network: a quantum field theory view of deep learning},
  author={Bondesan, Roberto and Welling, Max},
  booktitle={International Conference on Machine Learning},
  pages={1038--1048},
  year={2021},
  organization={PMLR}
}

@book{hinton1984boltzmann,
  title={Boltzmann machines: Constraint satisfaction networks that learn},
  author={Hinton, Geoffrey E and Sejnowski, Terrence J and Ackley, David H},
  year={1984},
  publisher={Carnegie-Mellon University, Department of Computer Science Pittsburgh, PA}
}

@article{hopfield1982neural,
  title={Neural networks and physical systems with emergent collective computational abilities.},
  author={Hopfield, John J},
  journal={Proceedings of the national academy of sciences},
  volume={79},
  number={8},
  pages={2554--2558},
  year={1982}
}

@article{jumper2021highly,
  title={Highly accurate protein structure prediction with AlphaFold},
  author={Jumper, John and Evans, Richard and Pritzel, Alexander and Green, Tim and Figurnov, Michael and Ronneberger, Olaf and Tunyasuvunakool, Kathryn and Bates, Russ and {\v{Z}}{\'\i}dek, Augustin and Potapenko, Anna and others},
  journal={nature},
  volume={596},
  number={7873},
  pages={583--589},
  year={2021},
  publisher={Nature Publishing Group}
}

@INPROCEEDINGS{imagenet,
  author={Deng, Jia and Dong, Wei and Socher, Richard and Li, Li-Jia and Kai Li and Li Fei-Fei},
  booktitle={2009 IEEE Conference on Computer Vision and Pattern Recognition}, 
  title={ImageNet: A large-scale hierarchical image database}, 
  year={2009},
  volume={},
  number={},
  pages={248-255},
  doi={10.1109/CVPR.2009.5206848}}

@book{griffiths2020introduction,
  title={Introduction to elementary particles},
  author={Griffiths, David},
  year={2020},
  publisher={John Wiley \& Sons}
}

@article{krizhevsky2012imagenet,
  title={Imagenet classification with deep convolutional neural networks},
  author={Krizhevsky, Alex and Sutskever, Ilya and Hinton, Geoffrey E},
  journal={Advances in neural information processing systems},
  volume={25},
  year={2012}
}

@article{rudin2019stop,
  title={Stop explaining black box machine learning models for high stakes decisions and use interpretable models instead},
  author={Rudin, Cynthia},
  journal={Nature machine intelligence},
  volume={1},
  number={5},
  pages={206--215},
  year={2019},
  publisher={Nature Publishing Group UK London}
}

@inproceedings{sohl2015deep,
  title={Deep unsupervised learning using nonequilibrium thermodynamics},
  author={Sohl-Dickstein, Jascha and Weiss, Eric and Maheswaranathan, Niru and Ganguli, Surya},
  booktitle={International conference on machine learning},
  pages={2256--2265},
  year={2015},
  organization={pmlr}
}

@ARTICLE{bruna2013invariant,
  author={Bruna, Joan and Mallat, St{\'e}phane},
  journal={IEEE Transactions on Pattern Analysis and Machine Intelligence}, 
  title={Invariant Scattering Convolution Networks}, 
  year={2013},
  volume={35},
  number={8},
  pages={1872-1886},
  doi={10.1109/TPAMI.2012.230}}

@article{lindeberg1994scale,
  title={Scale-space theory: A basic tool for analyzing structures at different scales},
  author={Lindeberg, Tony},
  journal={Journal of applied statistics},
  volume={21},
  number={1-2},
  pages={225--270},
  year={1994},
  publisher={Taylor \& Francis}
}

@article{karniadakis2021physics,
  title={Physics-informed machine learning},
  author={Karniadakis, George Em and Kevrekidis, Ioannis G and Lu, Lu and Perdikaris, Paris and Wang, Sifan and Yang, Liu},
  journal={Nature Reviews Physics},
  volume={3},
  number={6},
  pages={422--440},
  year={2021},
  publisher={Nature Publishing Group}
}

@article{raissi2017physics,
  title={Physics informed deep learning (part i): Data-driven solutions of nonlinear partial differential equations},
  author={Raissi, Maziar and Perdikaris, Paris and Karniadakis, George Em},
  journal={arXiv preprint arXiv:1711.10561},
  year={2017}
}

@article{muller2023exact,
  title={Exact conservation laws for neural network integrators of dynamical systems},
  author={M{\"u}ller, Eike Hermann},
  journal={Journal of Computational Physics},
  volume={488},
  pages={112234},
  year={2023},
  publisher={Elsevier}
}

@inproceedings{grzeszczuk1998neuroanimator,
  title={Neuroanimator: Fast neural network emulation and control of physics-based models},
  author={Grzeszczuk, Radek and Terzopoulos, Demetri and Hinton, Geoffrey},
  booktitle={Proceedings of the 25th annual conference on Computer graphics and interactive techniques},
  pages={9--20},
  year={1998}
}

@misc{van2024energy,
      title={Energy-Conserving Neural Network for Turbulence Closure Modeling}, 
      author={Toby van Gastelen and Wouter Edeling and Benjamin Sanderse},
      year={2024},
      eprint={2301.13770},
      archivePrefix={arXiv},
      primaryClass={math.NA},
      url={https://arxiv.org/abs/2301.13770}, 
}

@article{raissi2018deep,
  title={Deep hidden physics models: Deep learning of nonlinear partial differential equations},
  author={Raissi, Maziar},
  journal={Journal of Machine Learning Research},
  volume={19},
  number={25},
  pages={1--24},
  year={2018}
}

@article{smets2023pde,
  title={PDE-based group equivariant convolutional neural networks},
  author={Smets, Bart MN and Portegies, Jim and Bekkers, Erik J and Duits, Remco},
  journal={Journal of Mathematical Imaging and Vision},
  volume={65},
  number={1},
  pages={209--239},
  year={2023},
  publisher={Springer}
}

@inproceedings{cohen2016group,
  title={Group equivariant convolutional networks},
  author={Cohen, Taco and Welling, Max},
  booktitle={International conference on machine learning},
  pages={2990--2999},
  year={2016},
  organization={PMLR}
}

@inproceedings{cohen2019gauge,
  title={Gauge equivariant convolutional networks and the icosahedral CNN},
  author={Cohen, Taco and Weiler, Maurice and Kicanaoglu, Berkay and Welling, Max},
  booktitle={International conference on Machine learning},
  pages={1321--1330},
  year={2019},
  organization={PMLR}
}

@article{alsallakh2025symmetries,
  title={On Symmetries in Convolutional Weights},
  author={Alsallakh, Bilal and Wroge, Timothy and Miglani, Vivek and Kokhlikyan, Narine},
  journal={arXiv preprint arXiv:2503.19215},
  year={2025}
}

@article{babaiee2024unveiling,
  title={Unveiling the unseen: Identifiable clusters in trained depthwise convolutional kernels},
  author={Babaiee, Zahra and Kiasari, Peyman M and Rus, Daniela and Grosu, Radu},
  journal={arXiv preprint arXiv:2401.14469},
  year={2024}
}

@article{aubury1996binomial,
  title={Binomial filters},
  author={Aubury, Matthew and Luk, Wayne},
  journal={Journal of VLSI signal processing systems for signal, image and video technology},
  volume={12},
  pages={35--50},
  year={1996},
  publisher={Springer}
}

@article{wells1986efficient,
  author={Wells, William M.},

  journal={IEEE Transactions on Pattern Analysis and Machine Intelligence}, 

  title={Efficient Synthesis of Gaussian Filters by Cascaded Uniform Filters}, 

  year={1986},

  volume={PAMI-8},

  number={2},

  pages={234-239},

  doi={10.1109/TPAMI.1986.4767776}}

@inproceedings{ruderman1993statistics,
 author = {Ruderman, Daniel and Bialek, William},
 booktitle = {Advances in Neural Information Processing Systems},
 editor = {J. Cowan and G. Tesauro and J. Alspector},
 pages = {},
 publisher = {Morgan-Kaufmann},
 title = {Statistics of Natural Images: Scaling in the Woods},
 url = {https://proceedings.neurips.cc/paper_files/paper/1993/file/4d5b995358e7798bc7e9d9db83c612a5-Paper.pdf},
 volume = {6},
 year = {1993}
}

@misc{chollet2015keras,
  title={Keras},
  author={Chollet, Fran\c{c}ois and others},
  year={2015},
  howpublished={\url{https://keras.io}},
}

@book{haar1909theorie,
  title={Zur theorie der orthogonalen funktionensysteme},
  author={Haar, Alfred},
  year={1909},
  publisher={Georg-August-Universitat, Gottingen.}
}

@article{gabor1946theory,
  title={Theory of communication. Part 1: The analysis of information},
  author={Gabor, Dennis},
  journal={Journal of the Institution of Electrical Engineers-part III: radio and communication engineering},
  volume={93},
  number={26},
  pages={429--441},
  year={1946},
  publisher={IET}
}

@inproceedings{luo2016understanding,
 author = {Luo, Wenjie and Li, Yujia and Urtasun, Raquel and Zemel, Richard},
 booktitle = {Advances in Neural Information Processing Systems},
 editor = {D. Lee and M. Sugiyama and U. Luxburg and I. Guyon and R. Garnett},
 pages = {},
 publisher = {Curran Associates, Inc.},
 title = {Understanding the Effective Receptive Field in Deep Convolutional Neural Networks},
 url = {https://proceedings.neurips.cc/paper_files/paper/2016/file/c8067ad1937f728f51288b3eb986afaa-Paper.pdf},
 volume = {29},
 year = {2016}
}

@article{krizhevsky2009learning,
  title={Learning multiple layers of features from tiny images},
  author={Krizhevsky, Alex and Hinton, Geoffrey and others},
  year={2009},
  publisher={Toronto, ON, Canada}
}

@inproceedings{nair2010rectified,
  title={Rectified linear units improve restricted boltzmann machines},
  author={Nair, Vinod and Hinton, Geoffrey E},
  booktitle={Proceedings of the 27th international conference on machine learning (ICML-10)},
  pages={807--814},
  year={2010}
}

@inproceedings{maas2013rectifier,
  title={Rectifier nonlinearities improve neural network acoustic models},
  author={Maas, Andrew L and Hannun, Awni Y and Ng, Andrew Y and others},
  booktitle={Proc. icml},
  volume={30},
  number={1},
  pages={3},
  year={2013},
  organization={Atlanta, GA}
}

@inproceedings{glorot10a,
  title={Understanding the difficulty of training deep feedforward neural networks},
  author={Glorot, Xavier and Bengio, Yoshua},
  booktitle={Proceedings of the thirteenth international conference on artificial intelligence and statistics},
  pages={249--256},
  year={2010},
  organization={JMLR Workshop and Conference Proceedings}
}

@inproceedings{he15,
author = {He, Kaiming and Zhang, Xiangyu and Ren, Shaoqing and Sun, Jian},
title = {Delving Deep into Rectifiers: Surpassing Human-Level Performance on ImageNet Classification},
booktitle = {Proceedings of the IEEE International Conference on Computer Vision (ICCV)},
month = {December},
year = {2015}
}

@inproceedings{glorot2011deep,
  title={Deep sparse rectifier neural networks},
  author={Glorot, Xavier and Bordes, Antoine and Bengio, Yoshua},
  booktitle={Proceedings of the fourteenth international conference on artificial intelligence and statistics},
  pages={315--323},
  year={2011},
  organization={JMLR Workshop and Conference Proceedings}
}
}

\clearpage
\setcounter{page}{1}
\maketitlesupplementary
\appendix

Here we provide additional results and visualizations to accompany our work in characterizing the mechanics the rectified convolution operation, demonstrate how even and odd filter components act upon image information as rest and kinetic energy operators, respectively, where the velocity of directional information is determined by the ratio of kinetic to total filter energy, i.e. the ratio of odd to total energy.

%\Cref{sym2d_definition} presents a framework for characterizing filters geometrically in terms of even and odd components, which in convolution followed by rectification cause information to diffuse symmetrically or propagate directionally.

In Section~\ref{sec:sym2d_definition} we provide definitions of even and odd symmetry for 2D filters.

In Section~\ref{sec:sup_training} we demonstrate the role of DCT coefficients and the dominance of primary DC and gradient components $\Sigma, \nabla_x, \nabla_y$ in training from scratch accuracy in spectral energy distributions, from VGG and Resnet models.

In Section~\ref{sec:sup_propagation} we provide additional demonstrations of our information propagation theory while mixing between even (e.g. DC $\Sigma$) and odd (e.g. gradient $\nabla_x,\nabla_y$) components, for various combinations of test patterns (pixel, circle), filter sizes (2x2, 3x3), types (DC, gradient, translation) and activation functions (none, ReLU, Modulus).

\section{Even-Odd Symmetry for 2D Images} \label{sec:sym2d_definition}

\begin{figure}[h]
\centering
        \includegraphics[width=0.25\textheight]{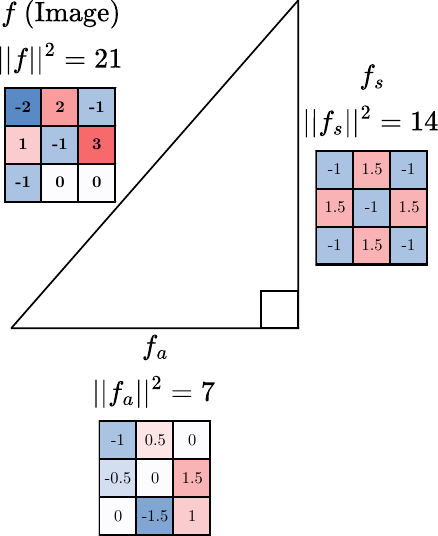}
        \caption{Illustrating the Pythagorean geometry of a discrete 2D image $f$ (e.g. $3 \times 3$ kernel) decomposed into orthogonal odd $f_o$ and even $f_e$ components}
        \label{fig:antisym_example-2d}
\end{figure}

Our work assumes even functions defined by rotational symmetry on a 2D lattice, as characterized by the dihedral group, here we introduce definitions used. Let $f(x,y)$ be a discrete 2D image or kernel of size $N \times N$ pixels, defined as a mapping $f: \mathbb{Z}^2 \rightarrow \mathbb{R}^1$ from 2D coordinates $(x,y) \in \mathbb{Z}^2$ to a scalar value $f \in \mathbb{R}^1$. Any function $f(x,y)$ may be decomposed into a sum $f(x,y) = f_e(x,y) + f_o(x,y)$ of an even rotationally symmetric component $f_e(x,y)$ and an odd component $f_o(x,y)$ whose magnitudes follow a Pythagorean relationship as shown in \Cref{fig:antisym_example-2d}.

Coordinates $(x,y)$ are taken with respect to the image centre $(\frac{N-1}{2},\frac{N-1}{2})$ without loss of generality and the primary properties of even and odd functions in 2D are as follows.

\begin{definition}[Even (Symmetric) Image]\label{def:sym_im}
An image $f_e(x,y) \in \mathbb{R}^{n\times n}$ is rotationally symmetric if 
\[ f_e(x,y) = f_e(\pm x, \pm y) = f_e(\pm y, \pm x),\]
where the unique value of $f_s(x,y)$ is the average of the set of equidistant points $\{(x,y): r = \sqrt{x^2+y^2}\}$ forming sign and coordinate permutations of $(x,y)$:
\begin{align}
    f_s(x,y) = \frac{1}{8}\sum_{s_x,s_y \in \{\pm 1\}} f(s_xx,s_yy) + f(s_yy,s_xx) \label{eq:sym_decomp}
\end{align}
\end{definition}

\begin{definition}[Odd (Anti-Symmetric) Image]\label{def:antisym_im}An image $f_o
(x,y) \in \mathbb{R}^{n\times n}$ is rotationally anti-symmetric if 
\[ f_o(x,y) = f(x,y) - f_e(x,y)\]
where the sum of $f_o(x,y)$ over the set of equidistant points $\{(x,y): r = \sqrt{x^2+y^2}\}$ forming sign and coordinate permutations of $(x,y)$ is 0:
\begin{align}
    \sum_{s_x,s_y \in \{\pm 1\}} f_o(s_xx,s_yy) + f_o(s_yy,s_xx) = 0
\end{align}
\end{definition}
% Add proof

\begin{lemma}[Orthogonality]\label{def:anti_sym_orthogonality}
Even and odd components are orthogonal
and their scalar or dot product is thus 0:
\[ f_e(x,y) \cdot f_o(x,y)  = \sum_{x,y}  f_e(x,y) f_o(x,y) = 0  \]
\end{lemma}

\begin{definition}[Energy]\label{def:filter_energy}
The energy of an image $f$ is defined as the squared magnitude $\|f(x,y)\|^2$, which equals the sum of squared magnitudes of the even and odd components:
\[ \|f(x,y)\|^2 = \sum_{x,y} f^2(x,y) = \sum_{x,y}  f_e^2(x,y) + \sum_{x,y} f_o^2(x,y).  \]
\end{definition}

\section{Additional Training Results}
\label{sec:sup_training}

Here we provide additional results and explanations regarding training experiments, even and odd components, and the discrete cosine transform (DCT) basis.

\subsection{The DCT basis and Even and Odd filter components}

Even and odd filters in 2D  may generally take on a variety of unique patterns, \eg DC $\Sigma$, gradients $\nabla$ and higher order patterns for larger filter sizes. Here, we show how even and odd filters may be generally grouped as components of the discrete cosine transform (DCT) frequency transform, as is commonly done in image and video compression. In the following training experiments, we interpret the contribution of each DCT basis towards the task of classification. 

%Here we demonstrate that these modes of propagation may be generalized to 2D using rotational symmetry as defined in the previous section and to larger filter sizes via the superposition of the discrete cosine transform (DCT) basis.

\begin{figure*}[h]
  \centering
  \begin{tabular}{cc}
         \includegraphics[width=.3\textwidth]{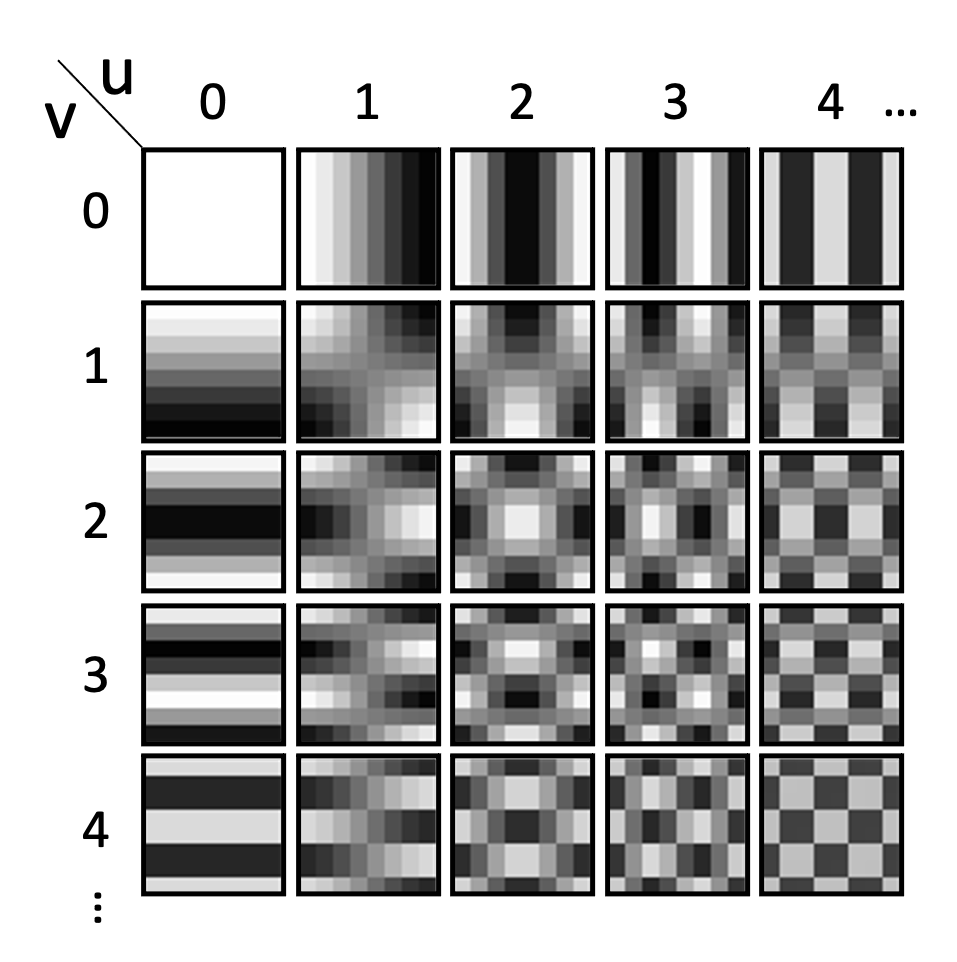}   &      \includegraphics[width=.3\textwidth]{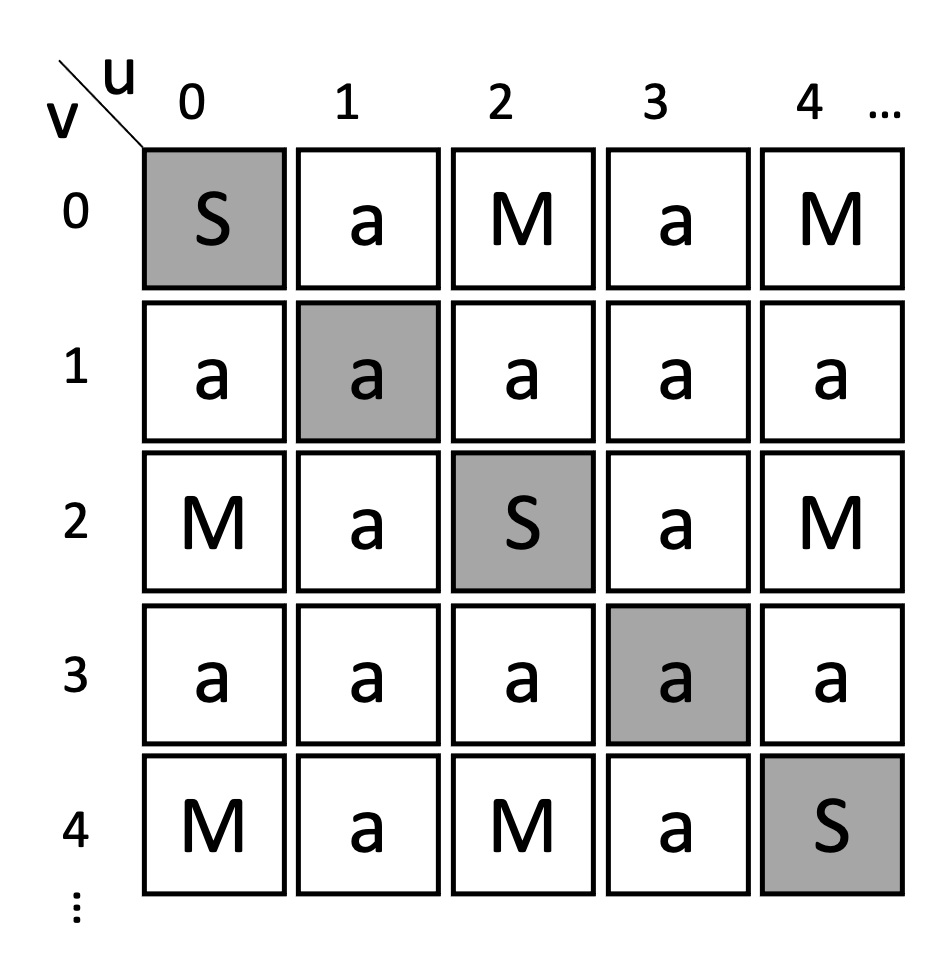} \\
    a) DCT Bases & b) Symmetry
  \end{tabular} 
  \caption{a) shows the DCT bases up to index (4,4), b) indicates whether the basis is even (symmetric) S, odd (antisymmetric) a or mixed M}
  \label{fig:DCT-Symmetry}
\end{figure*}

\begin{figure*}[h]
  \centering
  \begin{tabular}{cc}
         \includegraphics[width=.3\textwidth]{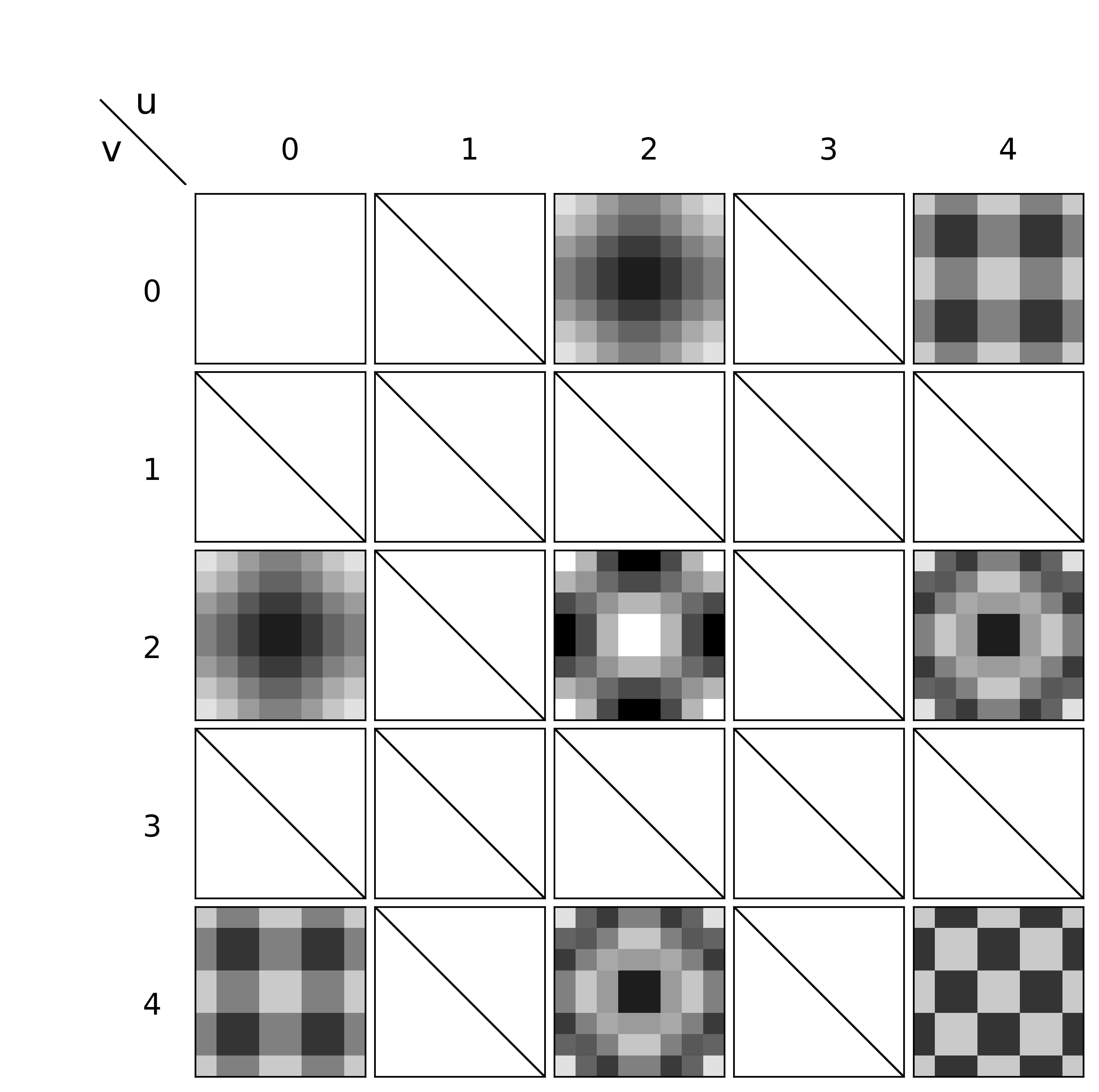}   &      \includegraphics[width=.3\textwidth]{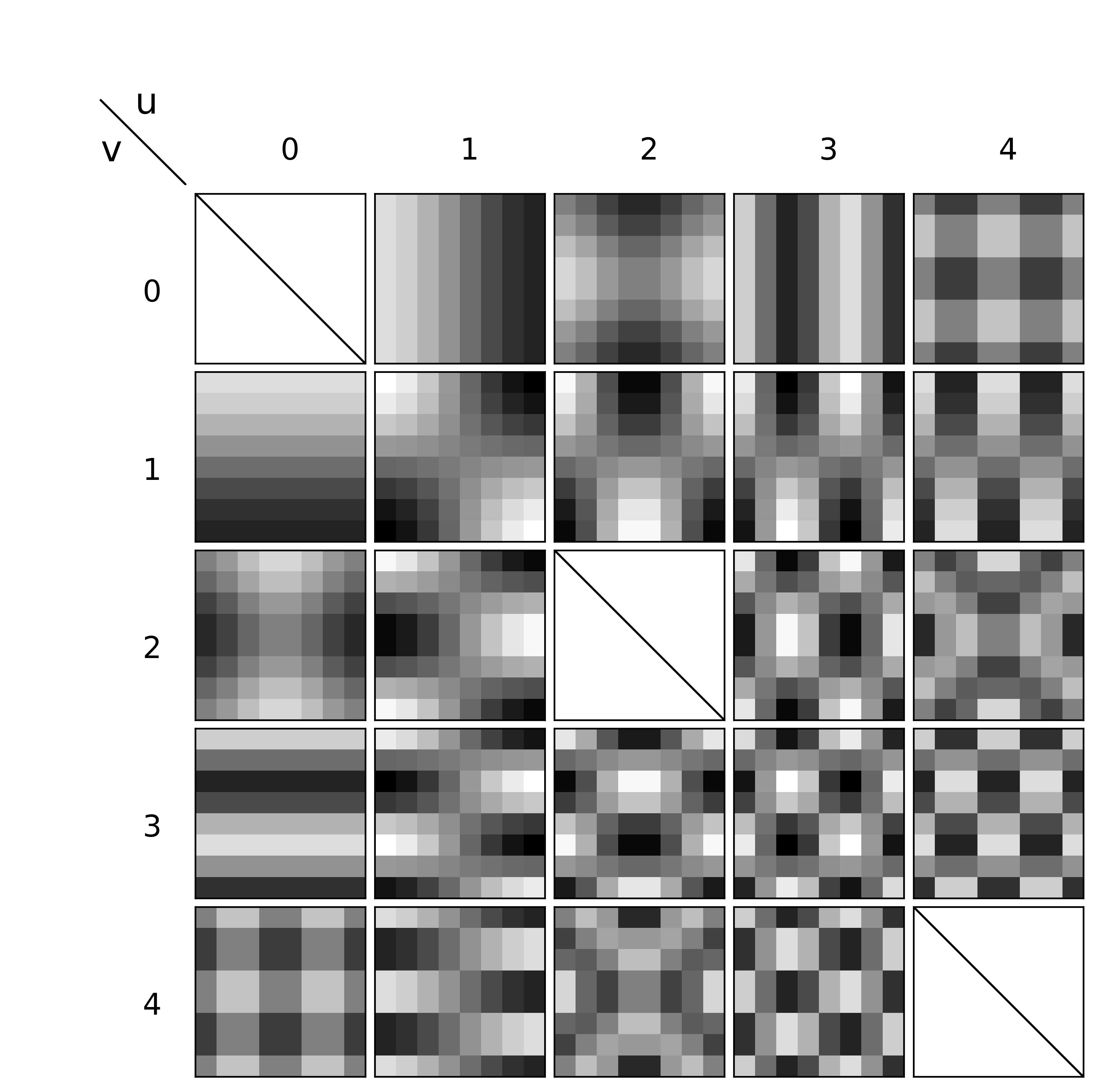} \\
    a) Even (symmetric) & b) Odd (antisymmetric)
  \end{tabular}
  \caption{Illustrating the even a) and odd b) components of DCT bases, black and white indicate negative and positive values.}
  \label{fig:DCT-Decomposed}
\end{figure*}

In general, an $N\times N$-pixel filter may be represented as a sum of $N^2$ discrete cosine transform (DCT) coefficients, each of which may be purely even (Symmetric, S), purely odd (Antisymmetric, A) or mixed even + odd (M). Figure~\ref{fig:DCT-Symmetry} a) shows the DCT basis functions up to index or wave number $(u,v) = (4,4)$. Figure~\ref{fig:DCT-Symmetry} b) shows the pattern of symmetry ascribed each DCT basis. As shown in Figure~\ref{fig:DCT-Decomposed}, we may observe that where one or two wave numbers $(u,v)$ are odd, the DCT basis is odd (A). If both indices $(u,v)$ are even and equal $u = v$, \ie along the main diagonal, the basis is even (S). If both indices $(u,v)$ are even but unequal $u \ne v$, then the basis is mixed even and odd (M).  Figure~\ref{fig:DCT-Decomposed} shows the pattern of even (symmetric) and odd (antisymmetric) components for each DCT basis. To our knowledge, this is the first time the DCT has been expressed as even and odd components, despite the widespread use of the DCT in data compression.

%Note for symmetric components in Figure~\ref{fig:DCT-Decomposed} a), off-diagonal coefficient pairs $(u,v)$ and $(v,u)$ are equal. Note that for antisymmetric components in Figure~\ref{fig:DCT-Decomposed} b), off-diagonal coefficient pairs $(u,v)$ and $(v,u)$ differ by an in-plane rotation angle of either $\pi/2$ or $3\pi/2$ and may be used as quadrature pairs to generate intermediate rotations.

%Furthermore, antisymmetric coefficients along anti-diagonal lines for $(u,v)$ such that $u+v=n$ represent derivatives of order $n$ and may be interpolated by the multi-angle formula:
%\begin{align}
%\sin(n\theta) &= \sum_{k=0}^n \frac{n!}{(2k-1)!(n-2k+1)!}\sin^{2k-1}(\theta)\cos^{n-2k+1}(\theta)\\
%\cos(n\theta) &= \sum_{k=0}^n \frac{n!}{(2k)!(n-2k)!}\sin^{2k}(\theta)\cos^{n-2k}(\theta)
%\end{align}

In practice, images or kernels representing natural images are dominated by low frequency DCT components. Particularly in our framework for small filters, \ie. $3\times3$ pixels, the even (or symmetric) component $f_e$ may be approximated by DC or sum:
\begin{align}
    f_e(x,y) &= \sum_{u,v~\in~Even} \omega_{u,v} D_{u,v} \approx \omega_{0,0}D_{0,0} = \omega_{0,0}\Sigma.
\end{align}

While the odd (or antisymmetric) component $f_o$ may be approximated by the gradient or difference $\nabla$, where a single angular parameter $\theta$ defines the gradient orientation as a linear mix between horizontal $\nabla_x$ and vertical $\nabla_y$ gradient components:
\begin{align}
    f_o(x,y) &= \sum_{u,v~\in~Odd~\cup~u \ne v} \omega_{u,v}D_{u,v}, \notag \\ &\approx \omega_{0,1}D_{0,1} + \omega_{1,0}D_{1,0} \propto \cos{\theta}\nabla_x  + \sin{\theta}\nabla_y .
\end{align}

This approximation is validated in experiments, where retraining CNNs with only three of nine filter components $(\Sigma,\nabla_x,\nabla_y)$ results in greater than 90\% of baseline accuracy for typical networks, e.g. VGG and Resnet.

\clearpage
\pagebreak

\subsection{Training from Scratch on the CIFAR-100 Dataset using DCT components}

This experiment consists of training a VGG16~\citep{simonyan2014very} and  Resent20~\cite{he2016deep} models with various numbers of DCT coefficients, from a single DC ($\Sigma$) parameter to 9 total components (full spectrum). DCT kernel weight parameters $\{\omega_i\}$ are used, which are updated during backpropagation and inverse transformed into filters for forward propagation. For all 6 runs, we use the same hyperparameters which yielded the best accuracy for the baseline. 

As we can see in \Cref{tab:dct_train}, the run with only DC ($\Sigma$) and low-order gradient ($\nabla_x$, $\nabla_y$) components, VGG16 achieves 91\% of baseline validation accuracy. We observe the same behaviour when training Resnet20 (see \Cref{tab:dct_train_resnet}).

\begin{table}[htbp]
  \centering
  \caption{Training VGG16  on CIFAR-100~\cite{krizhevsky2009learning} using convolutional kernels composed of progressively additional high-order DCT components. We find that only 3 low-frequency components (underlined) contribute to 91\% of VGG16 baseline accuracy.}
\begin{tabular}{l|l|l}
   {Number of DCT Components} & 
    Val-Accuracy ($\pm$ std) &
    $\%$ of Baseline \\
    \hline
    \hline
    1 $(\Sigma)$ & 0.3247 $\pm$ 0.0052 & 0.44 \\
    \underline{3 $(\Sigma, \nabla_x, \nabla_y)$} & 
    \underline{0.6664 $\pm$ 0.0017} & 
    \underline{0.91} \\
    4 & 0.6823 $\pm$ 0.0039 & 0.93 \\
    6 & 0.7162 $\pm$ 0.0055 & 0.98 \\
    8 & 0.7294 $\pm$ 0.0032 & 0.99 \\
    \textbf{9 (Baseline)} & 
    \textbf{0.7299 $\pm$ 0.0019} & 
    \textbf{1.00} \\
    \hline
    \hline
\end{tabular}
  \label{tab:dct_train}%
\end{table}%

\begin{table}[htbp]
  \centering
  \caption{Training Resnet20  on CIFAR-100~\cite{krizhevsky2009learning} using convolutional kernels composed of progressively additional high-order DCT components. We find that only 3 low-frequency components (underlined) contribute to 92\% of Resnet20 baseline accuracy.}
    \begin{tabular}{l|l|l}
    {Number of DCT Components} & Val-Accuracy ($\pm$ std) &
    $\%$ of Baseline \\
    \hline
    \hline
    1 $(\Sigma)$ & 0.4301 $\pm$ 0.022 & 0.63 \\
   \underline{3 $(\Sigma, \nabla_x, \nabla_y)$}  & \underline{0.6277 $\pm$ 0.0025} & \underline{0.92} \\
    4 & 0.6413 $\pm$ 0.0084 & 0.94 \\
    6 &  0.6675 $\pm$ 0.0074 & 0.98 \\
    8  & 0.6759 $\pm$ 0.0069 & 0.99\\
    \textbf{9 (Baseline)} & \textbf{0.6805 $\pm$ 0.0104} & \textbf{1.00} \\
    \hline
    \hline
    \end{tabular}%
    
  \label{tab:dct_train_resnet}%
\end{table}%

\clearpage
\pagebreak

\subsection{The Energy of ImageNet-trained Kernels is concentrated into $\Sigma$ and $\nabla$ DCT components.}

\begin{figure}
\centering
		\includegraphics[width=0.7\linewidth]{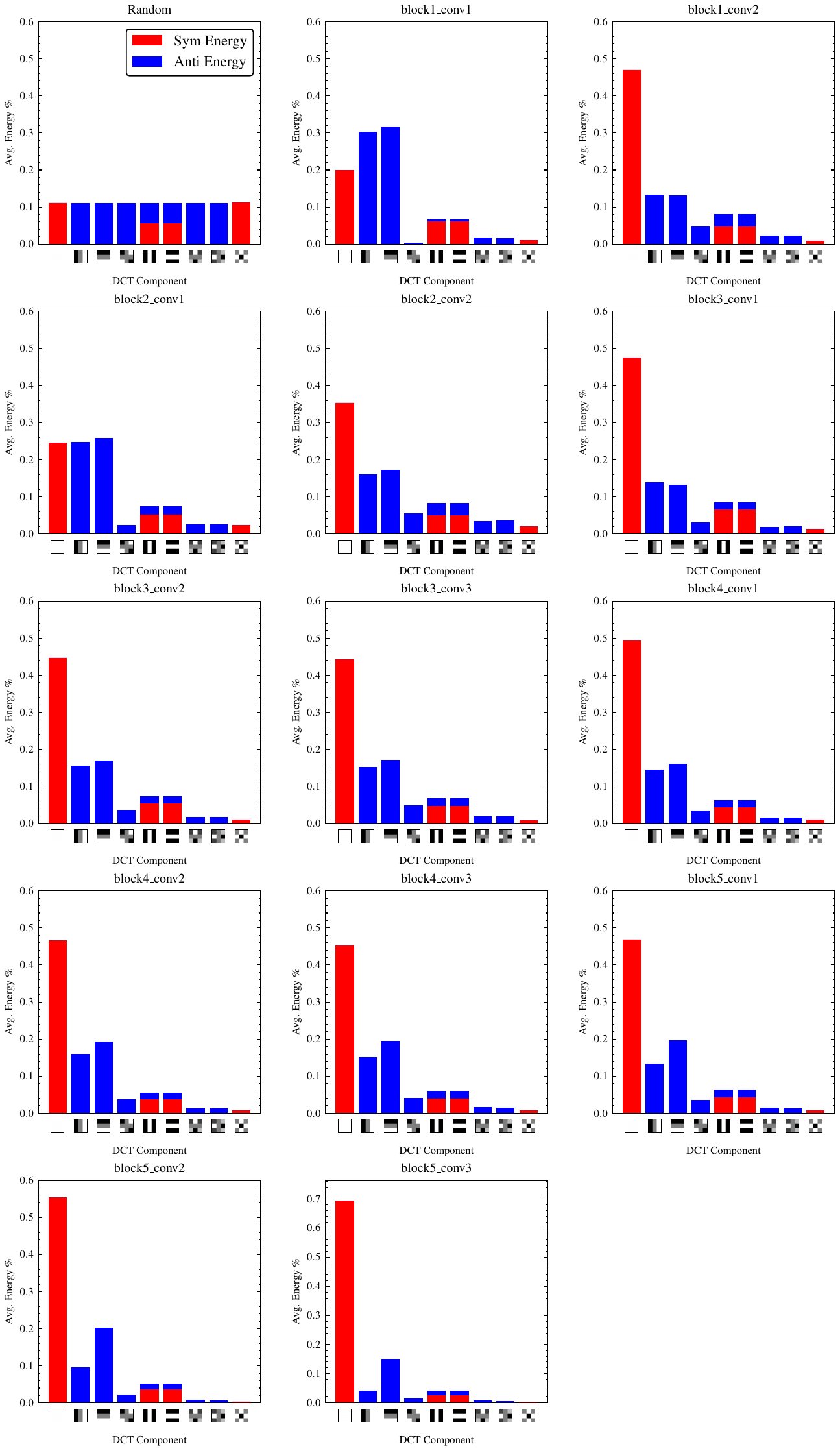}
\caption{Average energy distribution of DCT components ($\frac{\omega_i^2}{||\bm{\omega}||^2}$) in random and learned convolutional kernels (trained on Imagenet) throughout VGG16 layers.}
\label{fig:layer_dct_vgg}
\end{figure}

\begin{figure}
\centering
		\includegraphics[width=0.7\linewidth]{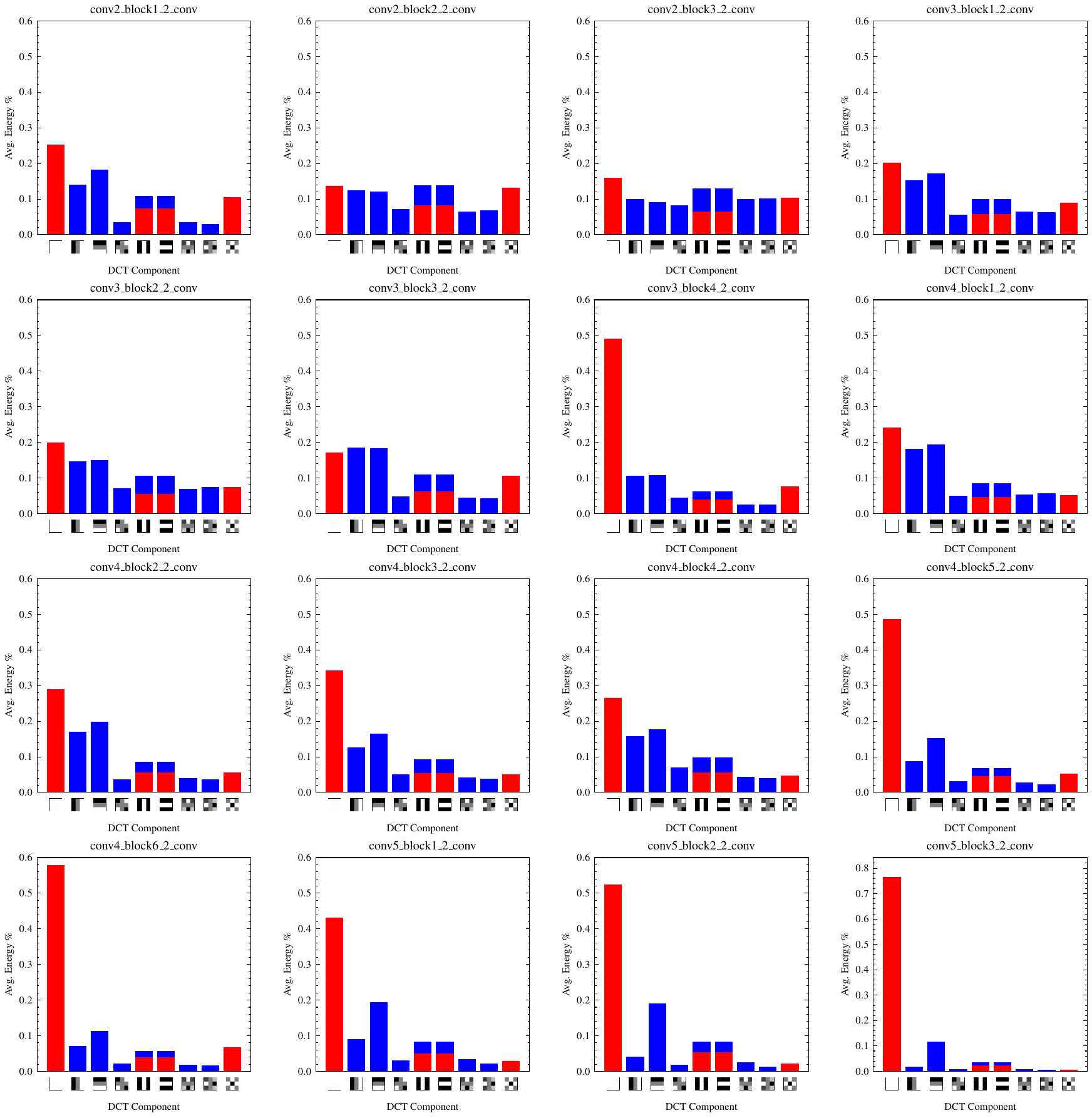}
\caption{Average energy distribution of DCT components ($\frac{\omega_i^2}{||\bm{\omega}||^2}$)  in learned convolutional kernels (trained on Imagenet) throughout Resnet50 layers.}
\label{fig:layer_dct_resnet}
\end{figure}

We report in \Cref{fig:layer_dct_vgg} and \Cref{fig:layer_dct_resnet} the average energy percentage for each frequency component $\omega_i$ of each kernel of VGG16~\citep{simonyan2014very} and Resnet50~\citep{he2016deep}, respectively, trained on Imagenet~\citep{imagenet} , across all layers. In \Cref{fig:dct-decomp-full}  we plot the average energy per spectral component $\omega_i$ for the entire network. As we can clearly see from \Cref{fig:dct-decomp-full}, the majority of the weights are either DC $\Sigma$ (even) or gradient $\nabla$ (odd) after training, whereas they are uniformly distributed across all components at initialization (\cref{fig:layer_dct_vgg}).

\begin{figure}[h]
  \centering
  \begin{subfigure}{0.35\linewidth}
    \includegraphics[width=\linewidth]{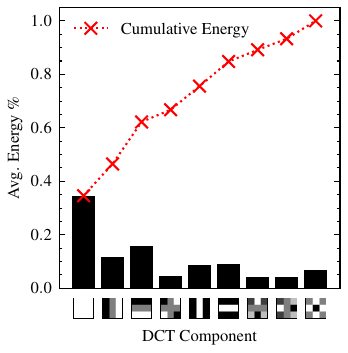}
    \caption{Resnet50}
    \label{fig:dct-full-resnet}
  \end{subfigure}
\hspace*{0.0000\linewidth}
\begin{subfigure}{0.35\linewidth}
    \includegraphics[width=\linewidth]{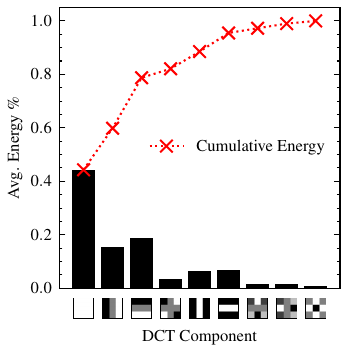}
    \caption{VGG16}
    \label{fig:dct-full-vgg}
  \end{subfigure}
  
\caption{Spectral DCT decomposition $\omega_i$ of all $3\times3$ convolutional filters in all layers of \subref{fig:dct-full-resnet})~Resnet50 and \subref{fig:dct-full-vgg})~VGG16. We find that in both models the majority of the weights are comprised of low order DC and Gradients ($\Sigma+\nabla$)  } 
  \label{fig:dct-decomp-full}
\end{figure}

\clearpage
\pagebreak

\section{Additional Propagation Demonstrations}
\label{sec:sup_propagation}

Here we demonstrate the results of rectified convolution from a single channel, and how the velocity of information is determined by the mixing ratio $\beta$ of even and odd filter components, similarly to the Lorentz transform in the theory of special relativity. Rectified convolution is repeatedly performed upon test patterns, and the velocity is measured in terms of the displacement of the centre of mass per convolution.

\subsection{Experimental Setup}

Here we demonstrate the mechanics by which even (e.g. DC $\Sigma$) and odd (e.g. gradient $\nabla_x,\nabla_y$) filter components act upon image information, similarly to \Cref{ssec:mec_prop} of the paper, for various combinations of test patterns (pixel, circle), filter sizes ($2\times2$, $3\times3$), types (DC, gradient, translation) and activation functions (none, ReLU, Modulus).

\Cref{tab:kernel_examples} shows the filter kernels used for various values of $\beta^2$. Most demonstrations mix $\Sigma$ and $\nabla_x$ components according to the $\beta^2$ parameter, and convolve a test pattern.  Note that 2x2 kernels are applied alternatingly within a 3x3 kernel in order to avoid a half-pixel shift following convolution. We also test a special case of propagation with a translation kernel, which is normally an offset impulse kernel (\Cref{tab:kernel_examples}, $3 \times 3$ translation for $\beta^2=0.75$).

\begin{figure}
  \centering
  \begin{subfigure}{0.3\linewidth}
    \includegraphics[width=\linewidth]{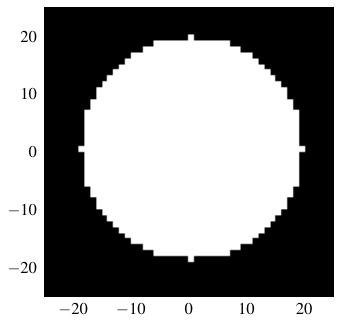}
    \caption{Test Circle (r=19)}
    \label{fig:test_circle}
  \end{subfigure}
\hspace*{0.001\linewidth}
\begin{subfigure}{0.3\linewidth}
    \includegraphics[width=\linewidth]{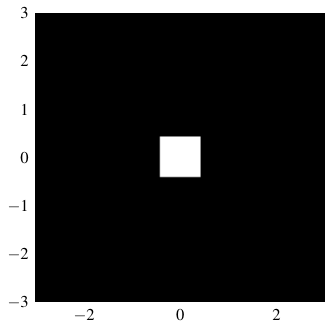}
    \caption{Test pixel (impulse)}
    \label{fig:test_pixel}
  \end{subfigure}
  \caption{Test Patterns }
  \label{fig:}
\end{figure}

Our demonstrations perform rectified convolution on two test image patterns including a circle ($r=19$) \cref{fig:test_circle}) and an impulse  (\cref{fig:test_pixel}). Between each iteration, the activation centre of mass $\mu_x$ and standard deviation $\sigma_x$ are computed from a normalized activation $f(x,y)$ as follows:
\begin{align}
    \mu_x &= \frac{\sum_{x} x \|f(x,0)\|}{\sum_{x} \|f(x,0)\|} ~~~~~~~~    \sigma^2 = \frac{\sum_{x} \|f(x,0)\| (x  - \mu_x)^2}{\sum_{x} \|f(x,0)\|}
\end{align}

\begin{table}[h]
\centering
\setlength{\tabcolsep}{6pt}
\renewcommand{\arraystretch}{1.6}

\begin{tabular}{l|ccccc}
\toprule
\textbf{Kernel size} & \multicolumn{5}{c}{$\boldsymbol{\beta^2}$} \\
\cmidrule(lr){2-6}
 & $0$ & $0.25$ & $0.5$ & $0.75$ & $1$ \\
\midrule

\textbf{2×2 (alternating)} &
\adjustbox{valign=m}{\includegraphics[width=.055\linewidth]{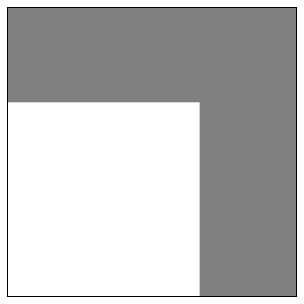}}
\adjustbox{valign=m}{\includegraphics[width=.055\linewidth]{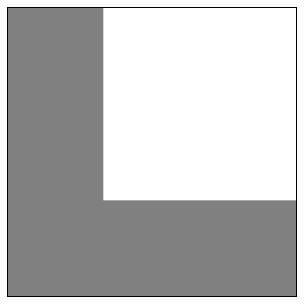}}
&
\adjustbox{valign=m}{\includegraphics[width=.055\linewidth]{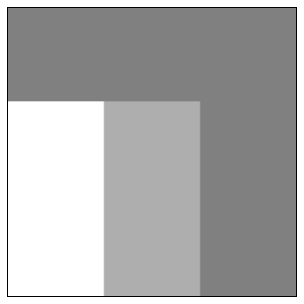}}
\adjustbox{valign=m}{\includegraphics[width=.055\linewidth]{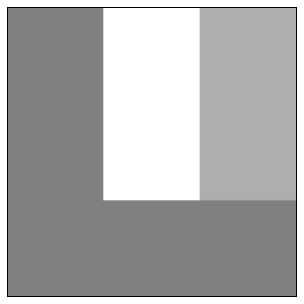}}
&
\adjustbox{valign=m}{\includegraphics[width=.055\linewidth]{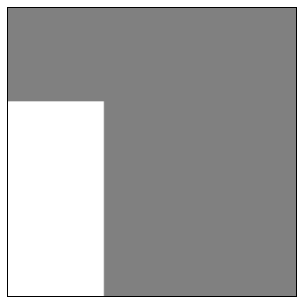}}
\adjustbox{valign=m}{\includegraphics[width=.055\linewidth]{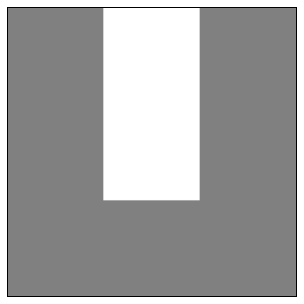}}
&
\adjustbox{valign=m}{\includegraphics[width=.055\linewidth]{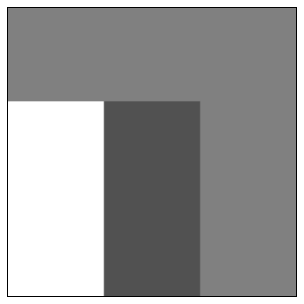}}
\adjustbox{valign=m}{\includegraphics[width=.055\linewidth]{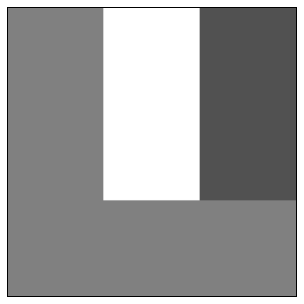}}
&
\adjustbox{valign=m}{\includegraphics[width=.055\linewidth]{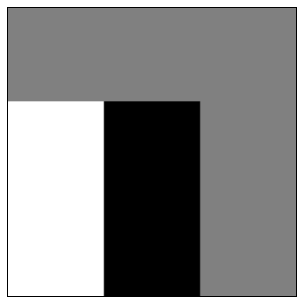}}
\adjustbox{valign=m}{\includegraphics[width=.055\linewidth]{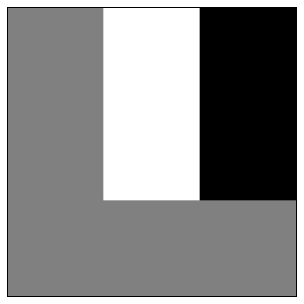}}
\\

\textbf{3×3} &
\adjustbox{valign=m}{\includegraphics[width=.055\linewidth]{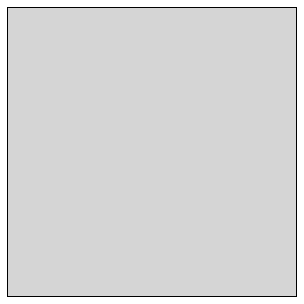}} &
\adjustbox{valign=m}{\includegraphics[width=.055\linewidth]{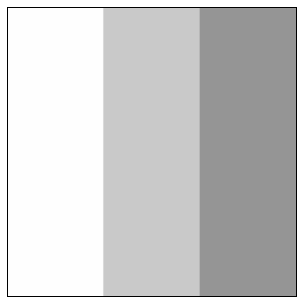}} &
\adjustbox{valign=m}{\includegraphics[width=.055\linewidth]{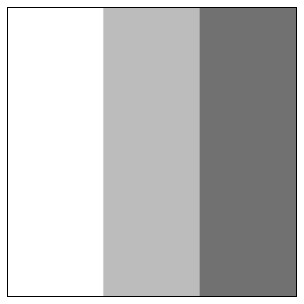}} &
\adjustbox{valign=m}{\includegraphics[width=.055\linewidth]{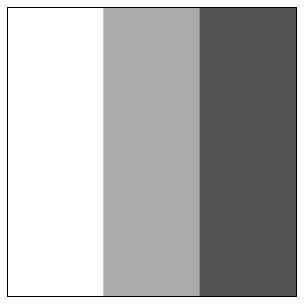}} &
\adjustbox{valign=m}{\includegraphics[width=.055\linewidth]{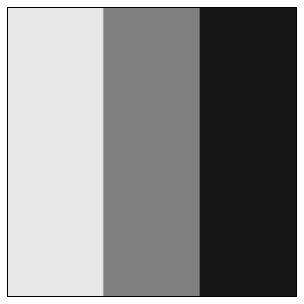}}
\\

\textbf{3×3 (translation)} &
\adjustbox{valign=m}{\includegraphics[width=.055\linewidth]{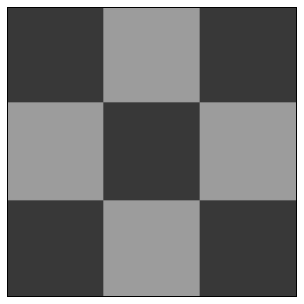}} &
\adjustbox{valign=m}{\includegraphics[width=.055\linewidth]{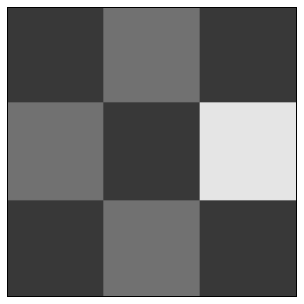}} &
\adjustbox{valign=m}{\includegraphics[width=.055\linewidth]{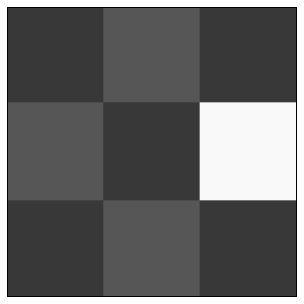}} &
\adjustbox{valign=m}{\includegraphics[width=.055\linewidth]{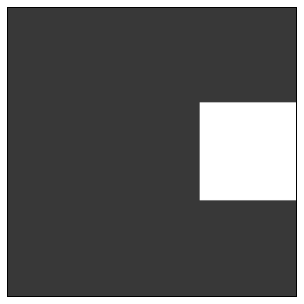}} &
\adjustbox{valign=m}{\includegraphics[width=.055\linewidth]{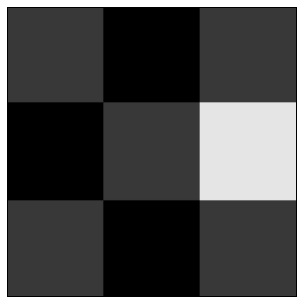}}
\\

\bottomrule
\end{tabular}
\caption{Examples of kernels used, where each $3\times 3$ kernel $f = \beta \hat{f_o} + \sqrt{1-\beta^2}\hat{f_e}$ is generated by mixing odd $f_o$ and even $f_e$ components according to mixing ratio $\beta$.}
\label{tab:kernel_examples}
\end{table}
%add kerenels

\clearpage
\pagebreak

\subsection{Convolution Without Activation}
\subsubsection{$3\times 3$ kernel, mixing unidirectional gradient $\nabla_x$ and sum $\Sigma$}
\begin{figure}[htbp]
  \centering
    \begin{minipage}{0.7\columnwidth}
      \centering
      \begin{subfigure}{\linewidth}
        \includegraphics[width=\linewidth]{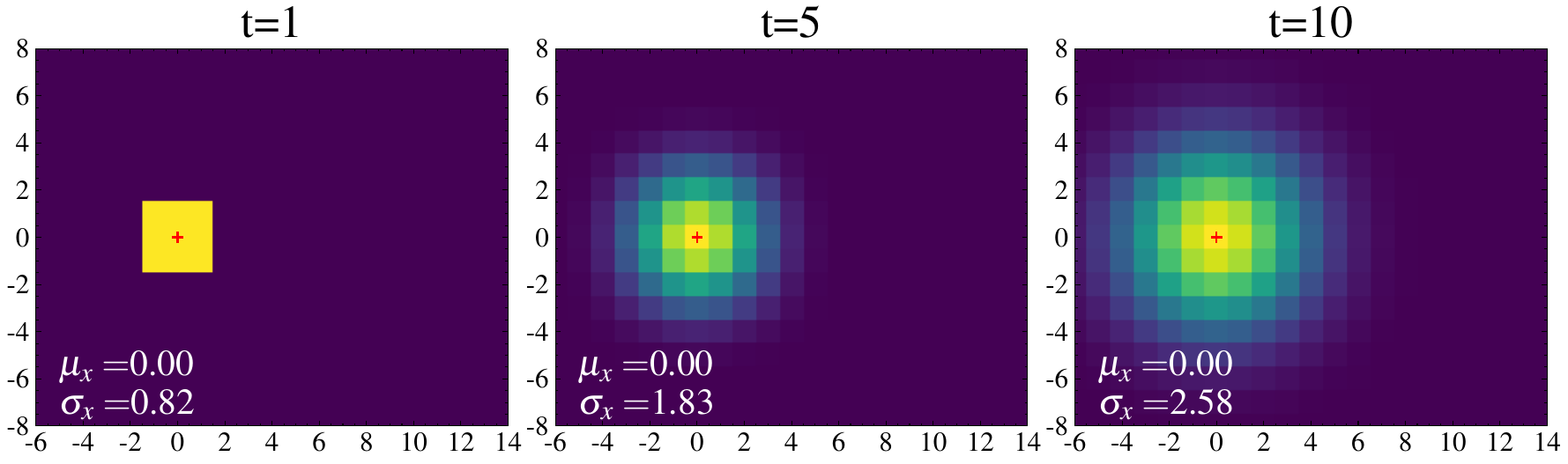}
        \caption{$\Sigma$ $(\beta^2=0)$}
        \label{fig:pixel_noact_uni_3x3_b0}
      \end{subfigure}\\
      \begin{subfigure}{\linewidth}
        \includegraphics[width=\linewidth]{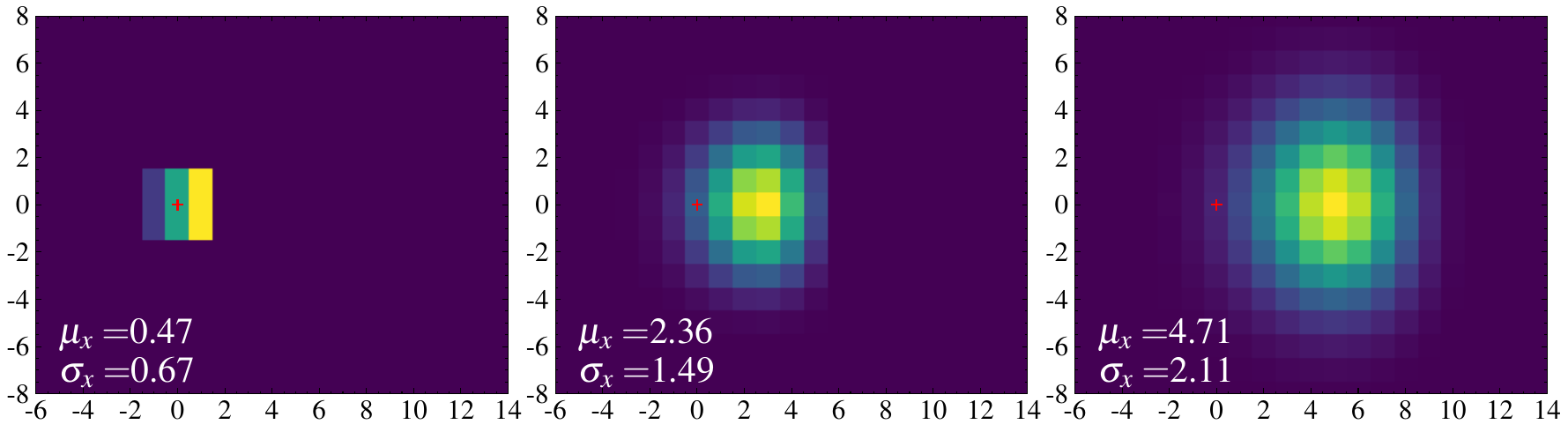}
        \caption{$\Sigma+\nabla_x$ $(\beta^2=0.25)$}
        \label{fig:pixel_noact_uni_3x3_b0.25}
      \end{subfigure}\\
      \begin{subfigure}{\linewidth}
        \includegraphics[width=\linewidth]{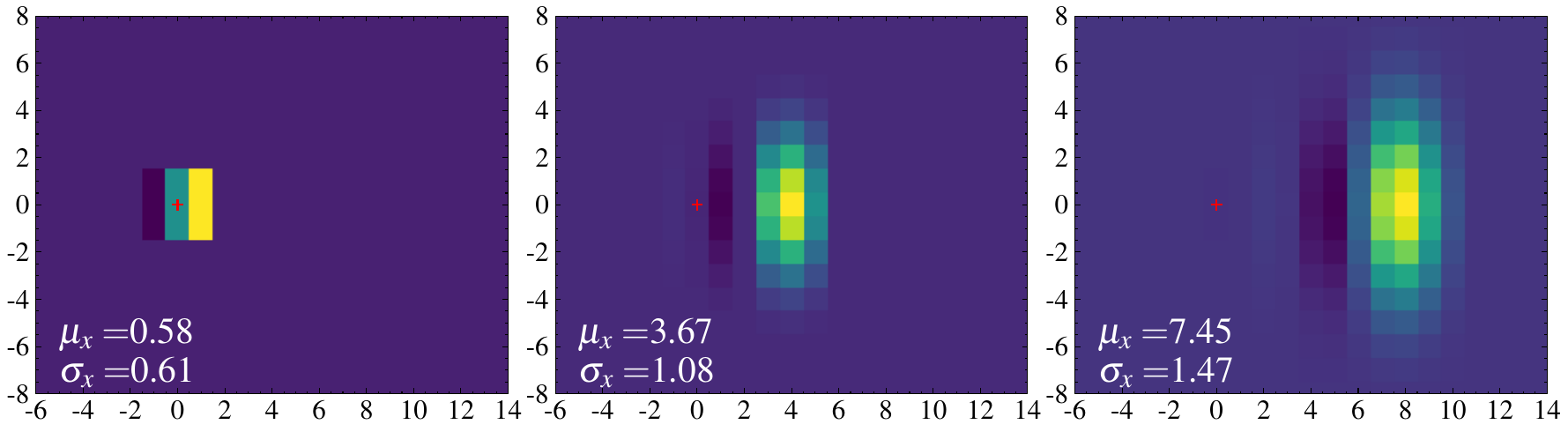}
        \caption{$\Sigma+\nabla_x$ $(\beta^2=0.5)$}
        \label{fig:pixel_noact_uni_3x3_b0.5}
      \end{subfigure}\\
      \begin{subfigure}{\linewidth}
        \includegraphics[width=\linewidth]{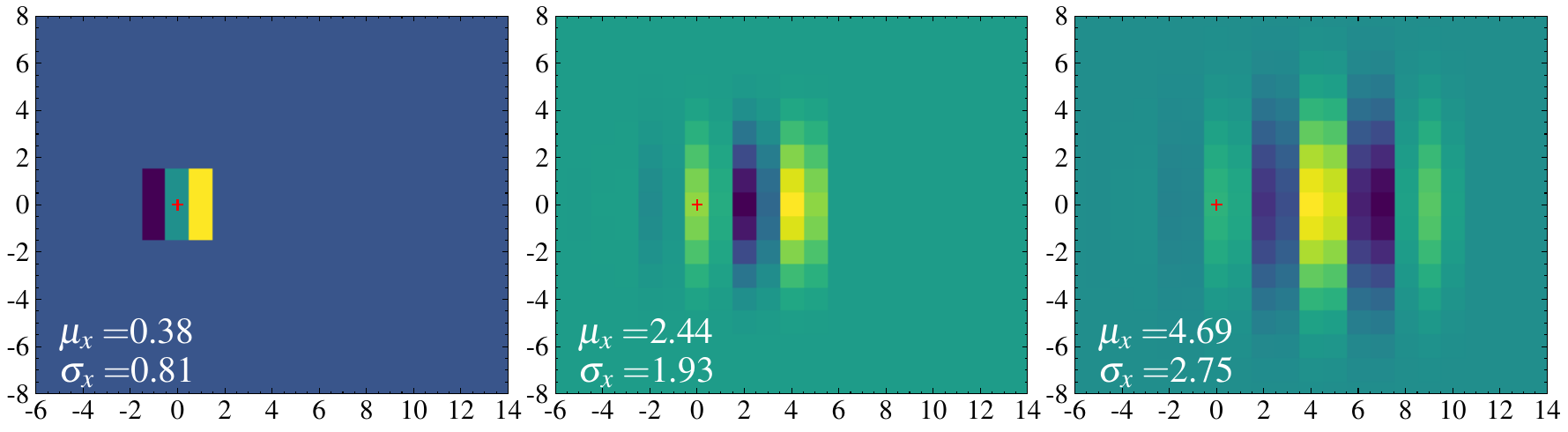}
        \caption{$\Sigma+\nabla_x$ $(\beta^2=0.75)$}
        \label{fig:pixel_noact_uni_3x3_b0.75}
      \end{subfigure}\\
      \begin{subfigure}{\linewidth}
        \includegraphics[width=\linewidth]{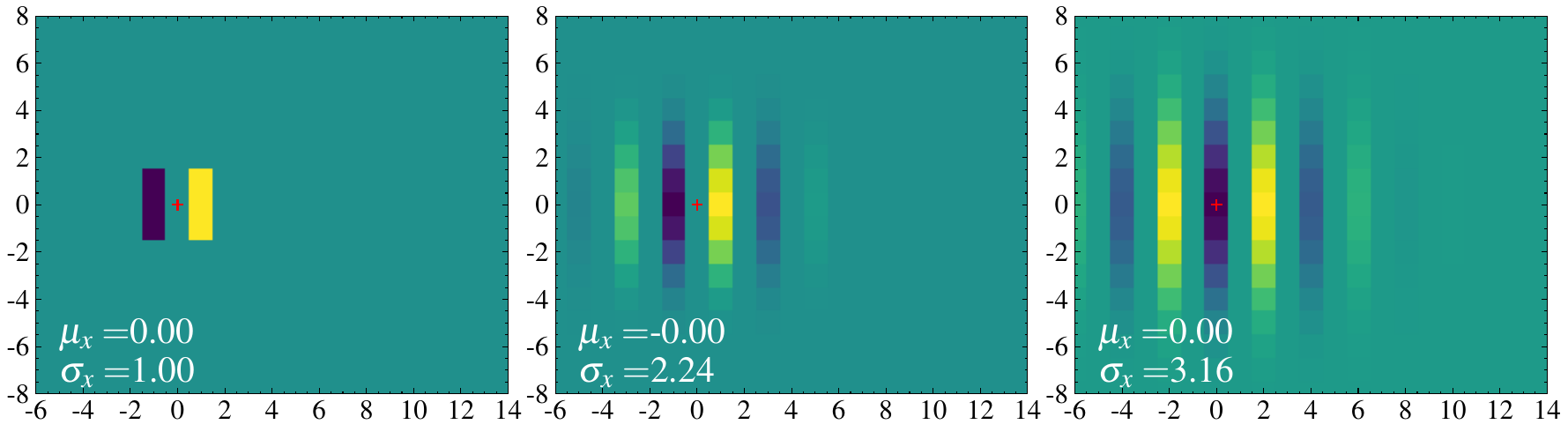}
        \caption{$\nabla_x$ $(\beta^2=1)$}
        \label{fig:pixel_noact_uni_3x3_b1}
      \end{subfigure}
    \end{minipage}%
  \caption{Demonstrating the effect of repeated convolution (no activation function) of a test pattern (pixel) over $3 \times 3$ kernels, varying $\beta$ mixing between $\Sigma$ and $\nabla$. Note that for both $\beta=0$ and $\beta=1$, there is net displacement of the centre of mass, this is distinctly different from rectified convolution.}
  \label{fig:pixel_noact_uni_3x3}
\end{figure}

\subsection{Convolution With ReLU (rectification) Activation}

\subsubsection{$3\times 3$ kernel, mixing unidirectional gradient $\nabla_x$ and sum $\Sigma$ components}

\begin{figure}[htbp]
  \centering
    \begin{minipage}{0.68\columnwidth}
      \centering
      \begin{subfigure}{\linewidth}
        \includegraphics[width=\linewidth]{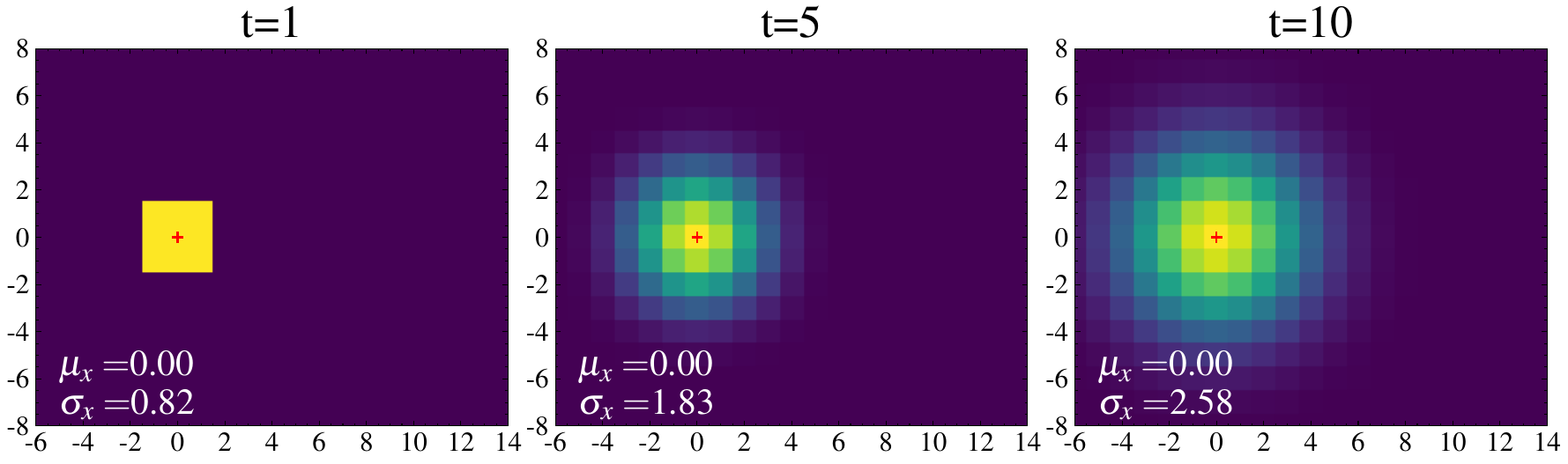}
        \caption{$\Sigma$ $(\beta^2=0)$}
        \label{fig:pixel_relu_uni_3x3_b0}
      \end{subfigure}\\
      \begin{subfigure}{\linewidth}
        \includegraphics[width=\linewidth]{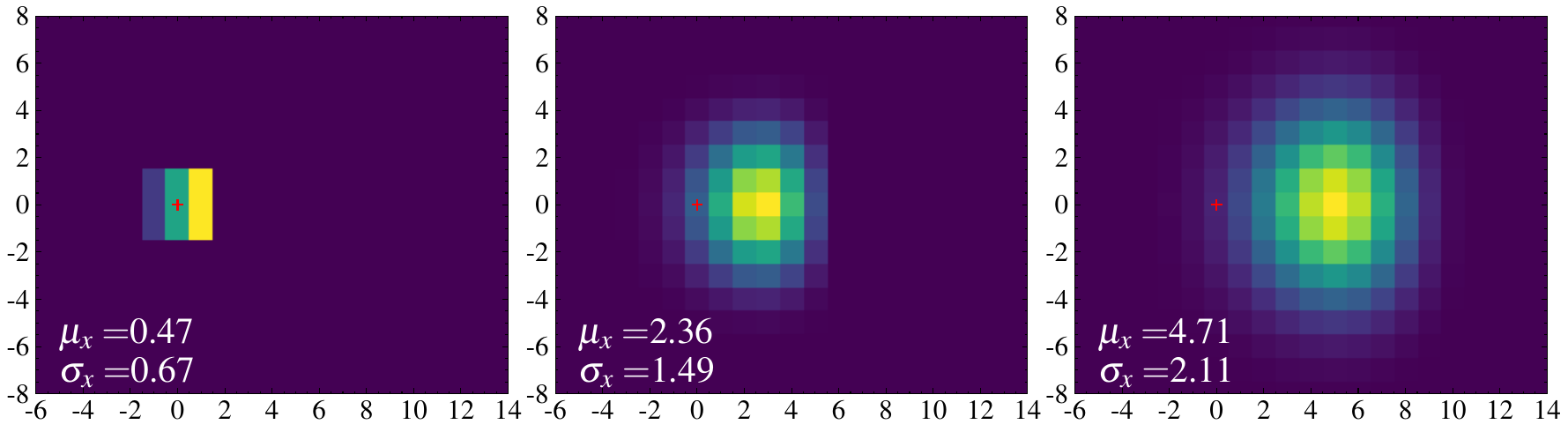}
        \caption{$\Sigma+\nabla_x$ $(\beta^2=0.25)$}
        \label{fig:pixel_relu_uni_3x3_b0.25}
      \end{subfigure}\\
      \begin{subfigure}{\linewidth}
        \includegraphics[width=\linewidth]{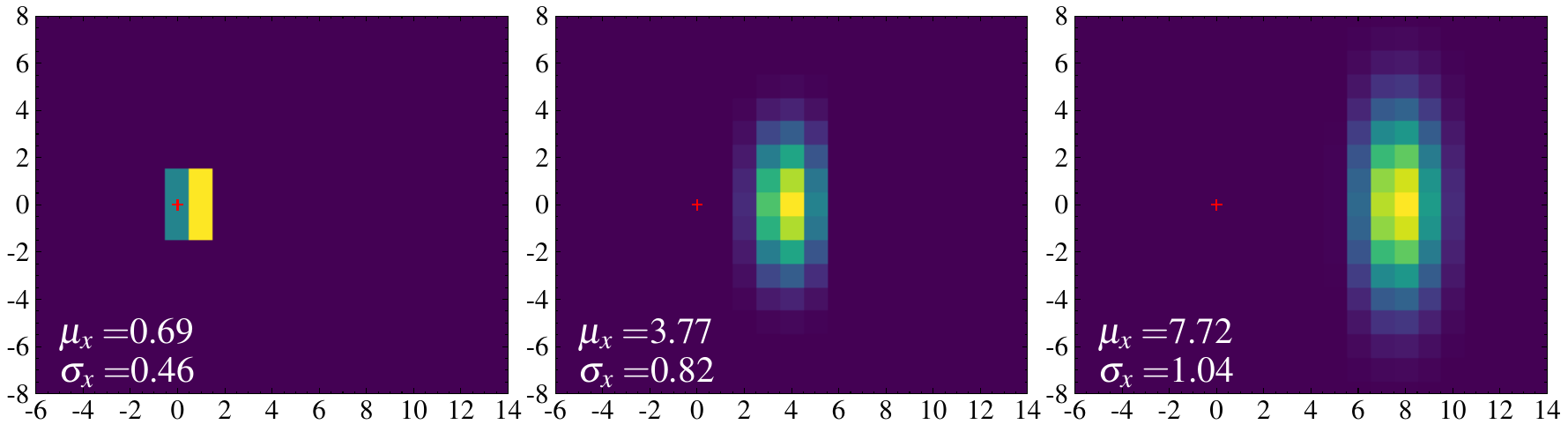}
        \caption{$\Sigma+\nabla_x$ $(\beta^2=0.5)$}
        \label{fig:pixel_relu_uni_3x3_b0.5}
      \end{subfigure}\\
      \begin{subfigure}{\linewidth}
        \includegraphics[width=\linewidth]{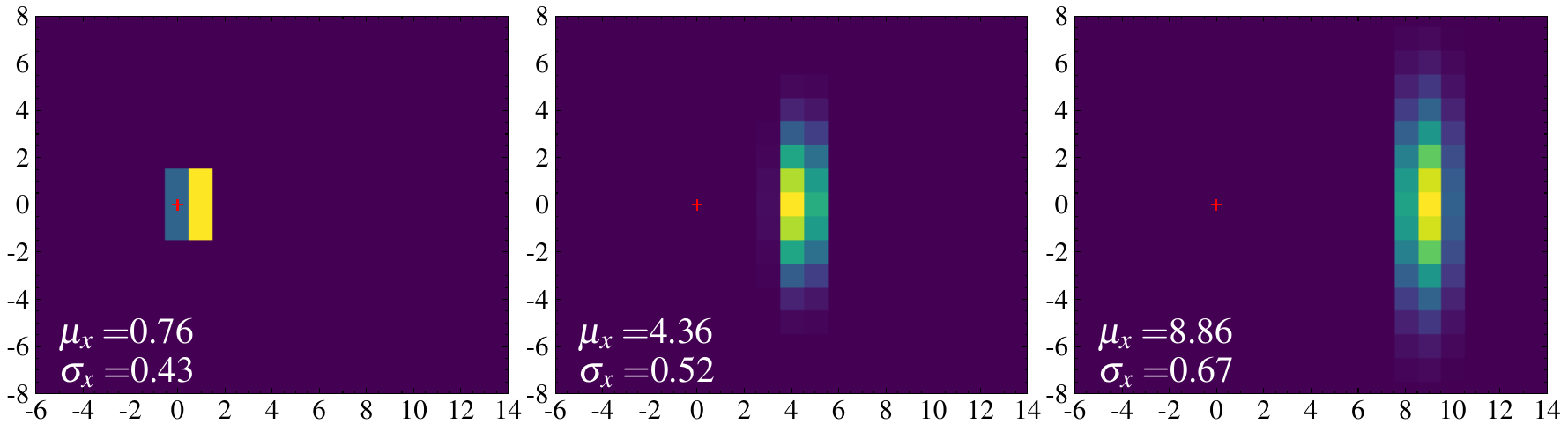}
        \caption{$\Sigma+\nabla_x$ $(\beta^2=0.75)$}
        \label{fig:pixel_relu_uni_3x3_b0.75}
      \end{subfigure}\\
      \begin{subfigure}{\linewidth}
        \includegraphics[width=\linewidth]{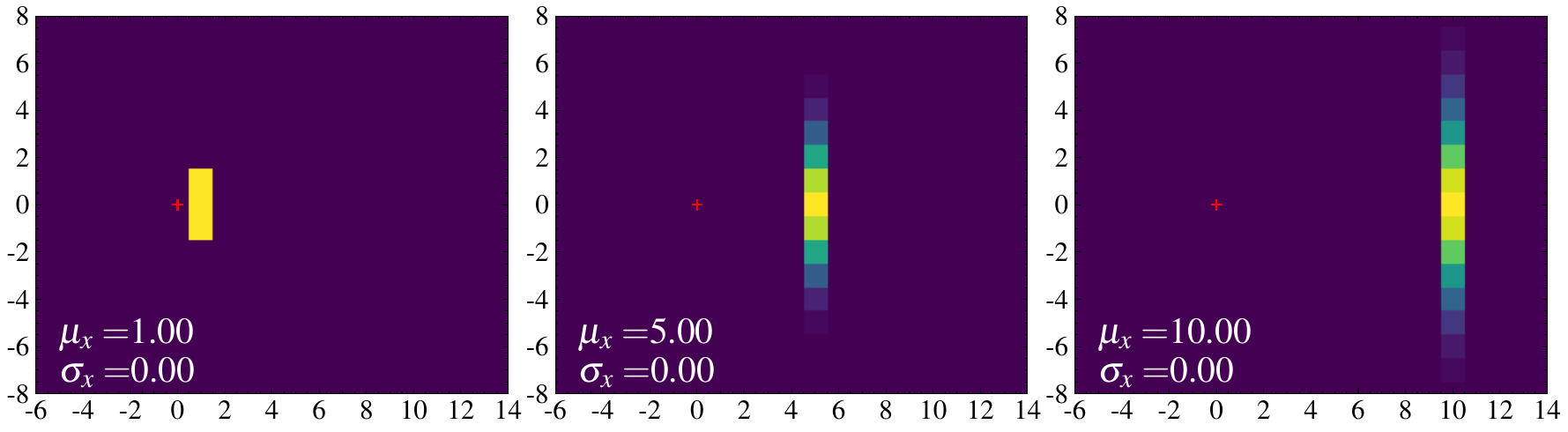}
        \caption{$\nabla_x$ $(\beta^2=1)$}
        \label{fig:pixel_relu_uni_3x3_b1}
      \end{subfigure}
    \end{minipage}%
  \caption{Demonstrating the effect of repeated convolution+ReLU of a test pattern over different types of $3\times3$ kernels (DC and Gradient). Note that for $\beta=0$ a), content diffuses symmetrically about a stationary centre of mass, while for $\beta=1$ the centre of mass translates rightward with maximum velocity.}
  \label{fig:pixel_relu_uni_3x3}
\end{figure}

\begin{figure}[htbp]
  \centering
    \begin{minipage}{0.7\linewidth}
      \centering
      \begin{subfigure}{\linewidth}
        \includegraphics[width=\linewidth]{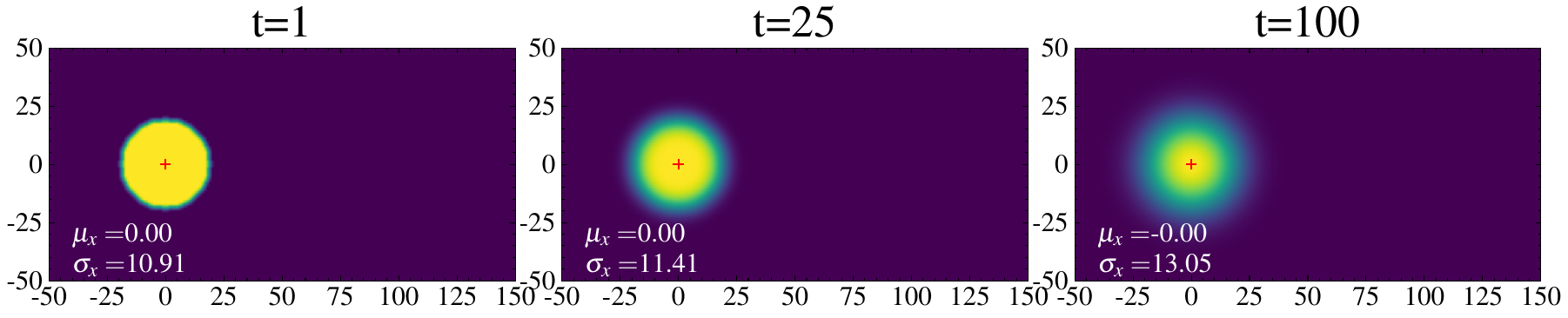}
        \caption{$\Sigma$ $(\beta^2=0)$}
        \label{fig:circle_relu_uni_3x3_b0}
      \end{subfigure}\\
      \begin{subfigure}{\linewidth}
        \includegraphics[width=\linewidth]{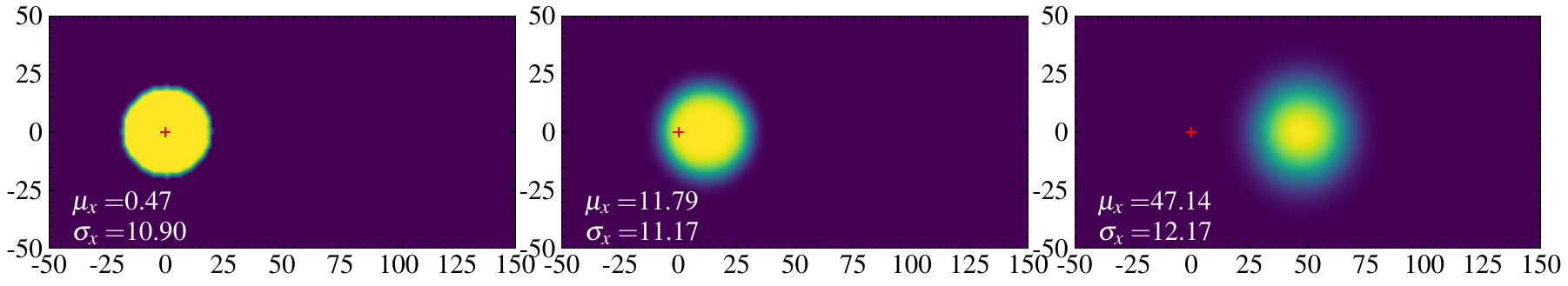}
        \caption{$\Sigma+\nabla_x$ $(\beta^2=0.25)$}
        \label{fig:circle_relu_uni_3x3_b0.25}
      \end{subfigure}\\
      \begin{subfigure}{\linewidth}
        \includegraphics[width=\linewidth]{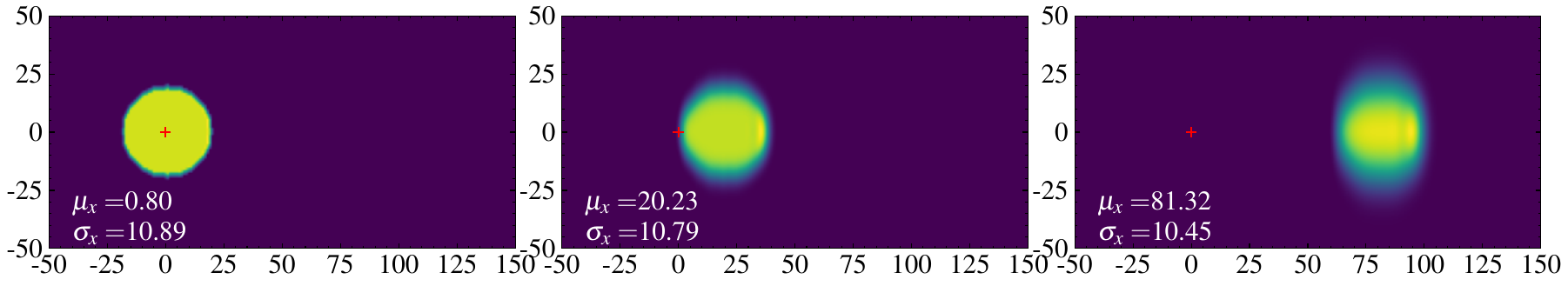}
        \caption{$\Sigma+\nabla_x$ $(\beta^2=0.5)$}
        \label{fig:circle_relu_uni_3x3_b0.5}
      \end{subfigure}\\
      \begin{subfigure}{\linewidth}
        \includegraphics[width=\linewidth]{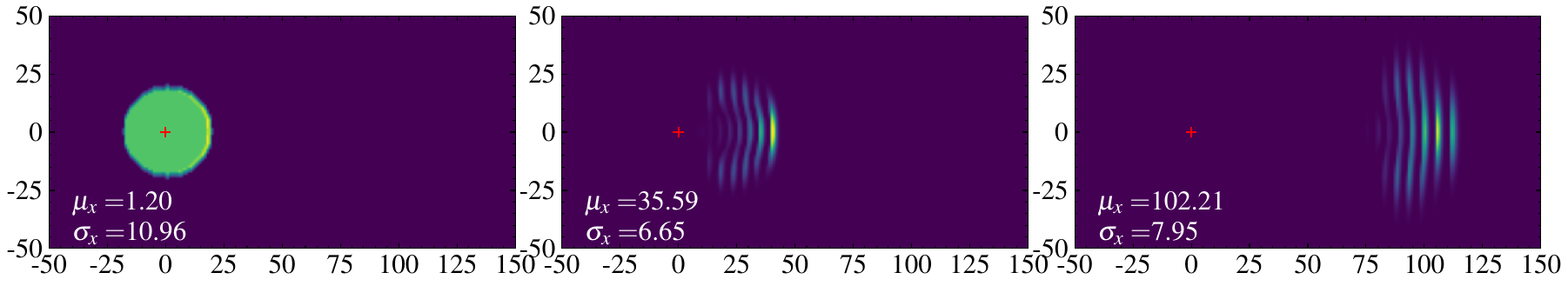}
        \caption{$\Sigma+\nabla_x$ $(\beta^2=0.75)$}
        \label{fig:circle_relu_uni_3x3_b0.75}
      \end{subfigure}\\
      \begin{subfigure}{\linewidth}
        \includegraphics[width=\linewidth]{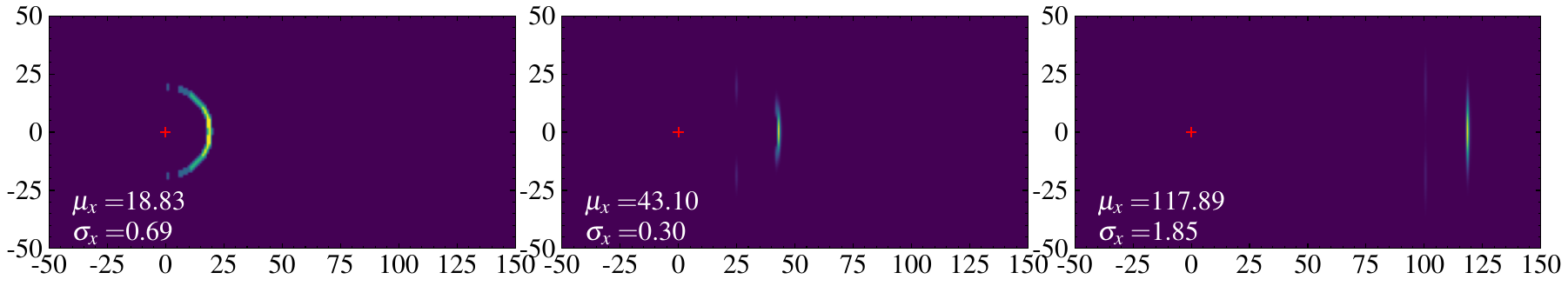}
        \caption{$\nabla_x$ $(\beta^2=1)$}
        \label{fig:circle_relu_uni_3x3_b1}
      \end{subfigure}
    \end{minipage}%
  
  \caption{Demonstrating the effect of repeated convolution+ReLU of a circular test pattern ($r=19$) with $3\times3$ kernels mixing DC $\Sigma$ and fixed direction gradient $\nabla_x$ over various mixing ratios $\beta \in \{0,0.25,0.5,0.75,1\}$.  Note that for $\beta=0$ a), content diffuses symmetrically about a stationary centre of mass, while for $\beta=1$, the circle bulk disappears and the rightmost edge translates right with maximum velocity.}
  \label{fig:circle_relu_uni_3x3}
\end{figure}

\clearpage
\pagebreak

\subsubsection{$3\times 3$ kernel, translation filter components}

\begin{figure}[htbp]
  \centering
    \begin{minipage}{0.68\columnwidth}
      \centering
      \begin{subfigure}{\linewidth}
        \includegraphics[width=\linewidth]{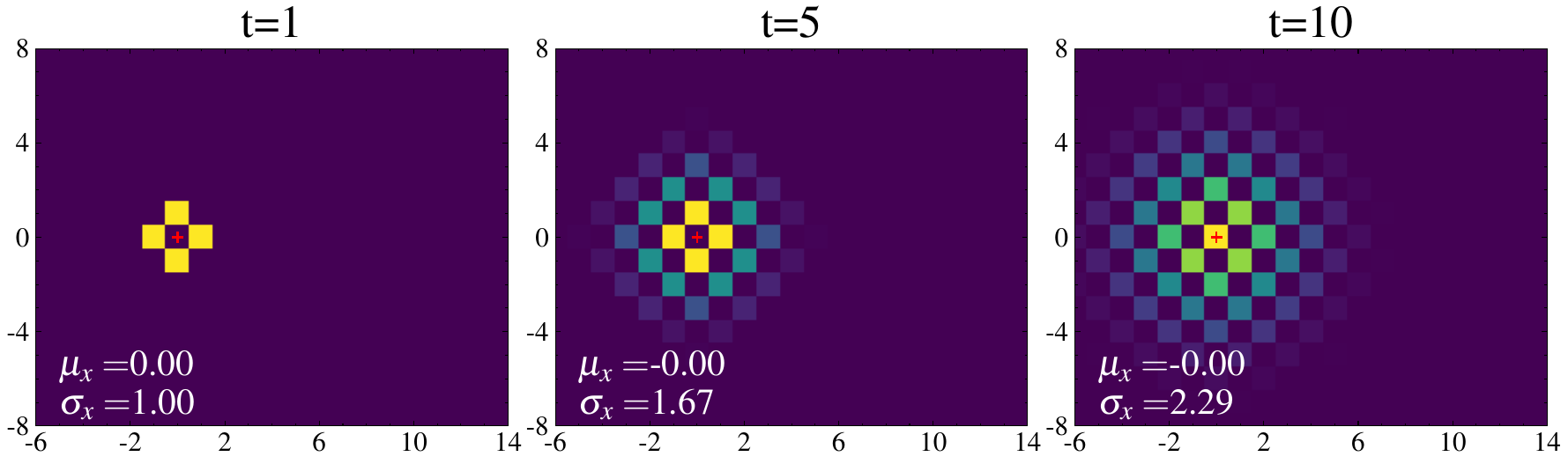}
        \caption{$f_e$ $(\beta^2=0)$}
        \label{fig:pixel_trans_relu_uni_3x3_b0}
      \end{subfigure}\\
      \begin{subfigure}{\linewidth}
        \includegraphics[width=\linewidth]{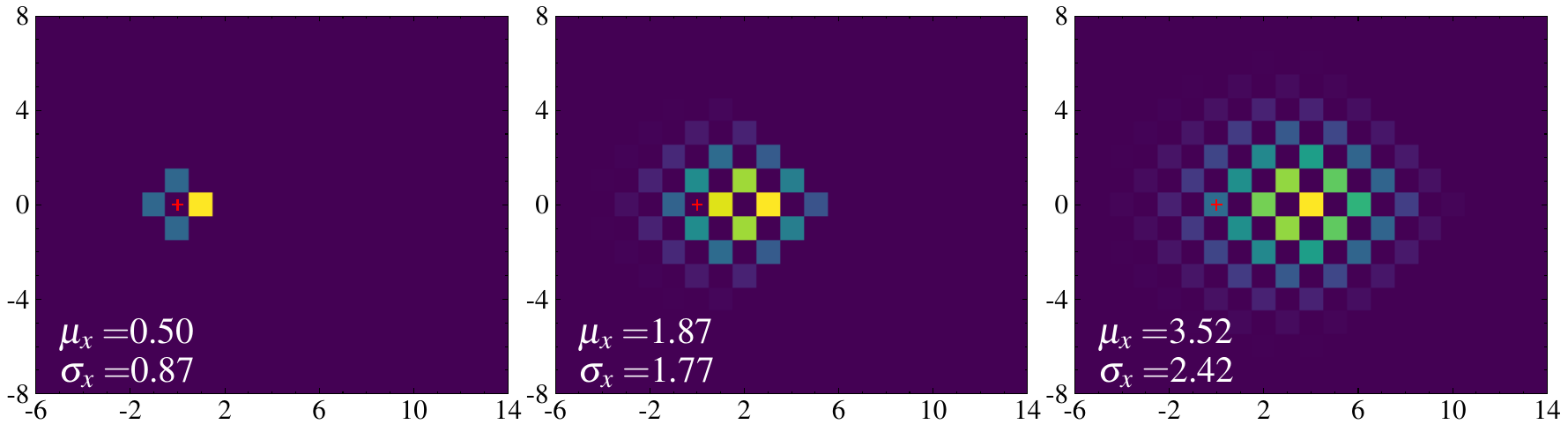}
        \caption{$f_e+f_o$ $(\beta^2=0.25)$}
        \label{fig:pixel_trans_relu_uni_3x3_b0.25}
      \end{subfigure}\\
      \begin{subfigure}{\linewidth}
        \includegraphics[width=\linewidth]{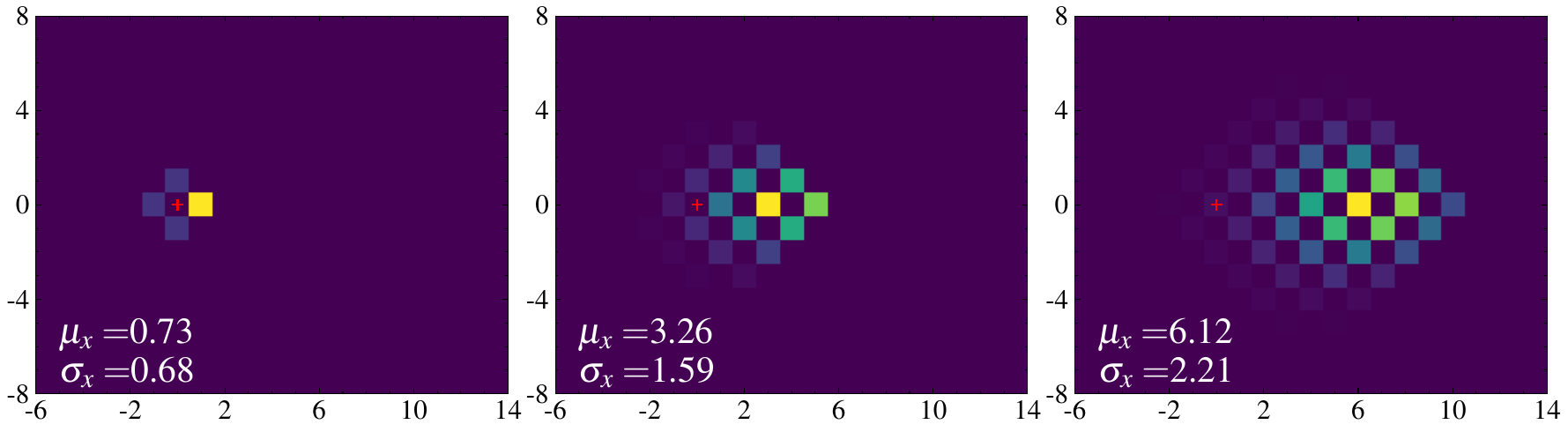}
        \caption{$f_e+f_o$ $\beta^2=0.5$}
        \label{fig:pixel_trans_relu_uni_3x3_b0.5}
      \end{subfigure}\\
      \begin{subfigure}{\linewidth}
        \includegraphics[width=\linewidth]{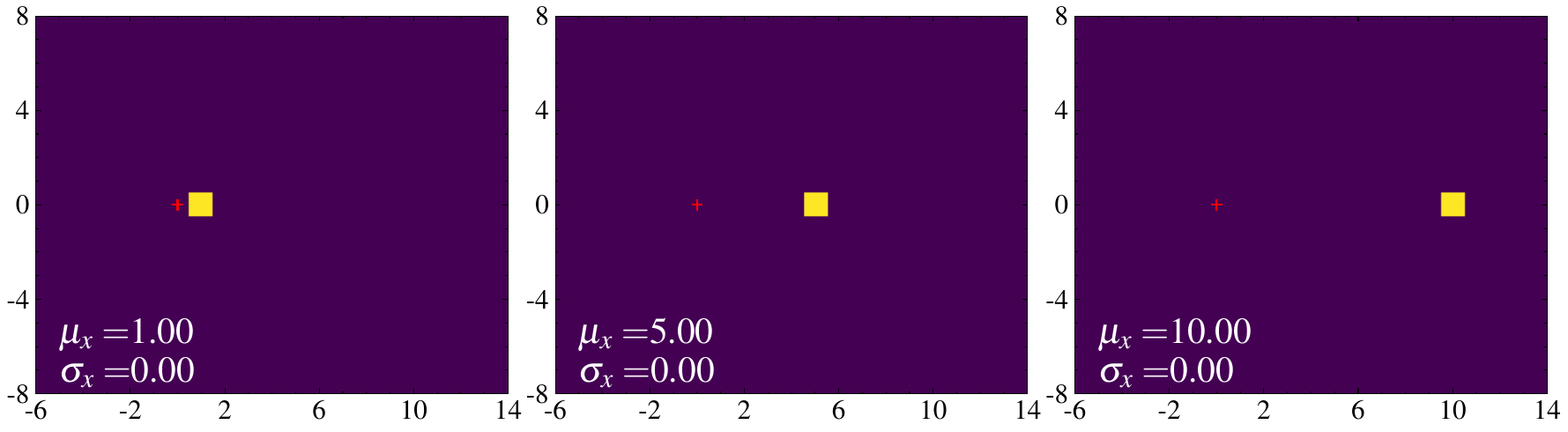}
        \caption{$f_e+f_o$ $\beta^2=0.75$}
        \label{fig:pixel_trans_relu_uni_3x3_b0.75}
      \end{subfigure}\\
      \begin{subfigure}{\linewidth}
        \includegraphics[width=\linewidth]{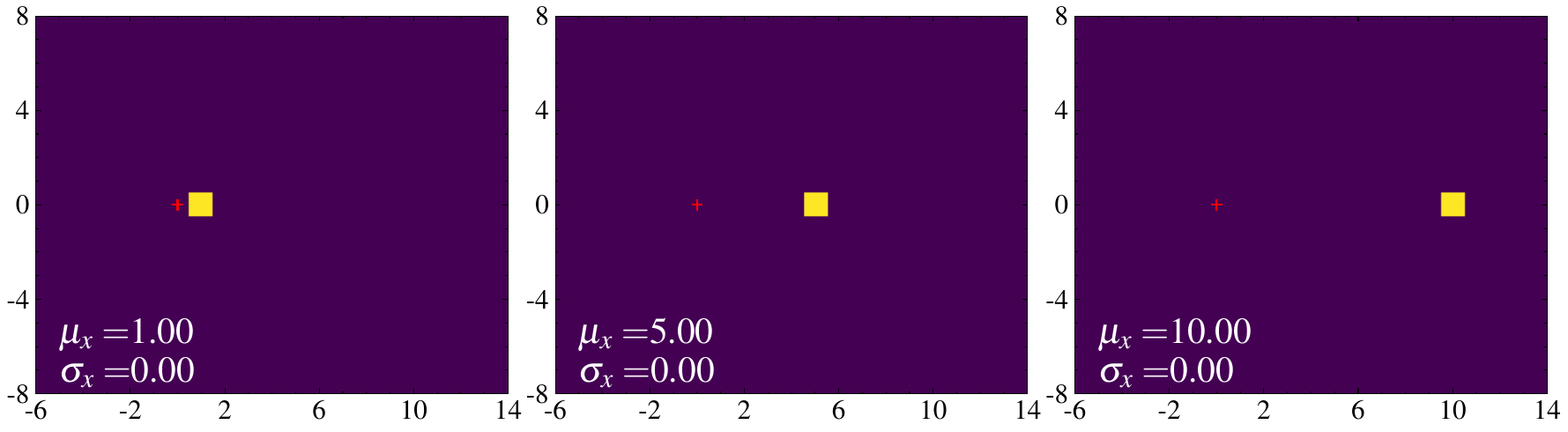}
        \caption{$f_o \  (\beta^2=1)$}
        \label{fig:pixel_trans_relu_uni_3x3_b1}
      \end{subfigure}
    \end{minipage}%

  \caption{Demonstrating the effect of repeated convolution+ReLU of a test pattern (pixel) over $3\times3$ translations, varying $\beta$. Note that for $\beta=0$ a), artificial checker-board structure  appears due to the complex non-DC even (symmetric) component, while the content maintains a stationary centre of mass. For $\beta=1$, the content translates rightward with maximum velocity. }
  \label{fig:pixel_trans_relu_uni_3x3}
\end{figure}

\begin{figure}[htbp!]
  \centering
    \begin{minipage}{0.7\columnwidth}
      \centering
      \begin{subfigure}{\linewidth}
        \includegraphics[width=\linewidth]{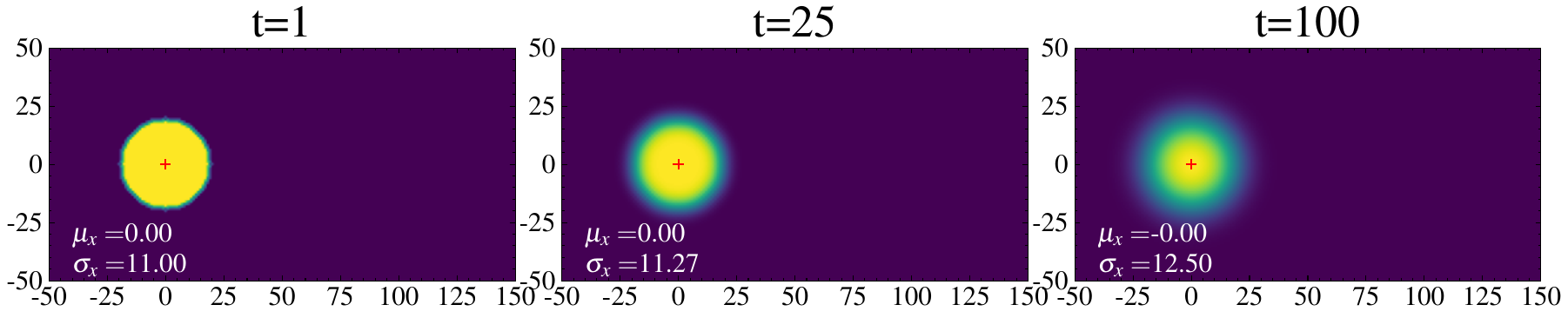}
        \caption{$\Sigma$ $\beta^2=0$}
        \label{fig:circle_trans_relu_uni_3x3_b0}
      \end{subfigure}\\
      \begin{subfigure}{\linewidth}
        \includegraphics[width=\linewidth]{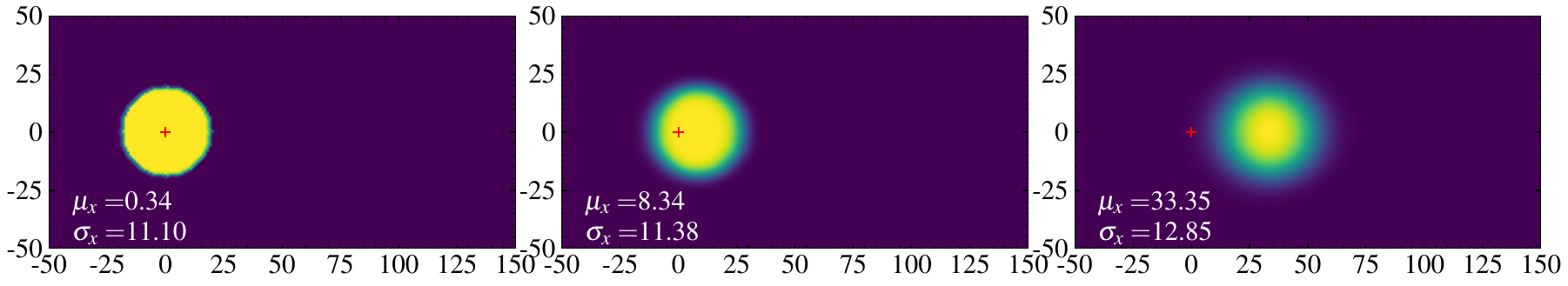}
        \caption{$\Sigma+\nabla_x$ $\beta^2=0.25$}
        \label{fig:circle_trans_relu_uni_3x3_b0.25}
      \end{subfigure}\\
      \begin{subfigure}{\linewidth}
        \includegraphics[width=\linewidth]{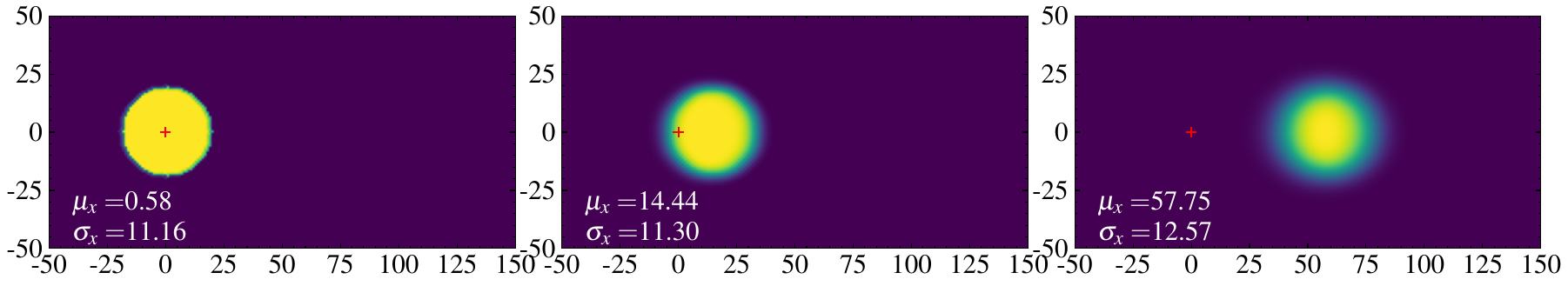}
        \caption{$\Sigma+\nabla_x$ $\beta^2=0.5$}
        \label{fig:circle_trans_relu_uni_3x3_b0.5}
      \end{subfigure}\\
      \begin{subfigure}{\linewidth}
        \includegraphics[width=\linewidth]{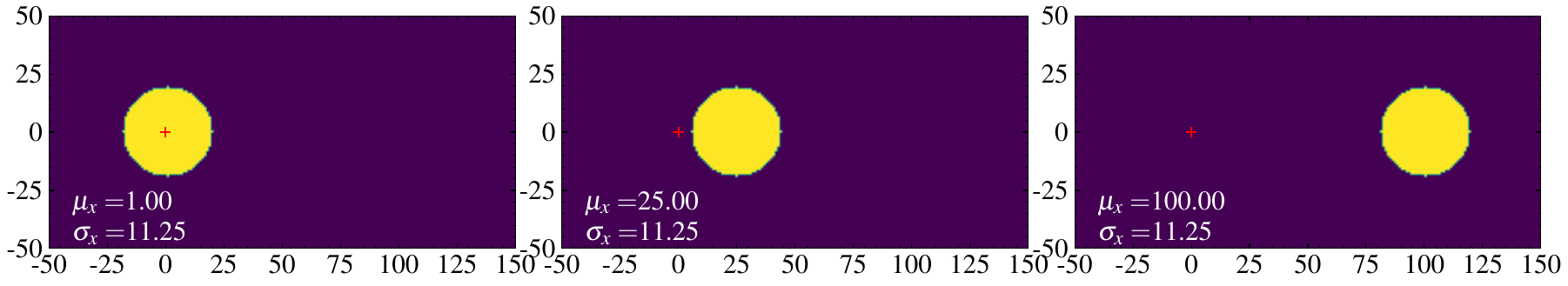}
        \caption{$\Sigma+\nabla_x$ $\beta^2=0.75$}
        \label{fig:circle_trans_relu_uni_3x3_b0.75}
      \end{subfigure}\\
      \begin{subfigure}{\linewidth}
        \includegraphics[width=\linewidth]{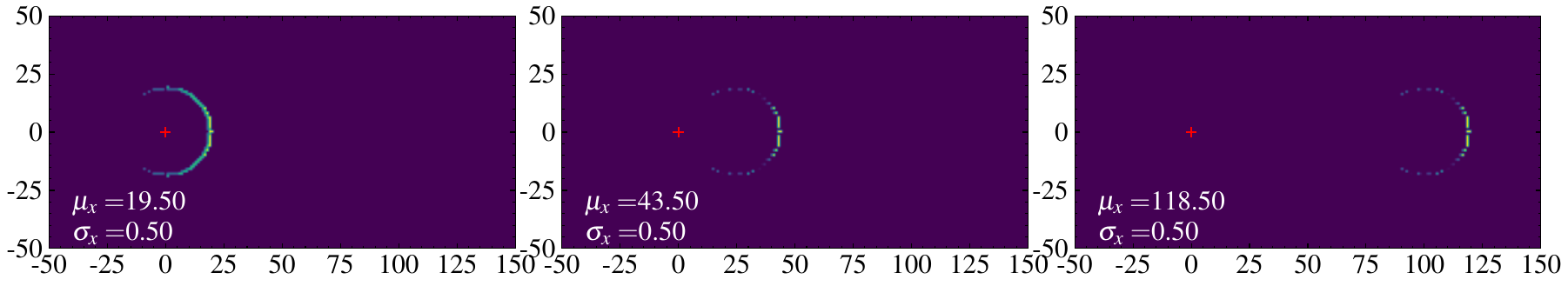}
        \caption{$\nabla_x$ $\beta^2=1$}
        \label{fig:circle_trans_relu_uni_3x3_b1}
      \end{subfigure}
    \end{minipage}%
  \caption{Demonstrating the effect of repeated convolution+ReLU of a test pattern (circle) over $3\times3$ translation filter components, varying $\beta$. Note that for $\beta=1$, the circle bulk disappears and the rightmost edge translates rightward with maximum velocity.}
  \label{fig:circle_trans_relu_uni_3x3}
\end{figure}

\clearpage
\pagebreak
\subsubsection{$3\times 3$ kernel, mixing alternating gradient $\{ \nabla_x, -\nabla_x \}$ and sum $\Sigma$ components}

\begin{figure}[htbp]
  \centering
    \begin{minipage}{0.7\columnwidth}
      \centering
      \begin{subfigure}{\linewidth}
        \includegraphics[width=\linewidth]{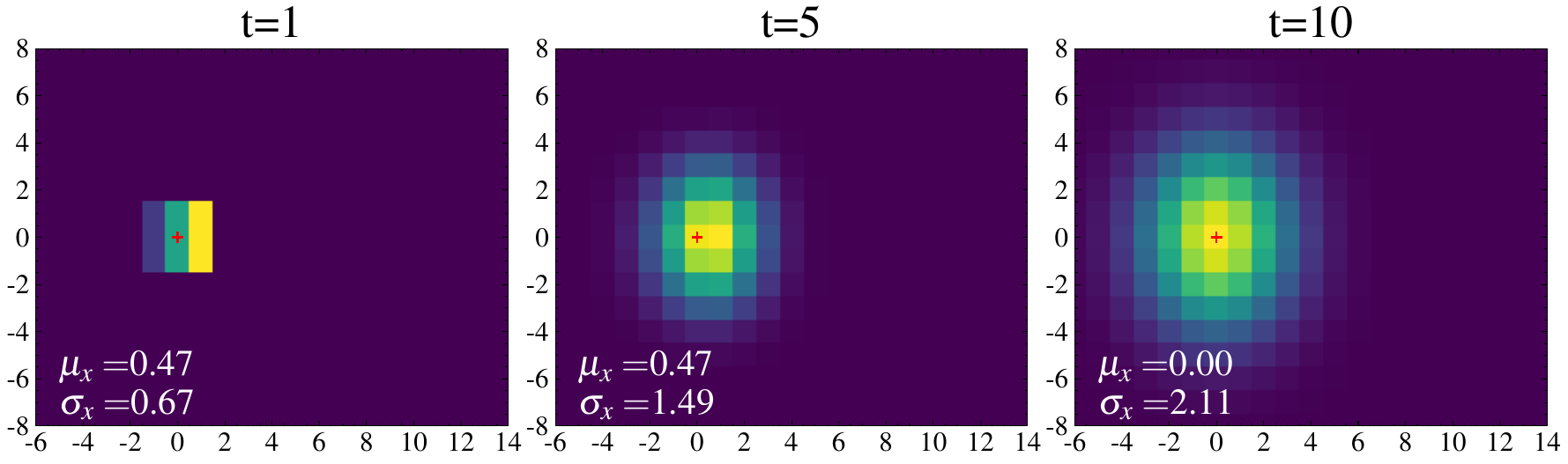}
        \caption{$\Sigma \pm \nabla_x$ $(\beta^2=0.25)$}
        \label{fig:pixel_relu_bi_3x3_b0.25}
      \end{subfigure}\\
      \begin{subfigure}{\linewidth}
        \includegraphics[width=\linewidth]{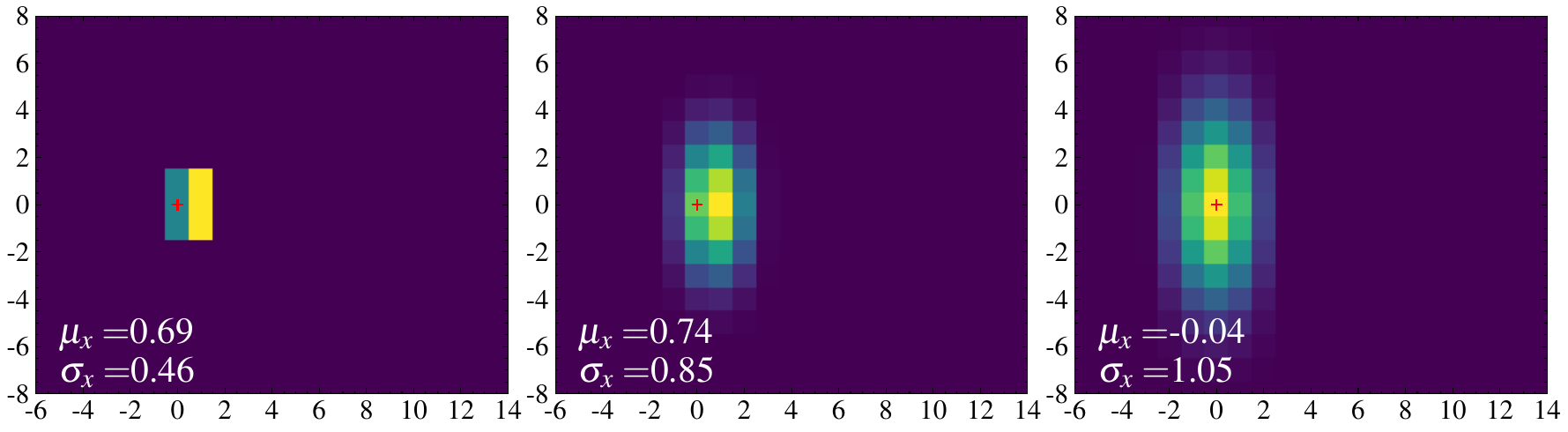}
        \caption{$\Sigma \pm \nabla_x$ $(\beta^2=0.5)$}
        \label{fig:pixel_relu_bi_3x3_b0.5}
      \end{subfigure}\\
      \begin{subfigure}{\linewidth}
        \includegraphics[width=\linewidth]{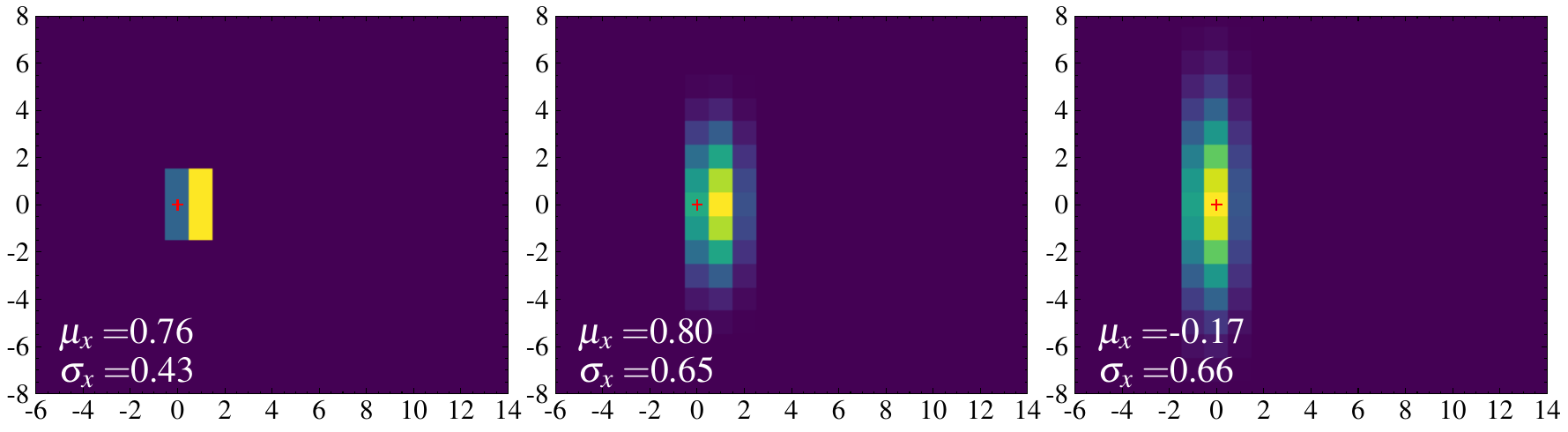}
        \caption{$\Sigma \pm \nabla_x$ $(\beta^2=0.75)$}
        \label{fig:pixel_relu_bi_3x3_b0.75}
      \end{subfigure}\\
      \begin{subfigure}{\linewidth}
        \includegraphics[width=\linewidth]{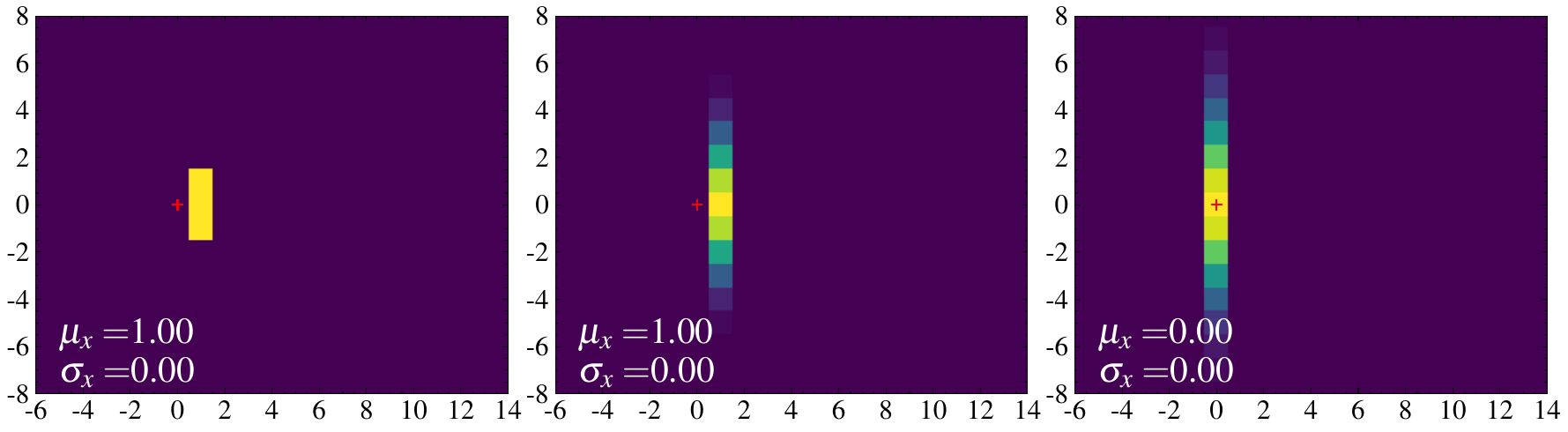}
        \caption{$\nabla_x$ $(\beta^2=1)$}
        \label{fig:pixel_relu_bi_3x3_b1}
      \end{subfigure}
    \end{minipage}%
  \caption{Demonstrating the effect of repeated convolution+ReLU with alternating orientation. Note that for $\beta^1=1$ in d), information vibrates left and right, and there is no net translation of the centre of mass.}
  \label{fig:pixel_relu_bi_3x3}
\end{figure}

\begin{figure}[htbp]
  \centering
    \begin{minipage}{0.7\columnwidth}
      \centering
      \begin{subfigure}{\linewidth}
        \includegraphics[width=\linewidth]{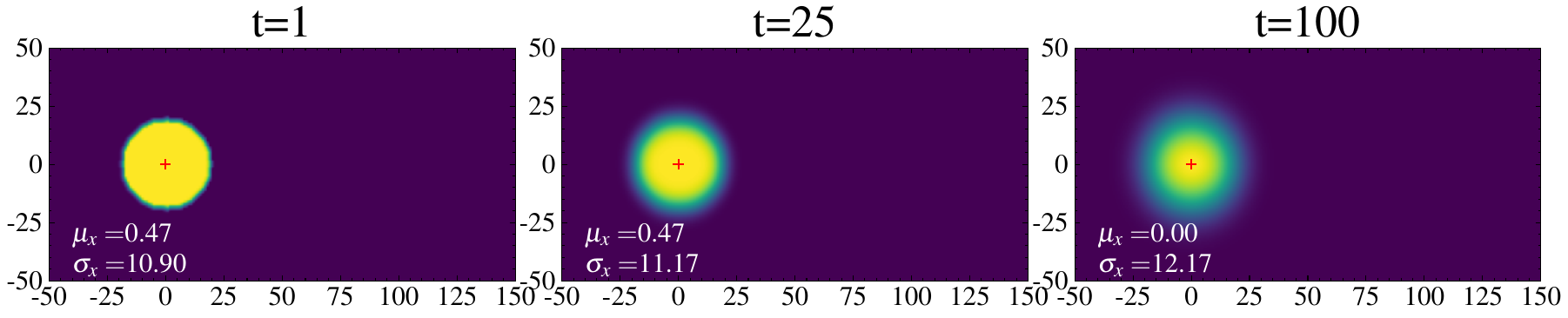}
        \caption{$\Sigma\pm\nabla_x$ $(\beta^2=0.25)$}
        \label{fig:circle_relu_bi_3x3_b0.25}
      \end{subfigure}
      \begin{subfigure}{\linewidth}
        \includegraphics[width=\linewidth]{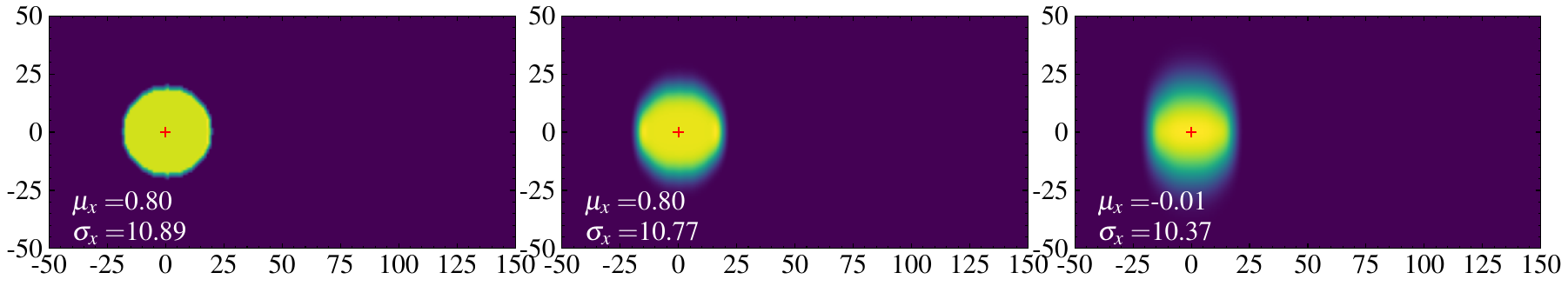}
        \caption{$\Sigma\pm\nabla_x$ $(\beta^2=0.5)$}
        \label{fig:circle_relu_bi_3x3_b0.5}
      \end{subfigure}
      \begin{subfigure}{\linewidth}
        \includegraphics[width=\linewidth]{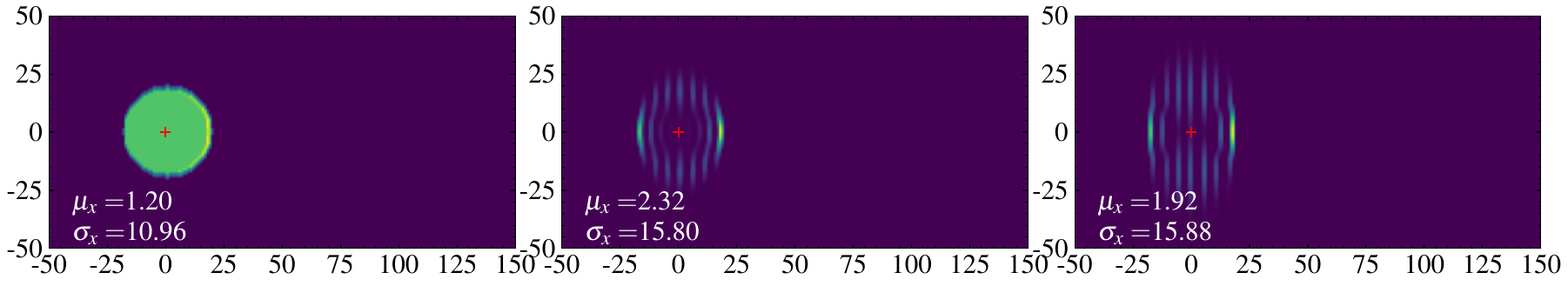}
        \caption{$\Sigma\pm\nabla_x$ $(\beta^2=0.75)$}
        \label{fig:circle_relu_bi_3x3_b0.75}
      \end{subfigure}
      \begin{subfigure}{\linewidth}
        \includegraphics[width=\linewidth]{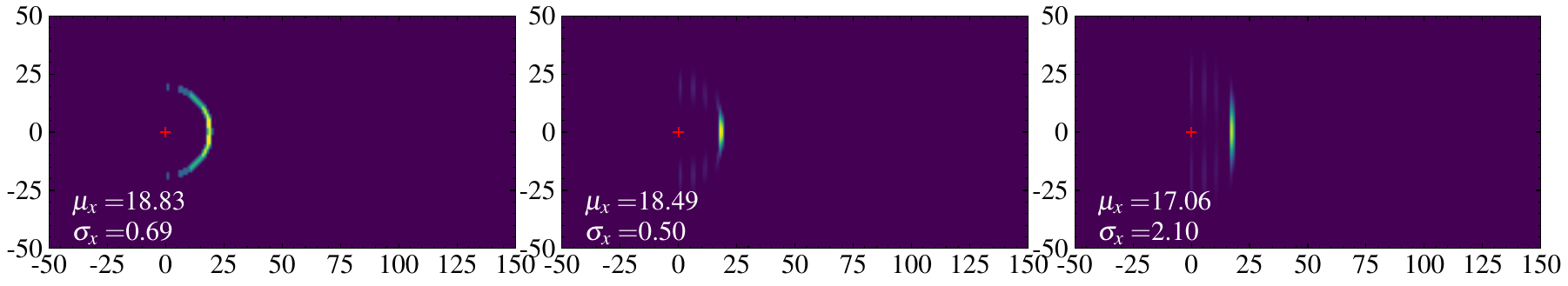}
        \caption{$\pm\nabla_x$ $(\beta^2=1)$}
        \label{fig:circle_relu_bi_3x3_b1}
      \end{subfigure}
    \end{minipage}%
  
  \caption{Demonstrating the effect of repeated convolution+ReLU of a circular test pattern over different types of $3\times3$ kernels mixing DC $\Sigma$ and alternating direction gradient $\pm\nabla_x$ components for different mixing ratios. Note that for $\beta^2=1$, the circle bulk disappears, and the right edge of the circle vibrates left to right with no net translation.}
  \label{fig:circle_relu_bi_3x3}
\end{figure}

\clearpage
\pagebreak
\subsubsection{$2\times 2$ kernel, mixing unidirectional gradient $\nabla_x$ and sum $\Sigma$ components}

\begin{figure}[htbp]
  \centering
    \begin{minipage}{0.7\columnwidth}
      \centering
      \begin{subfigure}{\linewidth}
        \includegraphics[width=\linewidth]{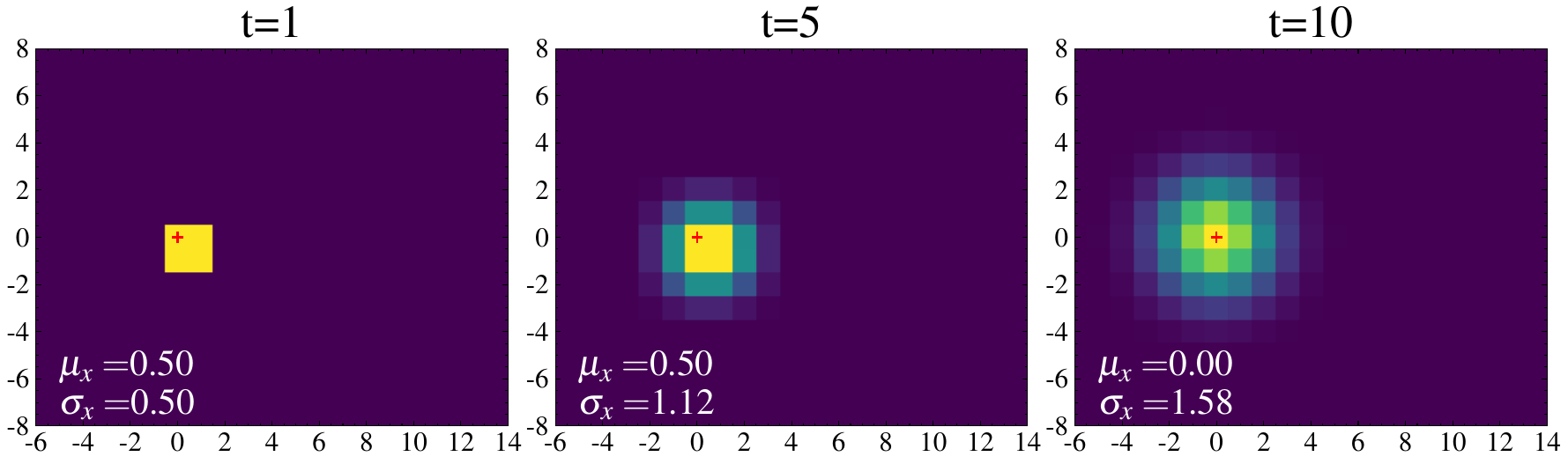}
        \caption{$\Sigma$ $(\beta^2=0)$}
        \label{fig:pixel_relu_uni_2x2_b0}
      \end{subfigure}\\
      \begin{subfigure}{\linewidth}
        \includegraphics[width=\linewidth]{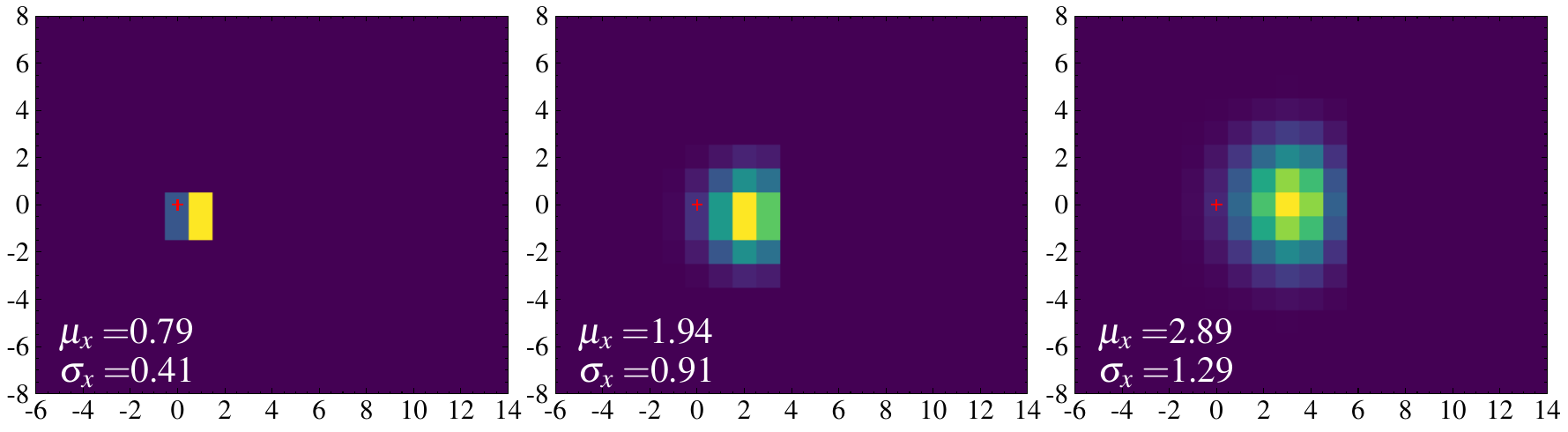}
        \caption{$\Sigma + \nabla_x$ $(\beta^2=0.25)$}
        \label{fig:pixel_relu_uni_2x2_b0.25}
      \end{subfigure}\\
      \begin{subfigure}{\linewidth}
        \includegraphics[width=\linewidth]{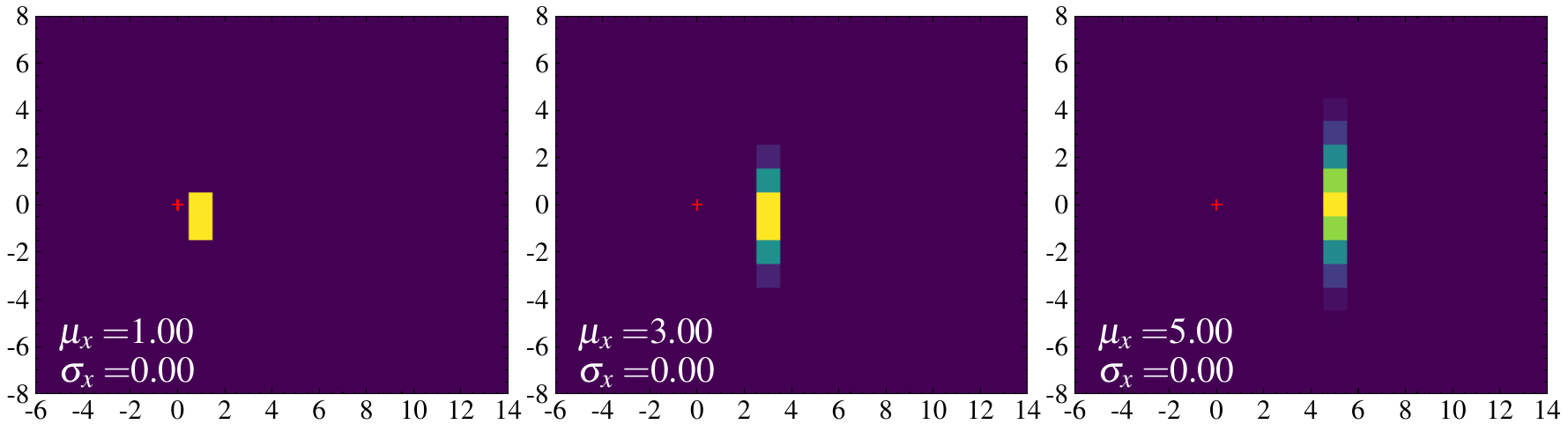}
        \caption{$\Sigma + \nabla_x$ $\beta^2=0.5$}
        \label{fig:pixel_relu_uni_2x2_b0.5}
      \end{subfigure}\\
      \begin{subfigure}{\linewidth}
        \includegraphics[width=\linewidth]{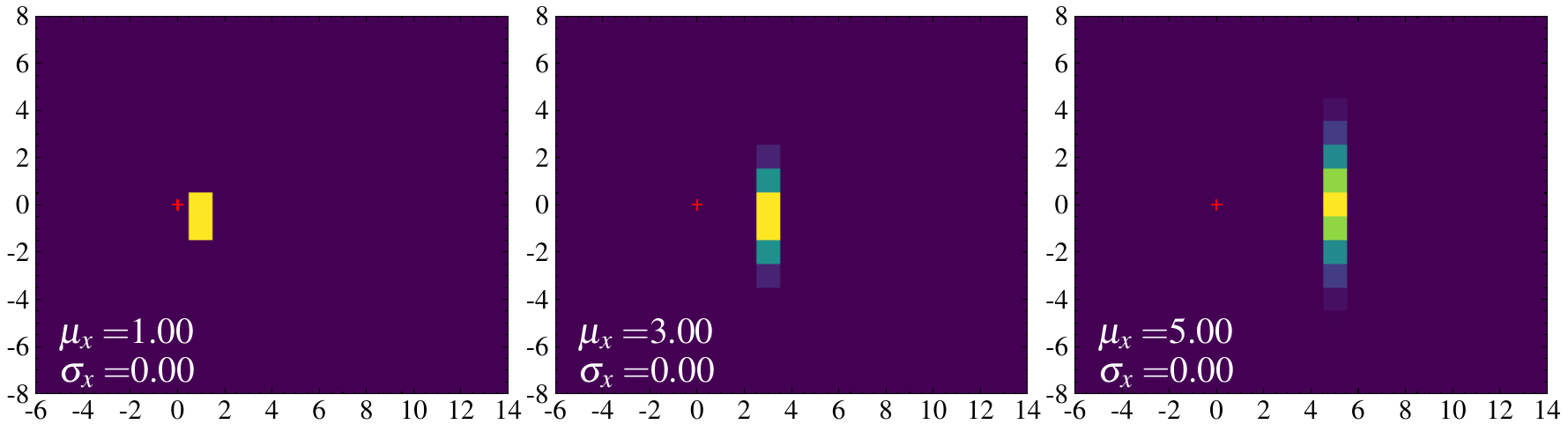}
        \caption{$\Sigma + \nabla_x$ $\beta^2=0.75$}
        \label{fig:pixel_relu_uni_2x2_b0.75}
      \end{subfigure}\\
      \begin{subfigure}{\linewidth}
        \includegraphics[width=\linewidth]{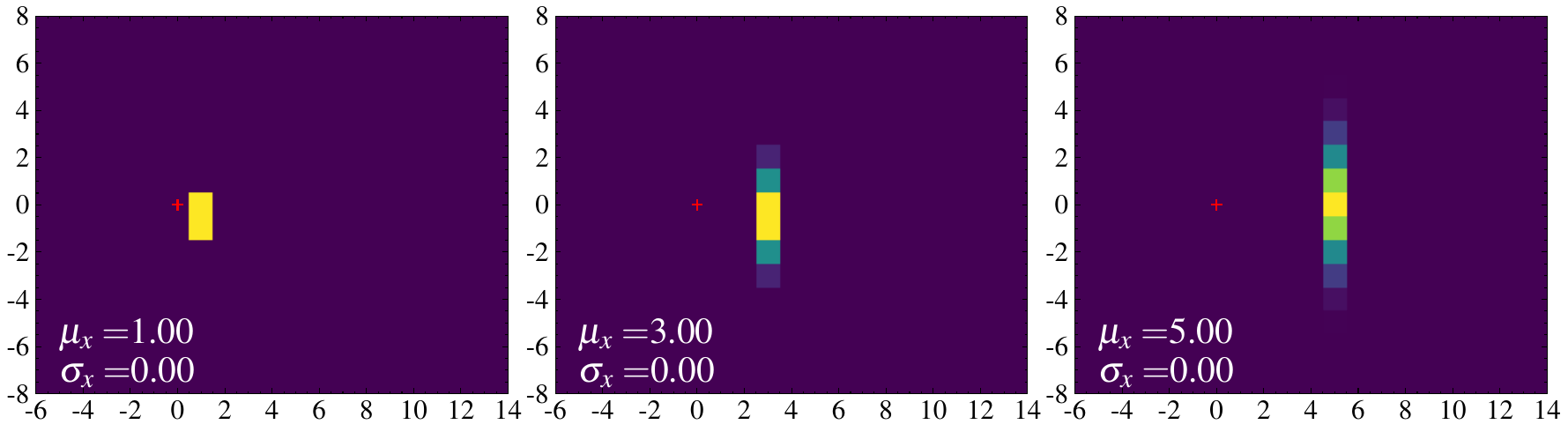}
        \caption{$\nabla_x$ $(\beta^2=1)$}
        \label{fig:pixel_relu_uni_2x2_b1}
      \end{subfigure}
    \end{minipage}%
  
  \caption{Demonstrating the effect of repeated convolution+ReLU of a test pattern over different types of $2\times2$ kernels (DC and Gradient).  Note that for $\beta^2 \ge 0.5 $, the content centre of mass travels rightward with maximum velocity.}
  \label{fig:pixel_relu_uni_2x2}
\end{figure}

\begin{figure}[h!]
  \centering
    \begin{minipage}{0.6\columnwidth}
      \centering
      \begin{subfigure}{\linewidth}
        \includegraphics[width=\linewidth]{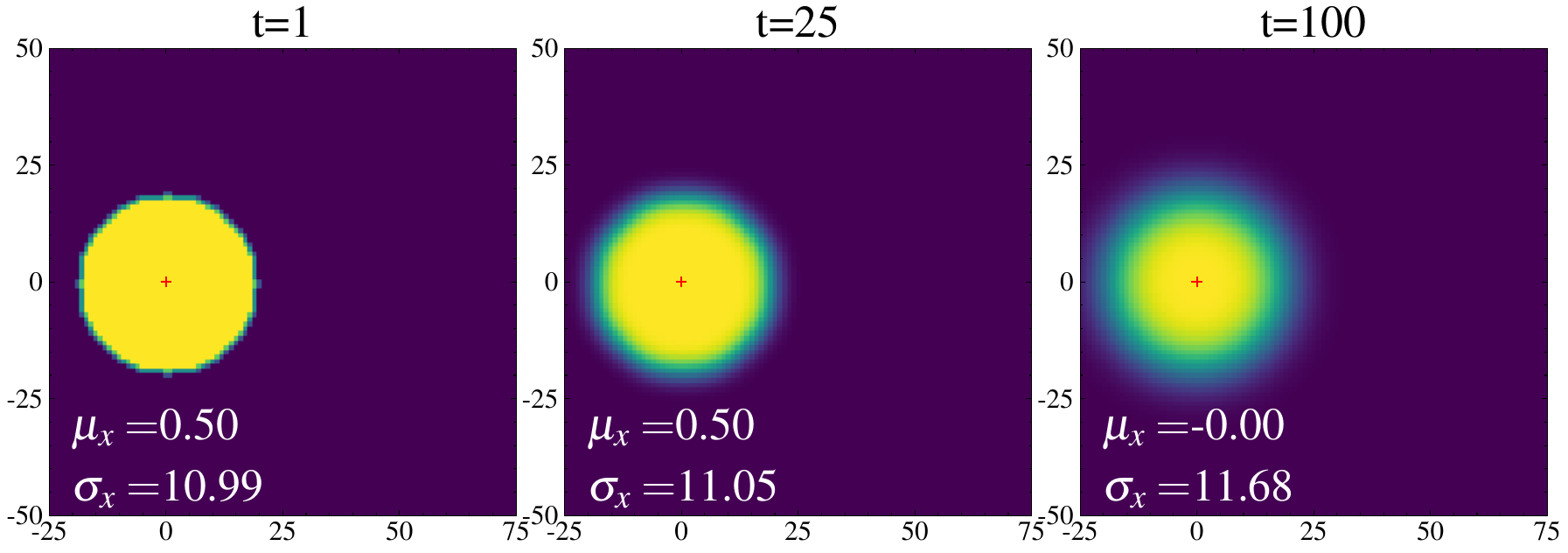}
        \caption{$\Sigma$ $(\beta^2=0)$}
        \label{fig:circle_relu_uni_2x2_b0}
      \end{subfigure}\\
      \begin{subfigure}{\linewidth}
        \includegraphics[width=\linewidth]{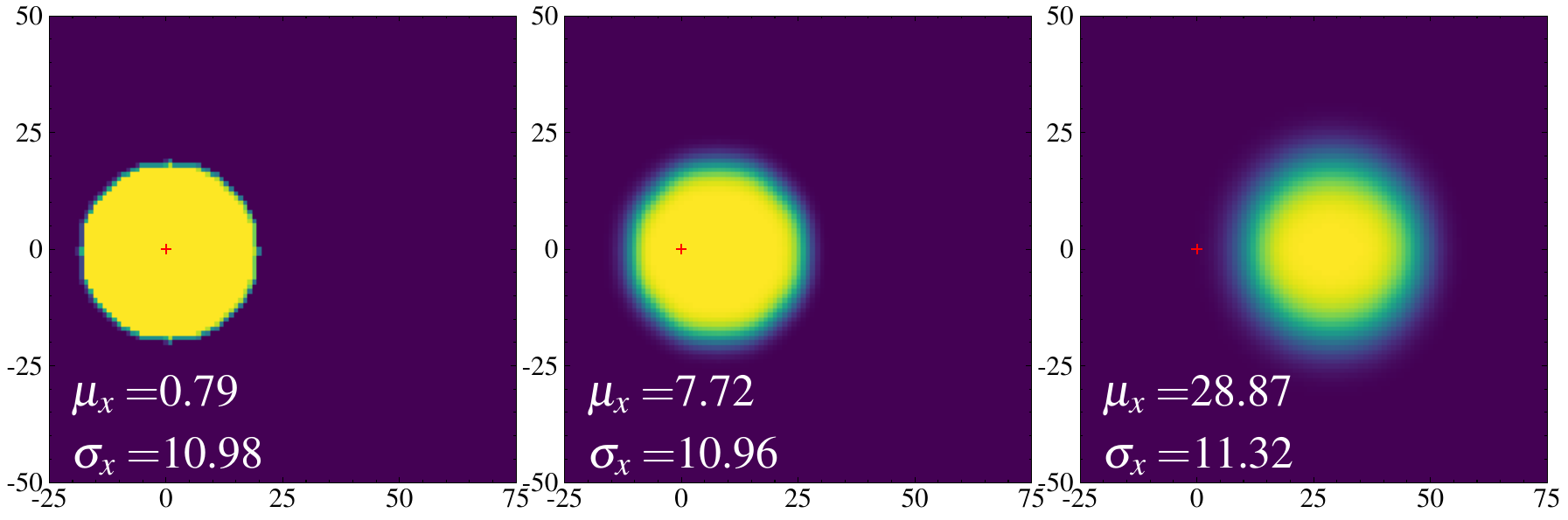}
        \caption{$\Sigma+\nabla_x$ $(\beta^2=0.25)$}
        \label{fig:circle_relu_uni_2x2_b0.25}
      \end{subfigure}\\
      \begin{subfigure}{\linewidth}
        \includegraphics[width=\linewidth]{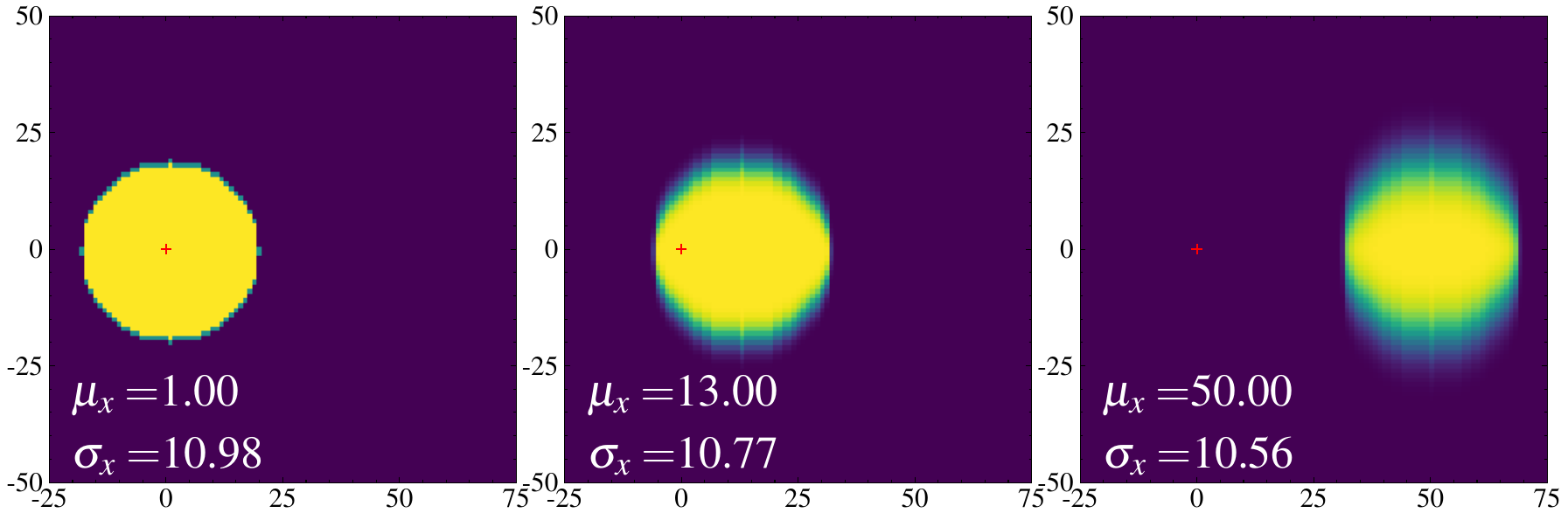}
        \caption{$\Sigma + \nabla_x$ $(\beta^2=0.5)$}
        \label{fig:circle_relu_uni_2x2_b0.5}
      \end{subfigure}\\
      \begin{subfigure}{\linewidth}
        \includegraphics[width=\linewidth]{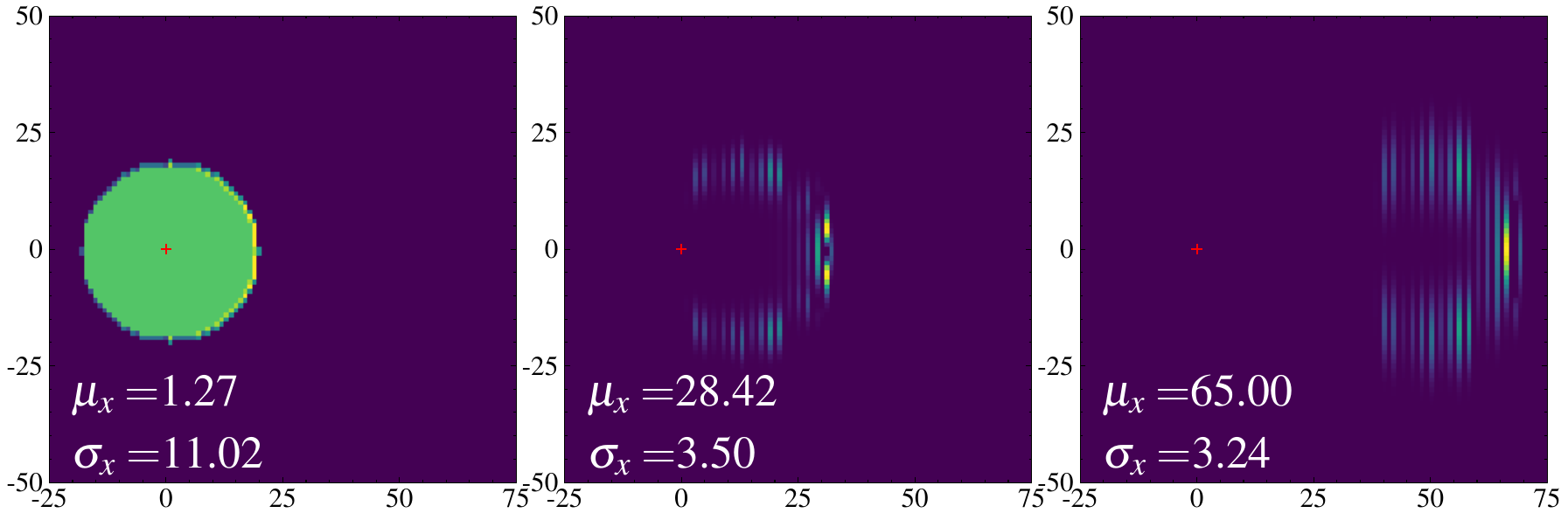}
        \caption{$\Sigma + \nabla_x$ $(\beta^2=0.75)$}
        \label{fig:circle_relu_uni_2x2_b0.75}
      \end{subfigure}\\
      \begin{subfigure}{\linewidth}
        \includegraphics[width=\linewidth]{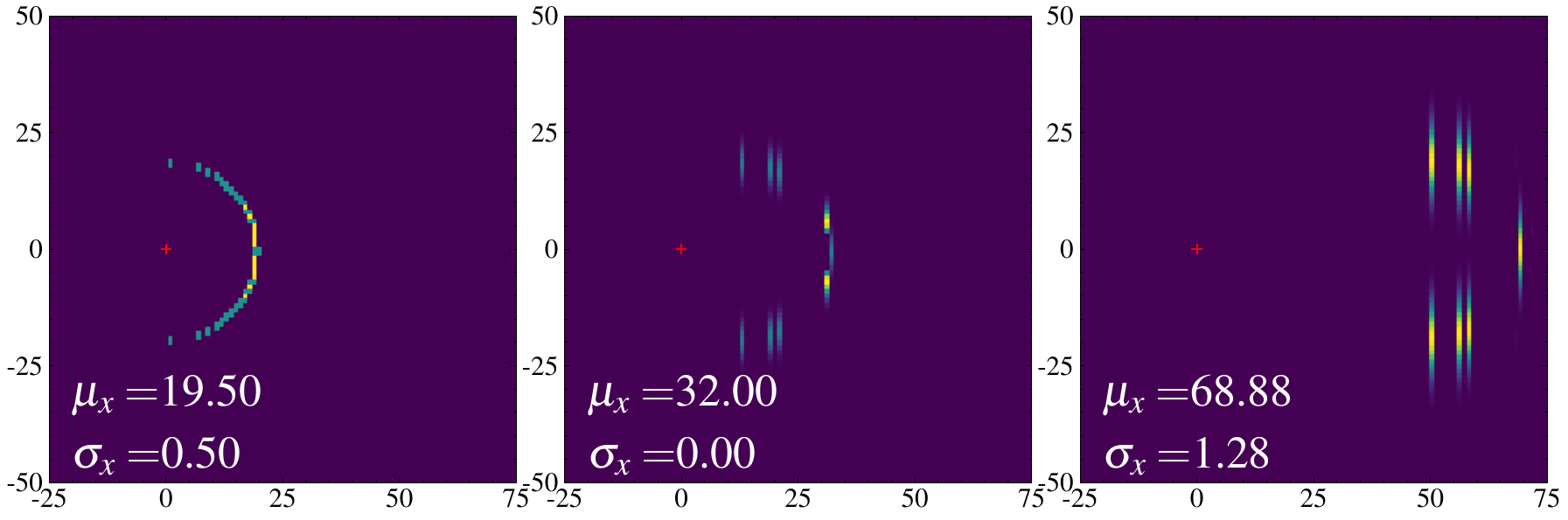}
        \caption{$\nabla_x$ $(\beta^2=1)$}
        \label{fig:circle_relu_uni_2x2_b1}
      \end{subfigure}
    \end{minipage}%

  \caption{Demonstrating the effect of repeated convolution+ReLU of a circular test pattern ($r=19$) with $2\times2$ kernels mixing DC $\Sigma$ and fixed direction gradient $\nabla_x$ components over various mixing ratios $\beta$.  Note that for $\beta=0$, the content diffuses symmetrically with a stationary centre of mass, while for $\beta=1$ the circle bulk disappears and the right edge of the circle travels rightward with maximum velocity.}
  \label{fig:circle_relu_uni_2x2}
\end{figure}

\clearpage
\pagebreak
\subsubsection{$3\times 3$ kernel mixing unidirectional gradient $\nabla_x$ and sum $\Sigma$ components, Modulus (Absolute value) activation.}

\begin{figure}[h!]
  \centering
    \begin{minipage}{0.6\columnwidth}
      \centering
      \begin{subfigure}{\linewidth}
        \includegraphics[width=\linewidth]{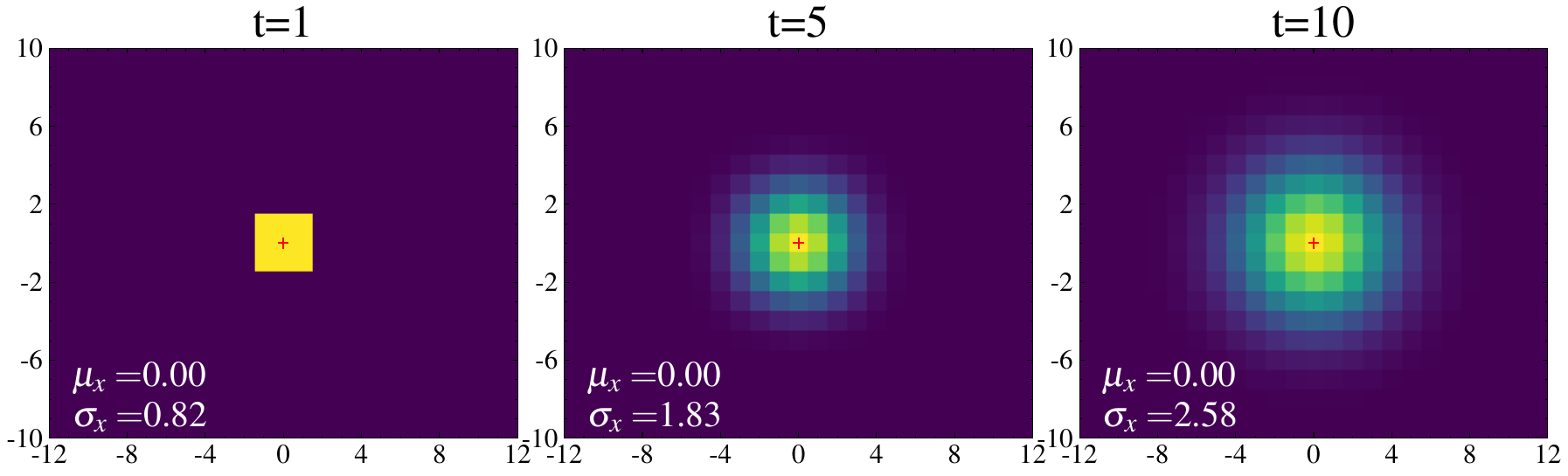}
        \caption{$\Sigma$ $(\beta^2=0)$}
        \label{fig:circle_mod_uni_3x3_b0}
      \end{subfigure}\\
      \begin{subfigure}{\linewidth}
        \includegraphics[width=\linewidth]{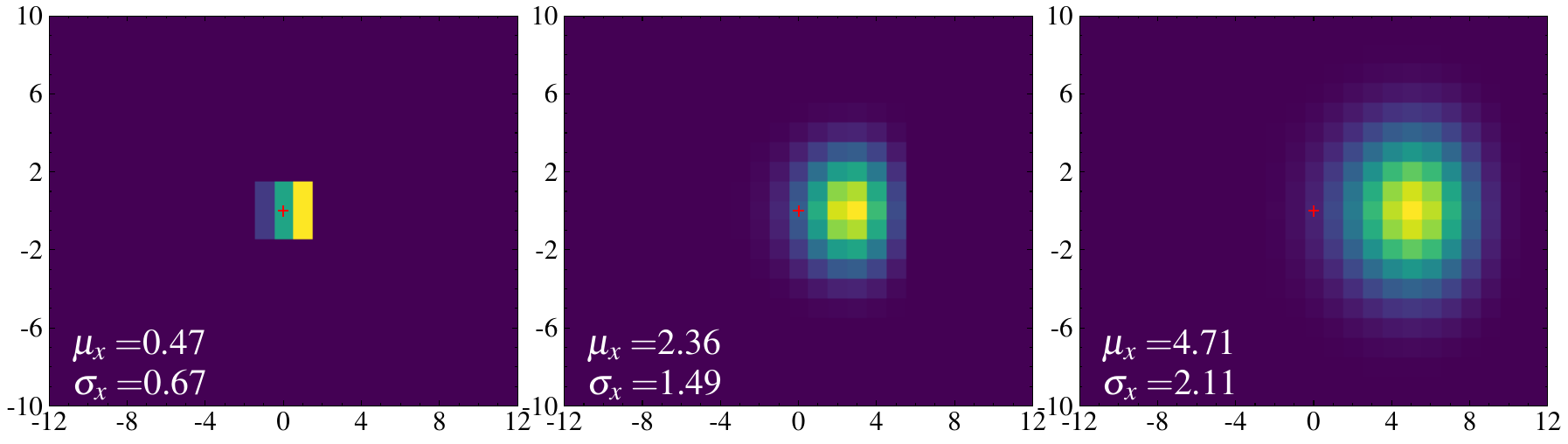}
        \caption{$\Sigma+\nabla_x$ $(\beta^2=0.25)$}
        \label{fig:circle_mod_uni_3x3_b0.25}
      \end{subfigure}\\
      \begin{subfigure}{\linewidth}
        \includegraphics[width=\linewidth]{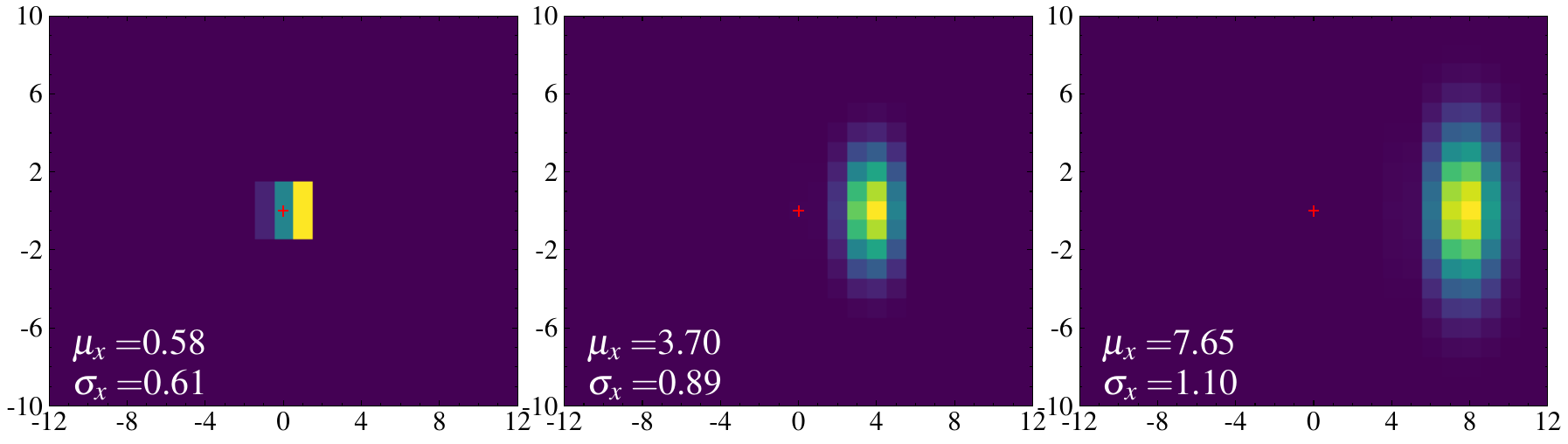}
        \caption{$\Sigma + \nabla_x$ $(\beta^2=0.5)$}
        \label{fig:circle_mod_uni_3x3_b1}
      \end{subfigure}
      \begin{subfigure}{\linewidth}
        \includegraphics[width=\linewidth]{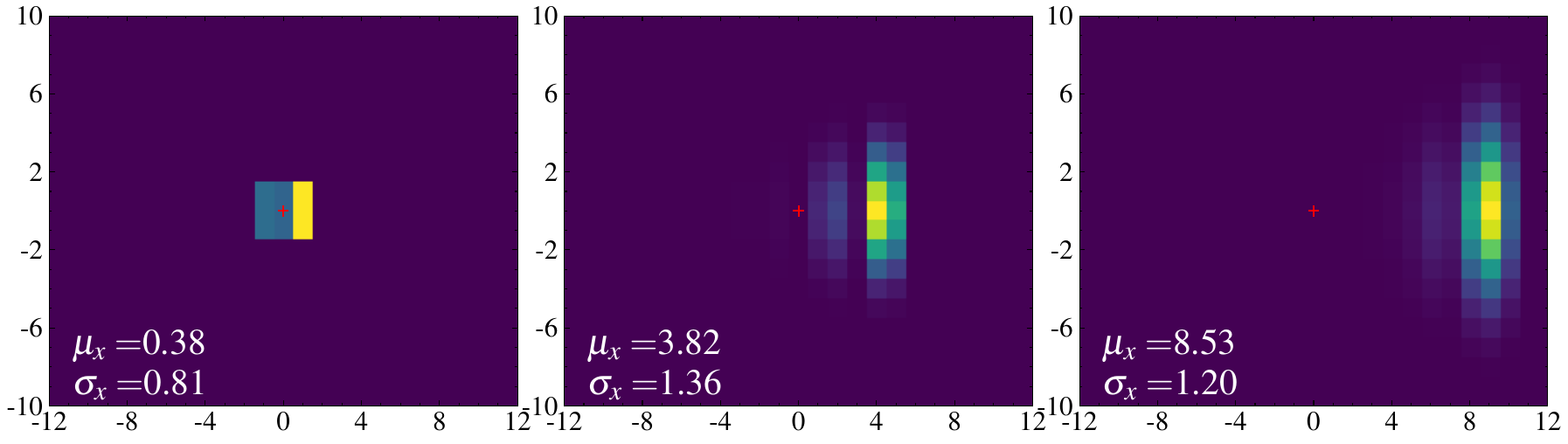}
        \caption{$\Sigma + \nabla_x$ $(\beta^2=0.75)$}
        \label{fig:circle_mod_uni_3x3_b1}
      \end{subfigure}
      \begin{subfigure}{\linewidth}
        \includegraphics[width=\linewidth]{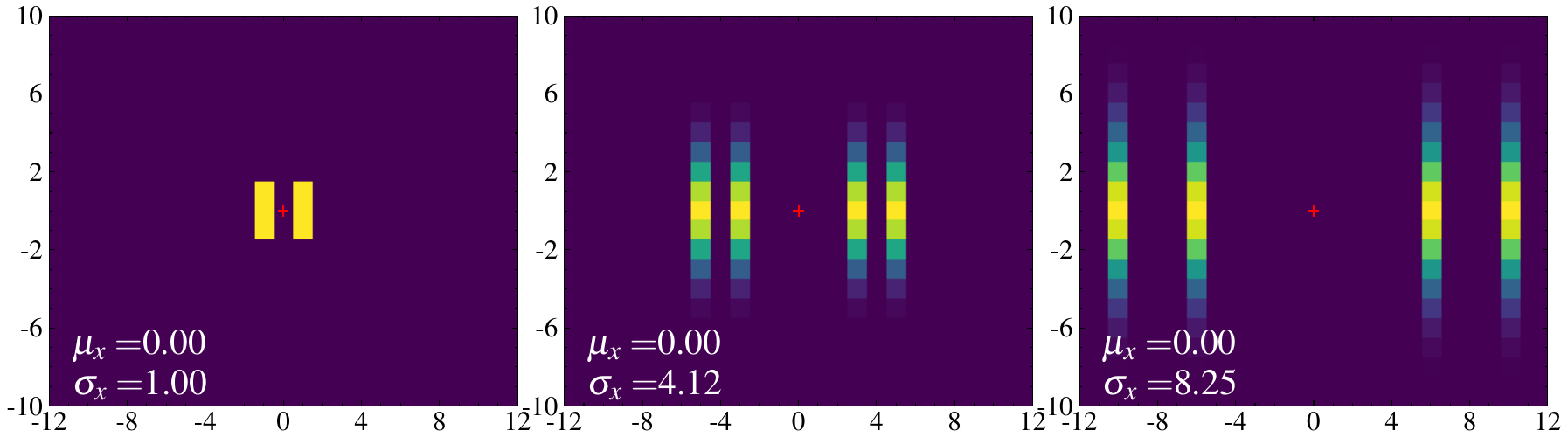}
        \caption{$\nabla_x$ $(\beta^2=1)$}
        \label{fig:circle_mod_uni_3x3_b1}
      \end{subfigure}
    \end{minipage}%

  \caption{Demonstrating the effect of repeated \textbf{convolution+mod} of a test pattern over different types of $3\times3$ kernels (DC and Gradient). Note that for $\beta=0$, the content diffuses symmetrically with a stationary centre of mass, while for $\beta=1$ the pattern propagates symmetrically in both directions at maximum velocity with a stationary centre of mass.}
  \label{fig:circle_mod_uni_3x3}
\end{figure}

\end{document}